\documentclass[10pt,twocolumn,letterpaper]{article}

\usepackage[pagenumbers]{cvpr} %

\usepackage[utf8]{inputenc} %
\usepackage[T1]{fontenc}    %
\usepackage{url}            %
\usepackage{booktabs}       %
\usepackage{amsfonts}       %
\usepackage{nicefrac}       %
\usepackage{microtype}      %
\usepackage{xcolor}         %
\usepackage[normalem]{ulem} %
\usepackage{multirow}
\usepackage[export]{adjustbox}
\usepackage{array}
\usepackage{bbding}
\usepackage{graphicx}
\usepackage{amsmath} 
\usepackage{mathbbol}
\usepackage{csquotes}
\usepackage{diagbox}
\usepackage[toc, title, titletoc]{appendix}
\usepackage{wrapfig}
\usepackage[export]{adjustbox}
\newcommand{\checkheight}[1]{%
	\par \penalty-100\begingroup%
	\setbox8=\hbox{#1}%
	\setlength{\dimen@}{\ht8}%
	\dimen@ii\pagegoal \advance\dimen@ii-\pagetotal
	\ifdim \dimen@>\dimen@ii
	\break
	\fi\endgroup}
\makeatother

\newcommand{\rowname}[1]{
	\rotatebox[origin=c]{90}{{#1}}
}

\usepackage{pifont}     %

\newcommand{\cmark}{\textcolor{green!60!black}{\ding{51}}}  %
\newcommand{\xmark}{\textcolor{red}{\ding{55}}}

\def\eg{\emph{e.g}.~} 

\def\ie{\emph{i.e}.~}

\definecolor{cvprblue}{rgb}{0.21,0.49,0.74}
\usepackage[pagebackref=true,breaklinks=true,colorlinks,allcolors=cvprblue,bookmarks=false]{hyperref}

\def\paperID{4302} %
\def\confName{CVPR}
\def\confYear{2026}

\title{\uline{SABER}: \uline{S}patially Consistent 3D Universal \uline{A}dversarial Objects for \uline{BE}V Detecto\uline{r}s}

\author{
Aixuan Li\textsuperscript{\rm 1},~ 
Mochu Xiang\textsuperscript{\rm 1},~
Bosen Hou\textsuperscript{\rm 1}, ~
Zhexiong Wan\textsuperscript{\rm 1}, ~
Jing Zhang\textsuperscript{\rm 2},~
Yuchao Dai\textsuperscript{\rm 1}~%
\thanks{Corresponding author: Yuchao Dai \emph{(daiyuchao@nwpu.edu.cn)}}
\\
\textsuperscript{\rm 1}School of Electronics and Information, Northwestern Polytechnical University \& \\ Shaanxi Key Laboratory of Information Acquisition and Processing, Xi'an, China \\
\textsuperscript{\rm 2} School of Computing, Australian National University
}

\begin{document}
\maketitle
\begingroup
\renewcommand\thefootnote{}
\footnotetext{Project page: \url{https://npucvr.github.io/SABER}}
\endgroup

\begin{abstract}
	Adversarial robustness of BEV 3D object detectors is critical for autonomous driving (AD). 
	Existing invasive attacks require altering the target vehicle itself (e.g. attaching patches), making them unrealistic and impractical for real-world evaluation. 
	While non-invasive attacks that place adversarial objects in the environment are more practical, current methods still lack the multi-view and temporal consistency needed for physically plausible threats. 
	In this paper, we present the first framework for generating universal, non-invasive, and 3D-consistent adversarial objects that expose fundamental vulnerabilities for BEV 3D object detectors. 
	Instead of modifying target vehicles, our method inserts rendered objects into scenes with an occlusion-aware module that enforces physical plausibility across views and time. 
	To maintain attack effectiveness across views and frames, we optimize adversarial object appearance using a BEV spatial feature-guided optimization strategy that attacks the detector's internal representations. 
	Extensive experiments demonstrate that our learned universal adversarial objects can consistently degrade multiple BEV detectors from various viewpoints and distances.
	More importantly, the new environment-manipulation attack paradigm exposes models' over-reliance on contextual cues and provides a practical pipeline for robustness evaluation in AD systems.

\end{abstract}

\section{Introduction}
3D object detection \cite{wang2021fcos3d,wang2022detr3d} is a foundational component of modern autonomous driving (AD) systems. While sensor-fusion and Lidar-based solutions \cite{li2024bevnext,wang2025hybridbev,bai2022transfusion,li2023bevdepth} often demonstrate state-of-the-art performance, their high hardware costs present a significant barrier to mass-market adoption. 
Consequently, pure vision solutions \cite{li2024bevformer,huang2021bevdet} have garnered immense industry and academic interest, offering a low-cost and highly scalable alternative preferred by many automakers. 
Among these vision-only approaches, 3D object detection models that construct Bird's-Eye-View (BEV) features from multi-view images have emerged as the mainstream framework~\cite{bevsurvey, harley2023simple, ye2025bevdiffuser}. 
As these cost-effective systems are being widely deployed, ensuring their robustness and reliability is of paramount importance \cite{chen2024efficient,tu2021exploring,wang2024attack,zhou2024stealthy}.
Despite their high accuracy, these BEV models remain vulnerable to adversarial perturbations that can cause false 3D bounding boxes \cite{wang2025physically,wang2025unified} or the complete disappearance of an object \cite{liadv3d,lehner20223d, chen2024efficient}, potentially leading to hazardous outcomes.

\begin{figure}
	\vspace{-3mm}
	\centering
	\begin{tabular}{@{}c@{}c@{}c@{}c@{}}
		\includegraphics[width=0.25\linewidth, trim={2 2 2 2}, clip]{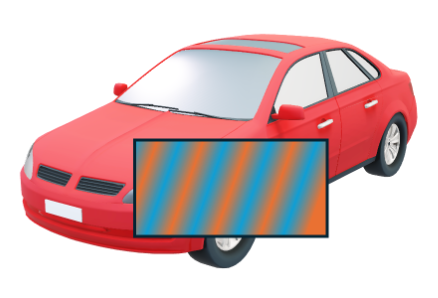} & 
		\includegraphics[width=0.25\linewidth, trim={2 2 2 2}, clip]{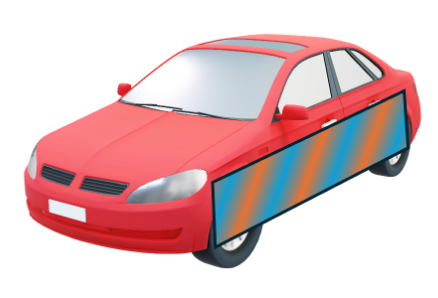} & 
		\includegraphics[width=0.25\linewidth, trim={2 2 2 2}, clip]{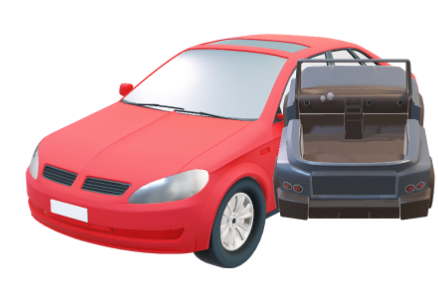} &
		\includegraphics[width=0.25\linewidth]{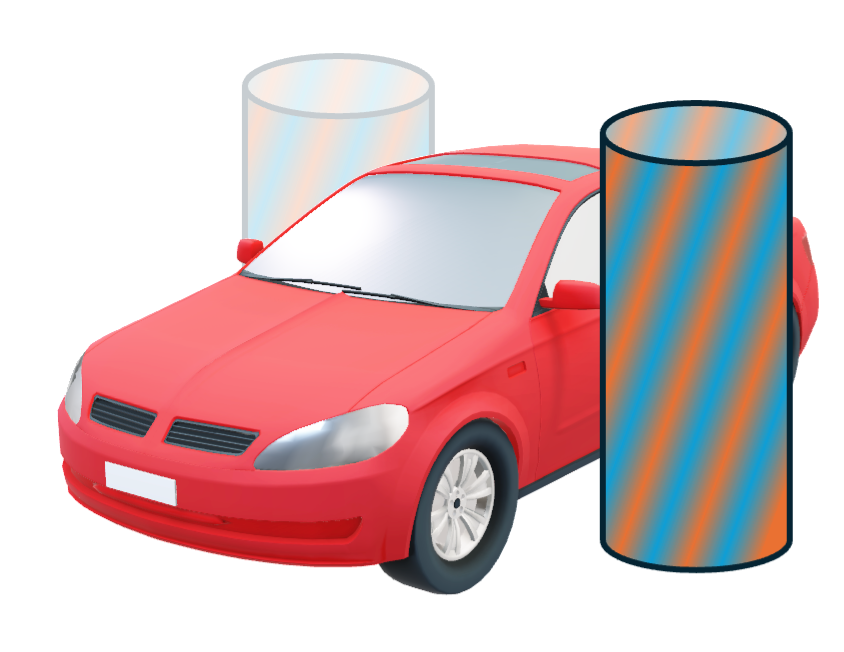} \\
		\small 3D-consistent \xmark & \small 3D-consistent \cmark & \small 3D-consistent \xmark & \small 3D-consistent \cmark \\
		\small ~non-invasive \xmark & \small ~non-invasive \xmark & \small ~~non-invasive \cmark & \small ~non-invasive \cmark \\ \midrule
		\cite{xie2023adversarial} &
		\cite{tu2021exploring, wang2025unified, zhu2023understanding, abdelfattah2021towards} &
		\cite{liadv3d, zhang2021evaluating}  &
		\textbf{Ours}\\
	\end{tabular}
	\vspace{-3mm}
	\caption{\textbf{Comparison of different adversarial attacks for 3D object detection} with respect to 3D-consistency and non-invasiveness. 
	} 
	\label{fig:teaser-image}
	\vspace{-5mm}
\end{figure}

Current adversarial approaches for BEV models are largely ``invasive'' \cite{xie2023adversarial,abdelfattah2021adversarial,tu2021exploring}, as shown in \cref{fig:teaser-image}. These methods require direct physical modification of the target objects, such as applying adversarial textures to vehicles or designing adversarial clothing for pedestrians. This paradigm is fundamentally impractical for two reasons. First, it assumes that the attacker has physical access to the target, which is unfeasible in most scenarios \cite{xie2023adversarial,abdelfattah2021adversarial}. Second, the invasive paradigm lacks scalability, as an attacker cannot possibly modify every vehicle on the road. This limitation is especially critical when aiming for a scene-level attack to cause widespread, unpredictable failures.

We argue that a far more practical and insidious threat comes from \textit{non-invasive environmental manipulation}. 
To formalize this, we define a more plausible and dangerous threat model, termed the scene-level, inter-object attack.
Instead of modifying individual vehicles, the attacker crafts a universal malicious 3D object (a ``rogue mesh'') and places it in the scene to induce systemic, scene-level failures, as \textbf{Ours} in \cref{fig:teaser-image}. 
The mesh exploits a contextual vulnerability: vehicles that perceive it suffer degraded detection of other legitimate vehicles (\eg mis-localization or missed detections), enabling scalable, non-invasive attacks without physical access to targets. 

Furthermore, existing attacks often rely on flat 2D perturbations \cite{wang2025unified,zhu2023understanding}, which overlook the 3D structure of objects, and fail to maintain effectiveness across varying viewpoints. 
Meanwhile, existing 3D object-based attacks \cite{liadv3d} ignore the scene geometry, resulting in unrealistic perspective effects and unreasonable occlusion.
Our core argument is that a practical BEV attack must be 3D-consistent: the adversarial object needs to preserve its malicious properties across views and time for a deployable, multi-view effective BEV attack. 
Addressing this constraint, while ensuring occlusion realism, is the central technical challenge we solve in this work.

To address these challenges, we propose the first framework to systematically study and generate non-invasive, universal, and 3D-consistent adversarial objects.  Specifically, we leverage the mesh's 3D properties to ensure the potential effectiveness of attacks across multiple viewpoints. First, we acquire the vehicles' position from 3D annotation. We then use a differentiable renderer to inject the adversarial mesh near the vehicle, similar to Augmented Reality object insertion, to accurately model the complex spatial relationships within the 3D scene. This non-invasive approach ensures 3D-consistency between the scene and the adversarial object. For multi-frame scenarios, the mesh remains static in its position relative to the scene, allowing for consistent rendering across frames.
Furthermore, to ensure realistic occlusions, we propose a Realistic Occlusion Processing Module that simulates the true occlusion between the attack mesh and other objects in the scene. 
Finally, to achieve efficient scene-level attacks and overcome the limitations of attacks targeting only final bounding box predictions, we augment our objective by introducing BEV feature-based scene confusion, which ensures the robustness of the adversarial effect across varying viewpoints and distances.

Our main contributions are:
\textbf{(1)} We propose the first 3D-consistent, non-invasive threat model where a universal adversarial object, placed nearby without physical contact, can mislead BEV detectors and cause hazards;
\textbf{(2)} We realize this attack with a novel pipeline that leverages differentiable rendering for 3D-consistency, a Realistic Occlusion Processing Module for physical realism, and a BEV feature-based scene confusion loss for robust feature-level attacks; 
\textbf{(3)} 
Results on public datasets and in physical experiments show that our attack
reveals 
a profound semantic vulnerability in current models: our non-invasive object manipulates the model's contextual reasoning about object co-occurrence, exposing a deep over-reliance on learned environmental priors and suggesting significant dataset deficiencies.

\section{Related Works}
\subsection{Attacks Targeting 3D Data}
3D adversarial attack aims to manipulate 3D input data~\cite{zeng2019adversarial,xiao2019meshadv,lu2021towards,eot} to deceive 2D models.
Foundational work showed that physical 3D objects could fool 2D classifiers~\cite{eot}. This concept was soon extended by demonstrating that altering 3D object textures~\cite{zeng2019adversarial} or mesh vertices~\cite{xiao2019meshadv} could also mislead 2D classification and detection. This line of research evolved into the ``adversarial camouflage'' paradigm, which aims to conceal objects from 2D detectors. Methods range from generating stealthy patches~\cite{wang2021dual} to full-coverage, high-resolution textures for vehicles~\cite{wang2022fca}, with later work employing neural renderers to simulate real-world transformations and improve physical realism~\cite{suryanto2022dta}. More recent efforts have focused on improving practicality and transferability, for instance, by creating optimized printable stickers~\cite{oslund2022multiview}, perturbing reconstructed meshes~\cite{huang2024towards}, or using triplanar mapping to generate geometry-independent textures~\cite{suryanto2023active}.

In contrast to prior approaches, our method targets realistically deployable, non-invasive attacks with broader impact. Rather than manipulating the prediction of the attack target itself, we design an adversarial object to disrupt the model’s perception of other vehicle targets without contact. %

\begin{figure*}[!t]
	\centering %
	\includegraphics[width=0.92\textwidth]{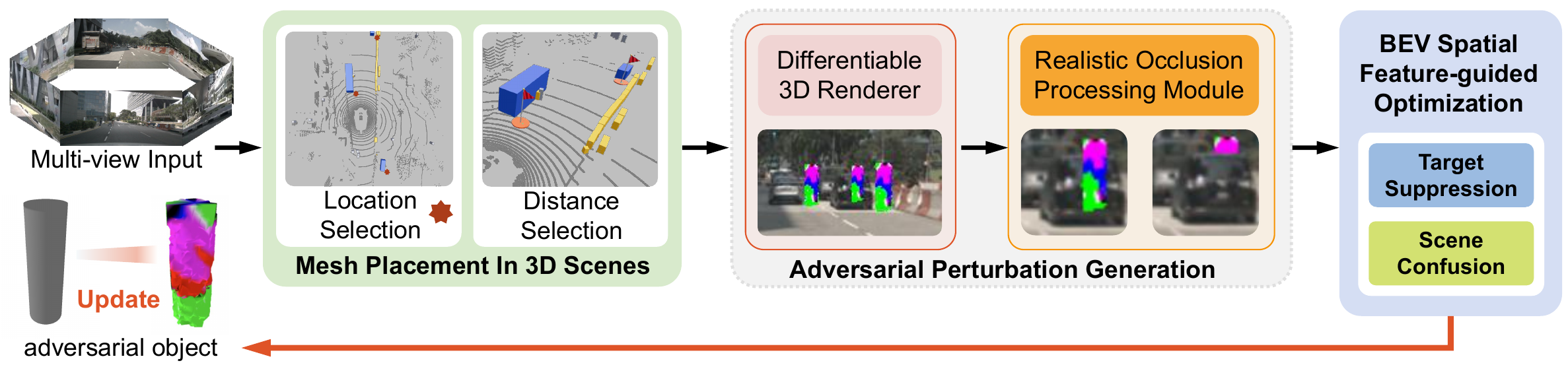} 
	\caption{\textbf{Overview of our adversarial object generation pipeline.} 
		Appropriate locations are first chosen to place adversarial meshes in the 3D scene (\cref{place}), which are then rendered onto the input images. Our differentiable renderer ensures 3D-consistent, multi-view renderings with correct perspective. The \enquote{Realistic Occlusion Processing Module} (\cref{render}) further simulates partial visibility for improved robustness. Finally, the adversarial object is optimized via \enquote{BEV Spatial Feature-Guided Optimization} (\cref{opt}) to enable effective attacks.
	}
	\label{Fig.framework} 
	\vspace{-4mm}
\end{figure*}

\subsection{Non-invasive Adversarial Attack}
As discussed, adversarial camouflage is a mainstream 3D attack, but it is ``invasive'' as it requires modifying the target object. In contrast, non-invasive attacks do not modify the target, leaving no trace and making them harder to detect~\cite{brown2017adversarial,yang2018building}. This concept was explored early in 2D, with foundational work using geometric transformations to create robust patches~\cite{brown2017adversarial}. A key insight came from \cite{yang2018building}, which verified that an attack could succeed even when the perturbation was not on the target object, \eg a human body. This idea of robust, off-target adversarial regions has since been advanced using modern generative models~\cite{casper2022robust}. The concept also extends to other modalities, such as stealthy optical attacks using reflected light~\cite{wang2023rfla}. However, these methods primarily target 2D models. The most relevant work in the 3D-to-3D domain is \cite{liadv3d}, which used a NeRF to generate a 3D adversarial vehicle. Critically, this attack was still rendered from a specific viewpoint and ``pasted'' onto the 2D images, failing to achieve true 3D scene-level consistency and correct geometric integration.

Different from all prior methods, we propose the first non-invasive adversarial attack that is truly \textbf{3D-consistent}. Instead of ``pasting'' a 2D render,
we place a full 3D adversarial mesh into the scene. Using differentiable rendering with a novel occlusion-aware module, we ensure it conforms to the 3D geometry and physics from all viewpoints.

\subsection{Attacks Against 3D Object Detectors}
Adversarial attacks against 3D object detection can be broadly categorized based on their target modality. \textbf{LiDAR-based attacks} \cite{lehner20223d, chen2024efficient, wang2023adversarial, zhang2024comprehensive, wang2025imperceptible} focus on modifying the spatial distribution or quantity of 3D point clouds to deceive perception modules.
\textbf{Camera-LiDAR fusion attacks} \cite{abdelfattah2021adversarial,tu2021exploring,tu2020physically} have explored placing adversarial meshes on vehicles. Key works have optimized both mesh geometry and texture to attack multi-modal pipelines \cite{abdelfattah2021adversarial} or used roof-mounted meshes for stealth \cite{tu2021exploring}. However, these methods are often invasive, as they physically alter the target vehicle's geometry, and are limited to single-view scenarios. Despite these limitations, they highlighted the efficiency of RGB-based attacks, inspiring our focus on the pure-vision domain.

In the \textbf{pure camera-based} domain \cite{xie2023adversarial, wang2025physically, wang2025unified, zhu2023understanding, zhang2021evaluating}, early studies evaluated pixel and patch-level vulnerabilities \cite{zhang2021evaluating, xie2023adversarial, zhu2023understanding}. These works found that BEV representations and multi-frame inputs generally improve robustness against such 2D-space perturbations \cite{xie2023adversarial, zhu2023understanding}.
Subsequently, %
\cite{wang2025physically} optimized a poster placed on the ground to induce false-positive detections, while \cite{wang2025unified} used patches to generate large-scale hallucinated objects.
Recently, Adv3D \cite{liadv3d} employs a NeRF-based renderer~\cite{mildenhall2021nerf} to synthesize and insert entire adversarial vehicles into the scene. However, it suffers from inaccurate occlusion modeling and incorrect perspective consistency.

\section{Methodology}
\label{sec3}

We propose a comprehensive workflow (\cref{Fig.framework}) to attack camera-based BEV 3D object detection, aiming to disrupt the system's predictions of vehicles by placing non-invasive adversarial meshes in proximity to target vehicles within the 3D scene during both optimization and inference. The workflow consists of the following steps:
(1) Automatic selection of suitable regions for adversarial mesh placement (\cref{place}). (2) Adversarial multi-view images generation with perspective projection and occlusion-aware simulation (\cref{render}), ensuring 3D scene consistency. (3) BEV Spatial Feature-Guided Optimization technique (\cref{opt}) to ensure the adversarial mesh remains adversarially effective across time and views.

\subsection{Problem Formulation}

The goal of 3D object detection is to localize and classify objects in 3D. Specifically, given a set of multi-camera images denoted as $\mathcal{I}=\{I_1,I_2,\ldots,I_N\}$, where each image $I_i \in \mathbb{R}^{H \times W \times 3}$ is captured from one of the $N$ calibrated cameras. The objective is to infer the 3D bounding boxes and semantic categories of all related objects in the scene. Each 3D bounding box is represented as $B = \{x, y, z, l, w, h, \theta\}$, where $(x, y, z)$ denotes the coordinates of the object’s center in the 3D space, $\{l, w, h\}$ represents the length, width, and height of bounding box, and $\theta$ denotes the yaw angle of the object. The corresponding object category is denoted by $\tau$.

The conventional optimization mechanism for attacking 3D object detection model~\cite{xie2023adversarial,wang2025imperceptible} aims to optimize the perturbation by maximizing the model's prediction error for the target class $\tau_t$ (in our case, vehicles). 
The objective function for the attack is formulated as:
\begin{equation}
	\begin{split}
		\mathcal{L}_{\text{attack}}(\mathcal{I}_{\text{adv}}, & B_{\tau_t}, f) = \alpha\mathcal{L}_{\text{cls}}(f(\mathcal{I}_{\text{adv}}), \tau_t) \\
		&\quad + \beta  \mathcal{L}_{\text{loc}}(f(\mathcal{I}_{\text{adv}}), B_{\tau_t})  + \lambda \cdot |\delta|_p,
	\end{split}
\end{equation}
where $\mathcal{L}_{\text{cls}}$ penalizes the model $f(\cdot)$ for correctly predicting targets, \ie to reduce
the confidence score of $\tau_t$. $\mathcal{L}_{\text{loc}}$ encourages incorrect positions. The  adversarial images $\mathcal{I}_{\text{adv}}$ are constrained by the raw images $\mathcal{I}$ by $|\delta|_p \leq \epsilon$, where $\epsilon$ is the maximum perturbation magnitude allowed and $p$ is typically 2 or $\infty$. $\alpha$ and $\beta$ are the weights. By minimizing the attack loss $ \mathcal{L}_{\text{attack}}$, the attack targets either disappear from predictions or appear at incorrect positions.

\subsection{Mesh Placement in 3D Scenes}
\label{place}

We explore whether a 3D object can consistently execute multi-frame and multi-view adversarial attacks in realistic 3D scenes for non-contact 3D object detection attack (see \cref{Fig.framework}). Unlike Adv3D~\cite{liadv3d}, which employs NeRF~\cite{mildenhall2021nerf} to optimize only vehicle textures, 
our mesh-based approach does not assume any object category
and allows for explicit modeling of perspective occlusions. Furthermore, by adopting mesh representations, we gain direct surface control and improved compatibility with physics-based environments.

A 3D adversarial mesh $\mathcal{M}$ is represented as a set of vertices $\mathcal{V_M}=\{v_j\}$, triangular faces $\mathcal{F_M}=\{f_j\}$, and optional vertex textures mapping to a texture map $\mathcal{T_M}$.
To realistically insert the mesh into a 3D scene, we place it adjacent to a target vehicle using its 3D bounding box $B_{veh}$, ensuring consistent alignment across views. We decompose the placement into location selection and distance computation.

\noindent\textbf{Location Selection.} The mesh is placed near one of the bottom corners of the target vehicle's bounding box $B_{veh}$. The 8 corners of the box, $C_{{veh}} = \{c_{{veh},j}\}_{j=1}^8$, are computed by offsetting from the vehicle's center $(x, y, z)$ and accounting for its orientation $\theta$. For example, the right-rear-bottom corner $c_{{veh},1}$ is calculated as $(x - \frac{l}{2}\cos\theta - \frac{w}{2}\sin\theta, y - \frac{l}{2}\sin\theta + \frac{w}{2}\cos\theta, z - \frac{h}{2})$. Other corners are defined analogously, with varying combinations of sign on ${\frac{l}{2}},{\frac{w}{2}},{\frac{h}{2}}$.

To ensure that the mesh is tangent to but not intersecting the vehicle, we define the position of the mesh 
as $p_{\mathcal{M}}$. Let $v_{\vec{n}_{\text{max}}}$ be the maximum distance from the mesh center to any vertex along a direction $\vec{n}$. When positioning from a corner $c_{{veh},j}$ along the direction $\vec{n}$, the mesh center is computed as: $p_{\mathcal{M}} = c_{{veh},j} + v_{\vec{n}_{\text{max}}} \cdot \hat{n}$, where $\hat{n}$ is the distance to $\vec{n}$. When the target vehicle is the same as the ego vehicle, the positioning direction from $c_{{veh},1}$ is $\vec{n} = -y$.

\noindent\textbf{Distance to Target.} The distance $d$ between the mesh and the vehicle is the Euclidean distance from the mesh position $p_{\mathcal{M}}$ to the reference corner $c_{{veh},j}$. For $d > 0$, the mesh is further offset: $p_{\mathcal{M}} = c_{{veh},j} + (v_{\vec{n}_{\text{max}}} + d) \cdot \hat{n}$.
The influence of $d$ is discussed in \cref{sec_abu} and \cref{dis_robust}. %

\subsection{Adversarial Perturbation Generation}
\label{render}

After placing the adversarial mesh, we render it onto $N$ calibrated camera images.
The mesh is centered at $p_{\mathcal{M}}$ 
in the world coordinate. For each of the $N$ cameras, the extrinsics $(R_i, T_i)$ map world points to its coordinate system.
For a mesh vertex $v_j$ in global coordinates, the transformation to the $i$-th camera's frame is: $v_j^{\text{cam}_i} = R_i \cdot v_j + T_i$, where $v_j^{\text{cam}_i}$ is the $j$-th vertex in the $i$-th camera coordinate frame.

We then employ the differentiable rendering pipeline PyTorch3D \cite{ravi2020accelerating} to project 3D vertices to 2D image coordinates via the camera’s intrinsic matrix $K_i$. Each vertex  $v_j^{\text{cam}_i}$ on mesh $\mathcal{M}$ is mapped to a 2D point, producing an RGB image $I_{\mathcal{M},i}^{\text{rgb}} \in \mathbb{R}^{H \times W \times 3}$ and a soft mask $I_{\mathcal{M},i}^{\text{mask}} \in \mathbb{R}^{H \times W \times 1}$, where the mask represents the mesh’s visibility and transparency in the rendered view.

\noindent\textbf{Realistic Occlusion Processing Module.}
The above 3D-consistent outputs ($I_{\mathcal{M},i}^{\text{rgb}}$, $I_{\mathcal{M},i}^{\text{mask}}$) do not account for real-world occlusions alone. Therefore, we handle occlusions per view 
between the mesh and
annotated scene objects $\{\mathcal{O}_k\}$.
Occlusion happens when one object overlaps another in the camera's view while also being closer in 3D space.
A two-stage filter is applied to identify the specific subset of true occluders, $\{\mathcal{O}_{occ}\} \subseteq \{\mathcal{O}_k\}$, which combines a 2D check for visual overlap with a BEV check for 3D spatial distance.

We first perform a coarse 2D check in each view $i$.
For this, we project a mesh $\mathcal{M}$ and each object $\mathcal{O}_k$ to get their 2D bounding boxes $B_{\mathcal{M},i}^{\text{2D}}$ and $B_{\mathcal{O}_k,i}^{\text{2D}}$, through the camera intrinsic and extrinsic matrix. We compute their overlap as:
\begin{equation}
	\text{Occ}_{\text{2D}}(B_{\mathcal{M},i}^{\text{2D}}, B_{\mathcal{O}_k,i}^{\text{2D}}) =  
	B_{\mathcal{M},i}^{\text{2D}} \cap  B_{\mathcal{O}_k,i}^{\text{2D}}.
\end{equation} 

If $\text{Occ}_{\text{2D}} > 0$, we consider the object $\mathcal{O}_k$ a candidate for occlusion and proceed with a more rigorous geometric check in the BEV plane to resolve the depth ambiguity.
Specifically, for any object $\mathcal{O}_k$ in the scene, we project its eight 3D bounding box corners $C_{\mathcal{O}_k} = \{c_{{\mathcal{O}_k},j}\}_{j=1}^8$ onto the BEV view. The BEV bounding box $B_{\mathcal{O}_k}^{\text{BEV}}$ is defined as the convex hull of its projected 3D corners: $B_{\mathcal{O}_k}^{\text{BEV}} = \text{Hull}(\Pi_{\text{BEV}}(C_{\mathcal{O}_k}))$. 
Next, we construct a visibility cone for our adversarial mesh $\mathcal{M}$ from the $i$-th camera origin $O_{\text{cam},i}$ in the BEV plane, denoted $\mathcal{F}_{\mathcal{M},i}^{\text{BEV}}$.  
This visibility cone is defined as the 2D convex hull of the union of two point sets: the BEV-projected camera origin $\Pi_{\text{BEV}}(O_{\text{cam},i})$, and the set of all BEV-projected mesh vertices $\Pi_{\text{BEV}}(\mathcal{V_M})$:
\begin{equation}
	\mathcal{F}_{\mathcal{M},i}^{\text{BEV}} = \text{Hull}\left (\Pi_{\text{BEV}}(O_{\text{cam},i}) \cup \Pi_{\text{BEV}}(\mathcal{V_M}) \right ). 
\end{equation}

The  BEV plane occlusion condition $\text{Occ}_{\text{BEV}}$ is defined as $B_{\mathcal{O}_k}^{\text{BEV}} \cap \mathcal{F}_{\mathcal{M},i}^{\text{BEV}} \neq \emptyset$.
An object $\mathcal{O}_k$ is formally identified as an occluder for a mesh $\mathcal{M}$ and is included in the set $\{\mathcal{O}_{occ}\}$, if both $\text{Occ}_{\text{BEV}}$ and $\text{Occ}_{\text{2D}}$ are satisfied (schematic in \cref{sec:real_occ_check}).

To correctly handle occlusions from multiple objects, we employ a two-stage process for each camera view $i$. First, we introduce the mask update process, using a single mesh $\mathcal{M}$ as an example. We identify the set of objects $\{\mathcal{O}_{occ}\}$ that occlude the mesh $\mathcal{M}$. For each occluder $\mathcal{O}_k \in \{\mathcal{O}_{occ}\}$, we prompt the SAM2 model \cite{ravi2024sam} with its 2D bounding box $B_{\mathcal{O}_k,i}^{2D}$ to acquire its segmentation mask $M_k^{\text{seg}}$. Then these masks are iteratively applied to update the transparency mask $I_{{\mathcal{M}},i}^{\text{mask}}$ of $\mathcal{M}$ as follows:
\begin{equation}
	\label{eq:mask_update}
	I_{\mathcal{M},i}^{\text{mask}} \leftarrow I_{{\mathcal{M}},i}^{\text{mask}} \cdot (1 - M_k^{\text{seg}}).
\end{equation}

Second, we generalize this process to a multi-mesh scene to compose the final image. For meshes $\{\mathcal{M}_s\}$, we generate the final pre-occluded masks $\{I_{\mathcal{M}_s,i}^{\text{mask}}\}$ by applying \cref{eq:mask_update} conditionally: only the masks of occluded meshes are updated, while unoccluded ones remain unchanged.
After all mesh masks $I_{\mathcal{M}_s,i}^{\text{mask}}$ have been pre-occluded, the meshes also need to blend correctly with each other. Thus, we use a process akin to the classic Painter's Algorithm~\cite{newell1972hidden}. We sort the list of adversarial meshes $\{\mathcal{M}_s\}$ in a \textit{far-to-near order} based on their depth. We initialize a canvas with the raw scene image, $I_i^{\text{canvas}} \leftarrow I_i^{\text{raw}}$. Following the sorted order, we sequentially paint each mesh $\mathcal{M}_s$ (using its rendered image $I_{\mathcal{M}_s,i}^{\text{rgb}}$ and pre-occluded mask $I_{\mathcal{M}_s,i}^{\text{mask}}$) onto the canvas using the alpha blending equation:
\begin{equation}
	\label{eq:painter_update}I_i^{\text{canvas}} \leftarrow I_{\mathcal{M}_s,i}^{\text{rgb}} \cdot I_{\mathcal{M}_s,i}^{\text{mask}} + (1 - I_{\mathcal{M}_s,i}^{\text{mask}}) \cdot I_i^{\text{canvas}},
\end{equation}
which ensures that nearer meshes correctly occlude farther ones. The final state of the canvas becomes the final adversarial image $I_i^{\text{adv}} = I_i^{\text{canvas}}$, and this process ultimately generates the final multi-camera adversarial set $\mathcal{I}_{\text{adv}} = \{ I_i^{\text{adv}} \}_{i=1}^N$.

\subsection{BEV Spatial Feature-guided Optimization}
\label{opt}
Our adversarial attack targets 3D object detectors by pursuing two primary goals: (1) suppressing correct vehicle predictions within a designated target region, and (2) inducing false predictions in non-target regions to disrupt overall scene understanding. To this end, we optimize the shape and appearance of an adversarial object by minimizing a composite loss function consisting of three components.

\noindent\textbf{Target Suppression.}
To suppress correct predictions within the annotated region, we minimize the sum of confidence responses in the target area:
\begin{equation} %
	\mathcal{L}_{\text{cls}} = \sum \left ( f(\mathcal{I}_{\text{adv}})_{confidence} \times \mathcal{R}_{\tau_t} \right),
\end{equation}
where $\mathcal{R}_{\tau_t}$ denotes the 2D projection of the target 3D box $B_{\tau_t}$ onto the confidence space, $f$ denotes the detection model and $\mathcal{I}_{\text{adv}}$ is the rendered adversarial multi-view images. 

Additionally, to disrupt accurate 3D localization, we maximize the discrepancy between the predicted 3D bounding box and the target vehicle ground-truth box $B$ using L1 loss:
\begin{equation}
	\mathcal{L}_{\text{loc}} = - \left|f(\mathcal{I}_{\text{adv}})_{\hat{B}} - B_{\tau_t} \right|, 
\end{equation}
where$f(\mathcal{I}_{\text{adv}})_{\hat{B}}$ represents the 3D bounding box predicted by the model in the region $B_{\tau_t}$. \\
\noindent\textbf{Scene Confusion.}
To encourage false positives in irrelevant regions, we disturb the global scene representation by minimizing the cosine similarity between BEV features extracted from the adversarial image and the unperturbed image:
\begin{equation}
	\mathcal{L}_{\text{sim}} =\frac{f_\phi(\mathcal{I}_{\text{adv}} )\cdot f_\phi(\mathcal{I}_{\text{raw}} )}{\left \|  f_\phi(\mathcal{I}_{\text{adv}} )\right \|\times\left \|f_\phi(\mathcal{I}_{\text{raw}} )\right \| },
\end{equation}
where $f_\phi(\cdot)$ denotes the BEV feature extractor in model $f$, and $\mathcal{I}_{\text{raw}}$ is the original multi-view image.

The overall objective combines all three components:
\begin{equation}
	\label{final_loss}
	\begin{split}
		\min_{\mathcal{V},\mathcal{T}} & \mathcal{L}_{\text{attack}}(\mathcal{V},\mathcal{T}) = \mathcal{L}_{\text{cls}}(\mathcal{I}_{\text{adv}}(\mathcal{V},\mathcal{T})) \\ + & \alpha \mathcal{L}_{\text{loc}}(\mathcal{I}_{\text{adv}}(\mathcal{V},\mathcal{T})) 
		+ \beta \mathcal{L}_{\text{sim}}(\mathcal{I}_{\text{adv}}(\mathcal{V},\mathcal{T})),
	\end{split}
\end{equation}
where $\alpha$ and $\beta$ balance the localization and scene confusion terms against the classification loss, and are both set to 10.
This formulation ensures the adversarial object prevents correct detection in the target region and induces false detections elsewhere, effectively disrupting the model’s scene-level reasoning. We optimize mesh vertices $\mathcal{V}$ and textures $\mathcal{T}$ with \cref{final_loss} to obtain effective attacks across viewpoints and time.

\section{Experiments}
\subsection{Experimental Setting}
\label{exp_setting}
\noindent\textbf{Dataset.} 
We use the nuScenes dataset \cite{caesar2020nuscenes}, a large-scale autonomous driving benchmark with 3D detection annotations and six calibrated cameras covering the full $360^\circ$ horizontal field of view.
Our 3D adversarial perturbations are trained on 28,130 keyframes from the training set and evaluated on 6,019 keyframes from the validation set. 

\begin{table*}[tp]
	\centering
	\setlength{\tabcolsep}{2pt} 
	\caption{\textbf{Performance comparison of BEV-based 3D object detection with our adversarial attack setting.} }
	\label{tab:white_box}
	\resizebox{0.9\linewidth}{!}{
		\renewcommand{\arraystretch}{0.8}
		\centering
		\begin{tabular}{l|c|ccc|ccc|ccc} 
			\toprule
			& Real Occ & Clean NDS &  Init NDS & Adv NDS & Clean mAP &Init mAP &Adv mAP    & $\text{ASR}_{0.3}$ & $\text{ASR}_{0.5}$ &  $\text{ASR}_{0.7}$  \\
			\midrule
			BEVDet~\cite{huang2021bevdet} &   \xmark  & 0.3942 &  0.3579 & 0.2097 & 0.3086 &0.2625 &0.1298    &  0.613    &   0.657 & 0.720   \\
			BEVDet~\cite{huang2021bevdet} & \cmark   & 0.3942 & 0.3682 & 0.2668 & 0.3086 & 0.2754 & 0.1597 &  0.454 & 0.515  & 0.555  \\
			BEVDet4D~\cite{huang2022bevdet4d} & \xmark  & 0.4471   &0.4158 & 0.2762 & 0.3138 &0.2799 &0.1564   & 0.530 &   0.590 &  0.670  \\
			BEVDet4D~\cite{huang2022bevdet4d} & \cmark & 0.4471 & 0.4268 & 0.3355 & 0.3138  & 0.2883 & 0.1838 &  0.438   &  0.501   &  0.586   \\ %
			BEVFormer~\cite{li2024bevformer} & \xmark  & 0.4784  &  0.4592 &  0.2876 & 0.3703 &0.3402 & 0.1652 &   0.405 &  0.472  &  0.621  \\
			BEVFormer~\cite{li2024bevformer} & \cmark & 0.4784  &  0.4654 &  0.3398   & 0.3703 &  0.3520  &  0.2097  & 0.245  &  0.322 & 0.467  \\
			
			\bottomrule
	\end{tabular}}
\end{table*}

\begin{table*}[tp]
	\centering
	\caption{\textbf{Performance comparison with Adv3d~\cite{liadv3d}}. Drop ($\downarrow$) highlights our attack effectiveness with significant performance degradation.}
	\resizebox{0.9\linewidth}{!}{ %
		\renewcommand{\arraystretch}{0.8} %
		\begin{tabular}{l|cccc|cccc|ccc} 
			\toprule
			\multirow{2}{*}{Method} & \multicolumn{4}{c|}{\textbf{NDS}} & \multicolumn{4}{c|}{\textbf{mAP}} & \multicolumn{3}{c}{\textbf{AP per Class}} \\
			\cmidrule(lr){2-5} \cmidrule(lr){6-9} \cmidrule(lr){10-12} %
			& Clean & Init & Adv & Drop ($\downarrow$) & Clean & Init & Adv & Drop ($\downarrow$) & car & truck & bicycle \\
			\midrule
			Ours  & 0.3942& 0.3579 & 0.2097 & \textbf{41.4\%} & 0.3086&  0.2625 & 0.1298 & \textbf{55.6\%} & 0.153 & 0.034 & 0.003 \\
			Adv3d~\cite{liadv3d} & 0.3942 & 0.2820 & 0.2277 & 19.3\%  & 0.3086 & 0.1727 & 0.0967 & 44.0\%          & 0.173 & 0.064 & 0.046 \\
			\bottomrule
	\end{tabular}}
	\label{tab:adv3d}
	\vspace{-8pt}
\end{table*}

\begin{table}[tp]
	\centering
	\caption{\textbf{Performance comparison with UAP~\cite{wang2025unified}.} }
	\resizebox{0.9\linewidth}{!}{
		\renewcommand{\arraystretch}{0.8}
		\begin{tabular}{l|cccc}
			\toprule
			Method & $\text{ASR}_{0.1}$ & $\text{ASR}_{0.3}$ & $\text{ASR}_{0.5}$ &  $\text{ASR}_{0.7}$ \\
			\midrule
			Ours & 0.568 & 0.613 & 0.657 & 0.720 \\
			UAP~\cite{wang2025unified}& 0.405  & 0.514 & 0.613 & 0.759 \\
			\bottomrule
		\end{tabular}
	}
	\vspace{-8pt}
	\label{tab:uap}
\end{table}

\begin{figure*}[tp]
	\centering
	\small
	\begin{tabular}{{c@{ } c@{ } c@{ } c@{ } c@{ }c@{ } c@{ }  }}
		{\includegraphics[width=0.185\linewidth, height=0.110\linewidth]{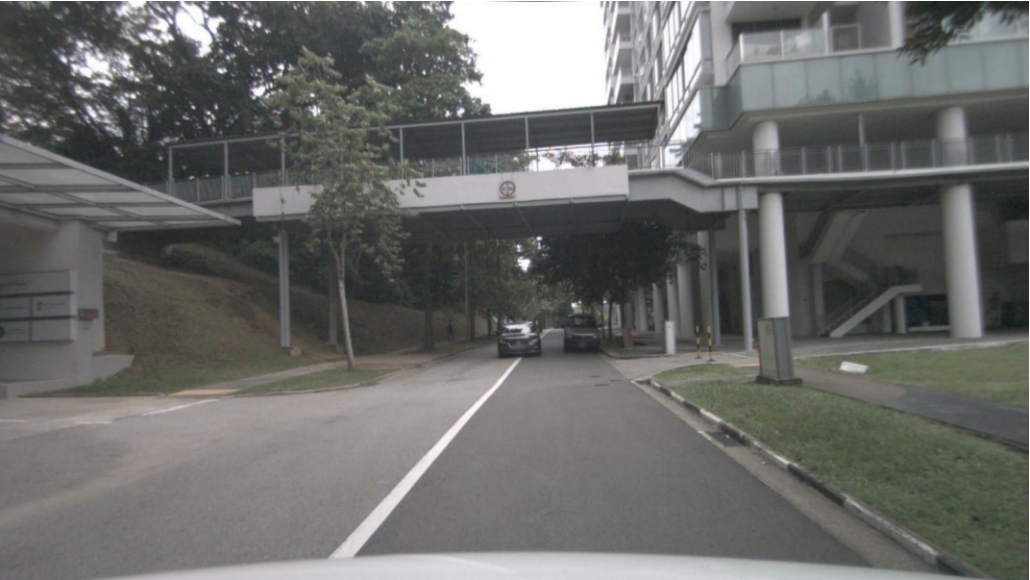}}&
		\multirow{1}*[11mm]{\rotatebox[origin=c]{90}{Init}} & 
		{\includegraphics[width=0.185\linewidth, height=0.110\linewidth]{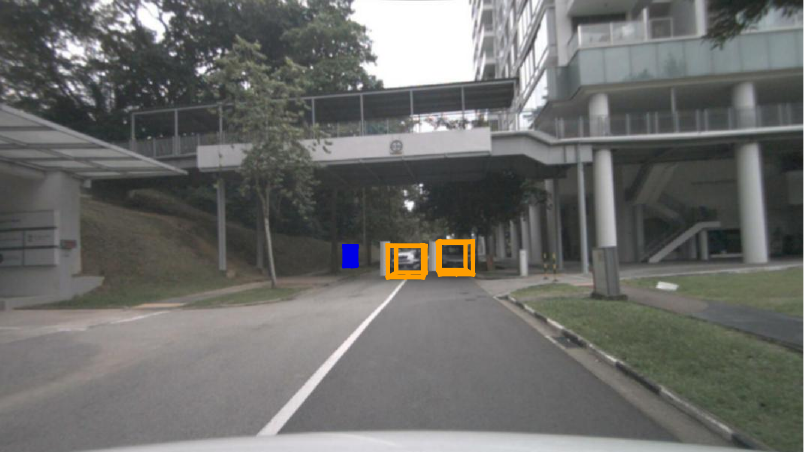}}&
		{\includegraphics[width=0.185\linewidth, height=0.110\linewidth]{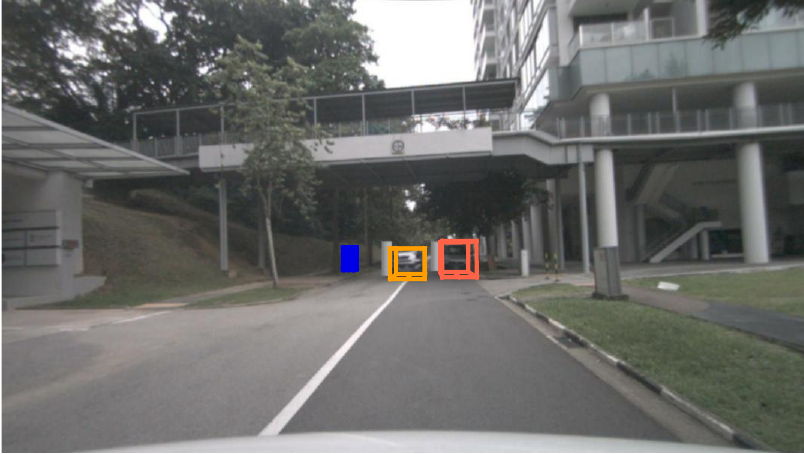}}&
		{\includegraphics[width=0.185\linewidth, height=0.110\linewidth]{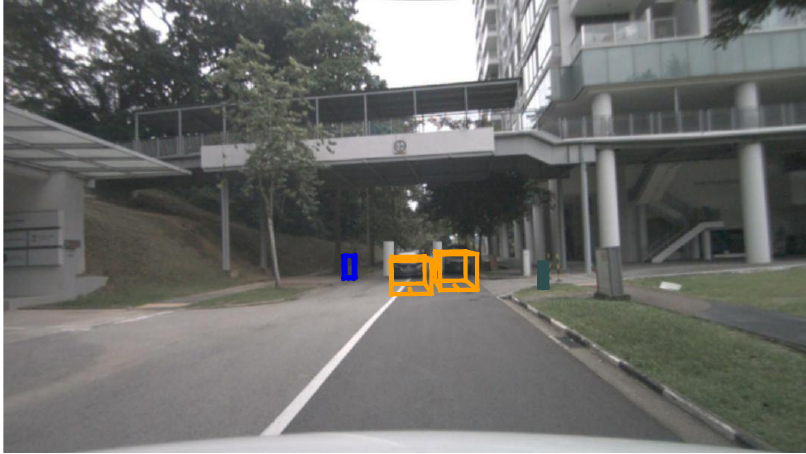}}&
		{\includegraphics[width=0.185\linewidth, height=0.110\linewidth]{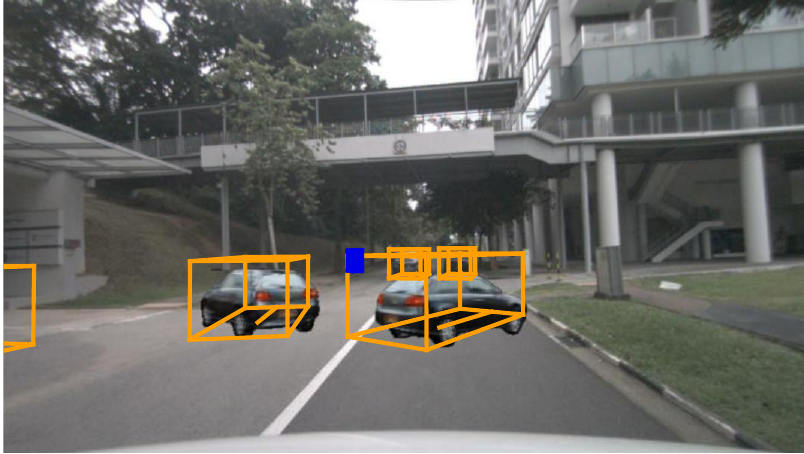}}
		\\
		
		{\includegraphics[width=0.185\linewidth, height=0.110\linewidth]{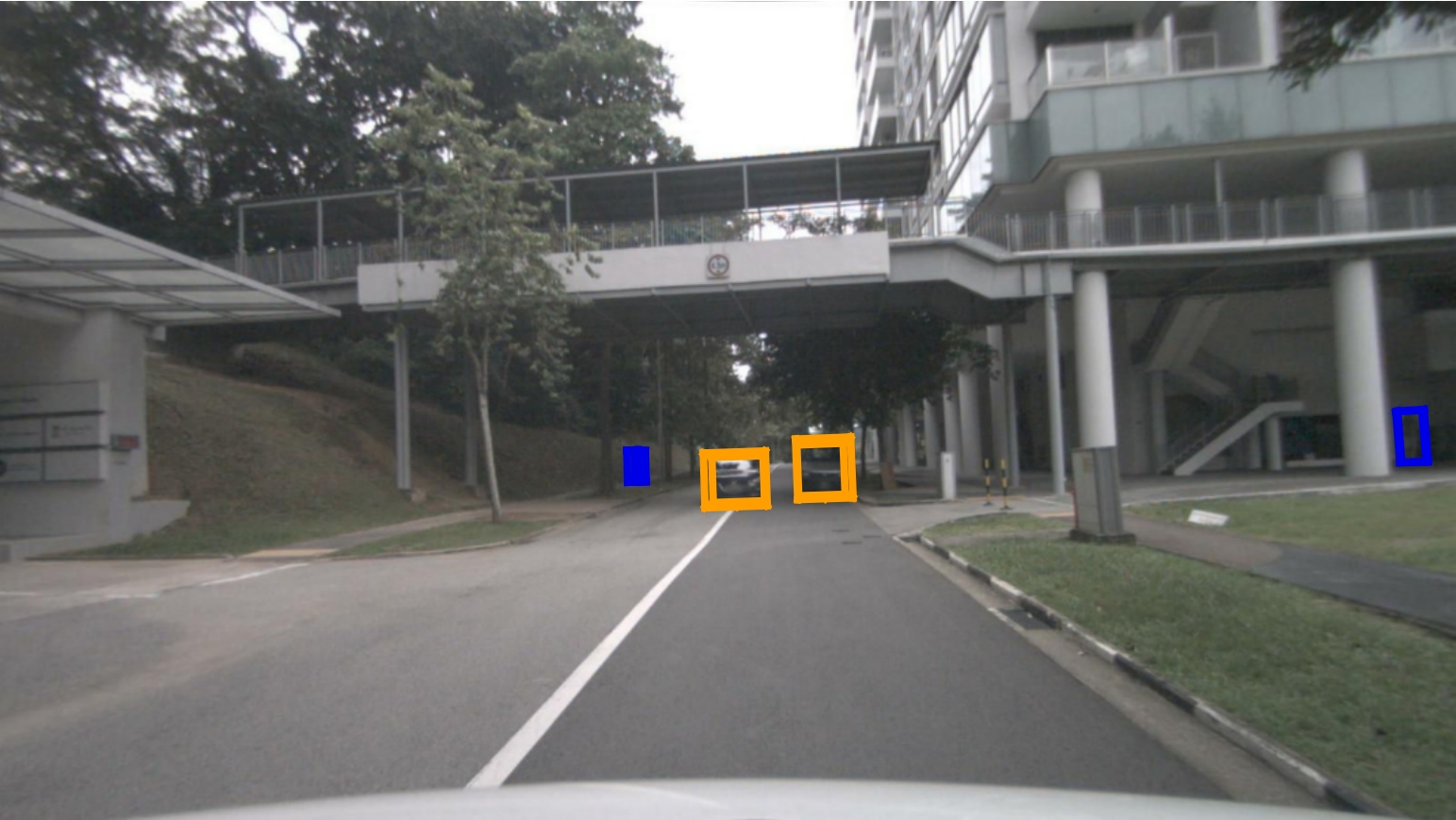}}&
		\multirow{1}*[13mm]{\rotatebox[origin=c]{90}{Attack}} & 
		{\includegraphics[width=0.185\linewidth, height=0.110\linewidth]{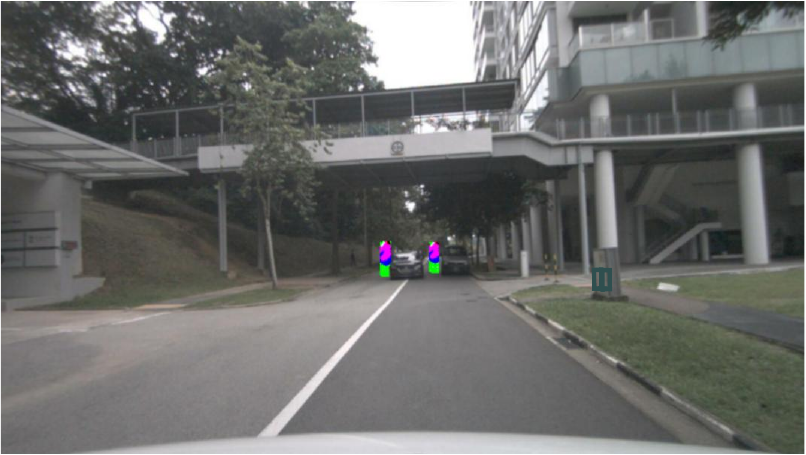}}&
		{\includegraphics[width=0.185\linewidth, height=0.110\linewidth]{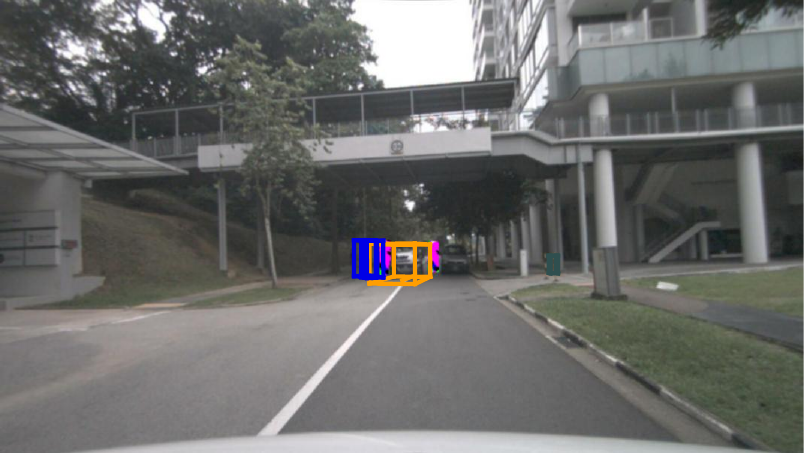}}&
		{\includegraphics[width=0.185\linewidth, height=0.110\linewidth]{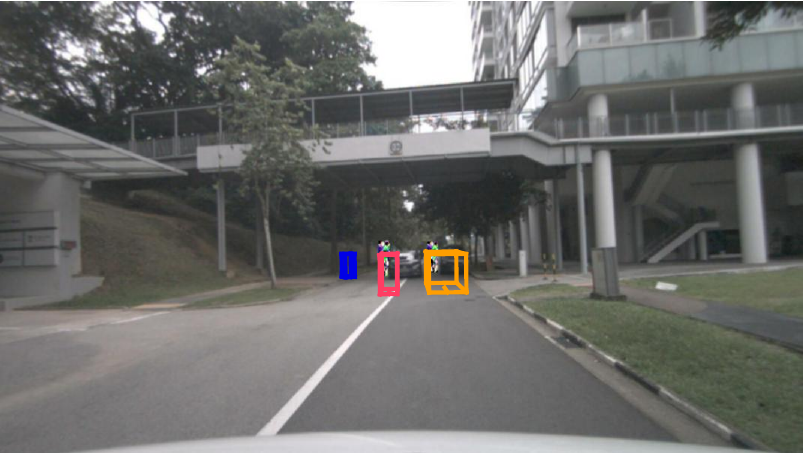}}&
		{\includegraphics[width=0.185\linewidth, height=0.110\linewidth]{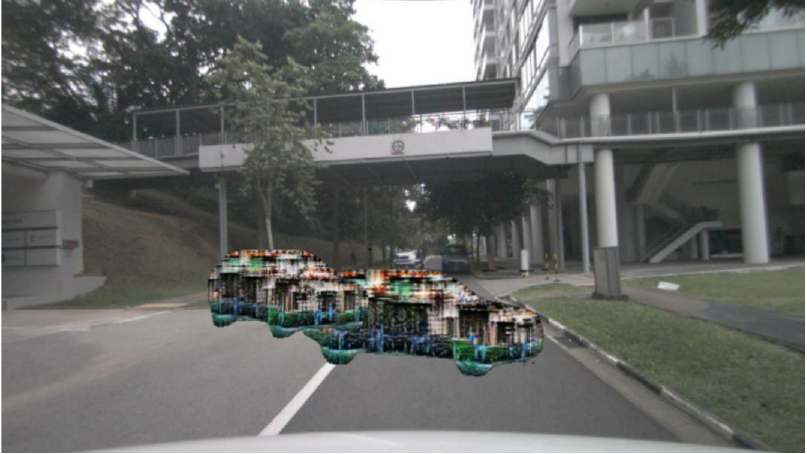}}
		\\
		
		{GT} & 
		& (a) Ours-BEVDet & (b) Ours-BEVDet4D & (c) Ours-BEVFormer  & (d) Adv3D-BEVDet \\
	\end{tabular} 
	\caption{\textbf{Visual Comparisons of attack effects with Adv3D~\cite{liadv3d}.} ``Init'' and ``Attack'' represent the detection of placing gray initial meshes and adversarial meshes. Results are labeled as [Method]-[Model] (\eg Ours-BEVDet denotes ``Our'' method on the ``BEVDet'' model).} 
	\label{fig:2d_compare}
	\vspace{-8pt}
\end{figure*}

\noindent\textbf{Evaluation Metrics.} We assess attack effectiveness using the Attack Success Rate (ASR), defined as $\mathrm{ASR}=1-{N_{\mathrm{adv}}}/{N_{\mathrm{init}}}$, where $N_{\mathrm{adv}}$ and $N_{\mathrm{init}}$ denote the number of successfully detected vehicles with adversarial and initial objects, respectively. 
A detection is regarded successful if its Intersection over Union (IoU)  with the ground truth exceeds a threshold.
We report ASR across IoU thresholds ranging from 0.3 to 0.7 to account for varying localization strictness. 
Furthermore, we assess the scene-level detection impact of our adversarial objects on the scene by evaluating the mean Average Precision (mAP) and the nuScenes Detection Score (NDS) \cite{caesar2020nuscenes}.

\noindent\textbf{Victim Models.} We train non-invasive 3D adversarial objects on three popular camera-based 3D detectors: BEVDet~\cite{huang2021bevdet}, BEVDet4D~\cite{huang2022bevdet4d}, and BEVFormer~\cite{li2022bevformer, li2024bevformer}. Following their original settings, both BEVDet and BEVDet4D adopt ResNet-50 backbone and utilize class-balanced grouping and sampling~\cite{zhu2019class} with $256 \times 704$ resolution of input images. BEVFormer uses ResNet-101 backbone with $736 \times 1280$ inputs. 
\cref{liadr_attack} also analyze the LiDAR-supervised training \& camera-only inference setup.

\noindent\textbf{Implementation Details.} 
We initialize the mesh as a \textit{cylinder} (radius $0.3$, height $2.0$). For quantitative evaluation, unless otherwise noted, we adopt a fixed object placement strategy, positioned $0.1$ meter from the target vehicle’s right-rear-bottom corner, to ensure controlled comparisons across scenes. See \cref{Implem}  for more details.

\subsection{Quantitative Results}

We present a comparison of white-box attack results on three victim models in \cref{tab:white_box}. 
``Clean'' denotes the baseline performance on the original, unmodified scenes. 
``Init'' reports performance after inserting a non-adversarial object (gray regular cylinder), while ``Adv'' shows results with our optimized adversarial mesh. 
``Real Occ'' denotes that our real occlusion handling module is enabled. 
This setup separates two effects: the drop from ``Clean'' to ``Init'' measures benign occlusion, and the further drop from ``Init'' to ``Adv'' demonstrates that our attack manipulates the model’s contextual reasoning about object co-occurrence, which in turn exposes its over-reliance on environmental priors. 
Our universal mesh maintains effective ASR across severity thresholds (0.3–0.7) and shows consistent effectiveness across views, frames, models, and vehicle types, indicating robust and consistent scene-level attack performance. 

Given that our non-invasive, environment-manipulation setting has not been studied before, direct evaluations are not available. 
Therefore, we perform comparisons below for two state-of-the-art baselines that are related yet distinct. 

\noindent\textbf{Comparison with a Non-Invasive Baseline (Adv3D).}
We first compare our method against Adv3D~\cite{liadv3d}, the closest existing non-invasive paradigm. As shown in \cref{tab:adv3d}, our analysis reveals not only the superiority of our targeted approach but also a fundamental weakness in Adv3D's evaluation. We observe that Adv3D's baseline performance is significantly compromised by its setting, which randomly renders two cars per view and introduces severe self-occlusions, causing a significant initial drop in NDS/mAP even before the attack. After accounting for this discrepancy, our method demonstrates substantially greater attack efficacy, achieving a remarkable 41.4\% drop in NDS and a 55.6\% drop in mAP. These results far exceed the drops induced by Adv3D, proving our method's superior effectiveness. By focusing specifically on the vehicle category, we achieve a lower Average Precision (AP) on the intended targets, which demonstrates a stronger, more focused adversarial attack.

\noindent\textbf{Conceptual Comparison with an Invasive Patch-Based Baseline (UAP).} We also compare with UAP~\cite{wang2025unified}, a leading invasive patch-based method. However, UAP's threat pattern is fundamentally different: 
unlike our targeting \textbf{multiple vehicles at various distances} setup, UAP~\cite{wang2025unified} 
only places 2D patch directly onto the \textbf{nearest one vehicle} in each view, a setting that inherently minimizes the chance of the patch being occluded. To ensure the most equitable comparison against UAP's less-occluded setting, we evaluate it against our method without Real Occ. applied (denoted as ``Ours'' in \cref{tab:uap}). The results show our method's superior ability to cause significant, practical perception failures. Specifically, our method significantly outperforms UAP at the lower and medium IoU thresholds ($ASR_{\le 0.5}$), which represent critical failures in detection systems. While UAP's invasive methodology, directly modifying the vehicle's surface, achieves a higher ASR at the strictest 0.7 IOU threshold, we argue this efficacy is limited to minor, fine-grained manipulations, not critical system failures. In contrast, our method, designed for real-world complexities and view-consistency, excels at causing the more significant, lower-IoU perception failures. %

\subsection{Qualitative  Results}
\begin{figure}[tp]
	\centering
	\setlength{\tabcolsep}{0pt} 
	\begin{tabular}{@{}c@{}c@{}c@{}c@{}c@{}c@{}} 
		
		& \small BEVFormer \quad & \small BEV4D \quad & \small BEVDet \quad & \small cube \quad & \small sphere \\
		\noalign{\smallskip\smallskip} 
		\raisebox{2em}{\rowname{front}} &
		\includegraphics[width=0.186\linewidth,height=0.170\linewidth]{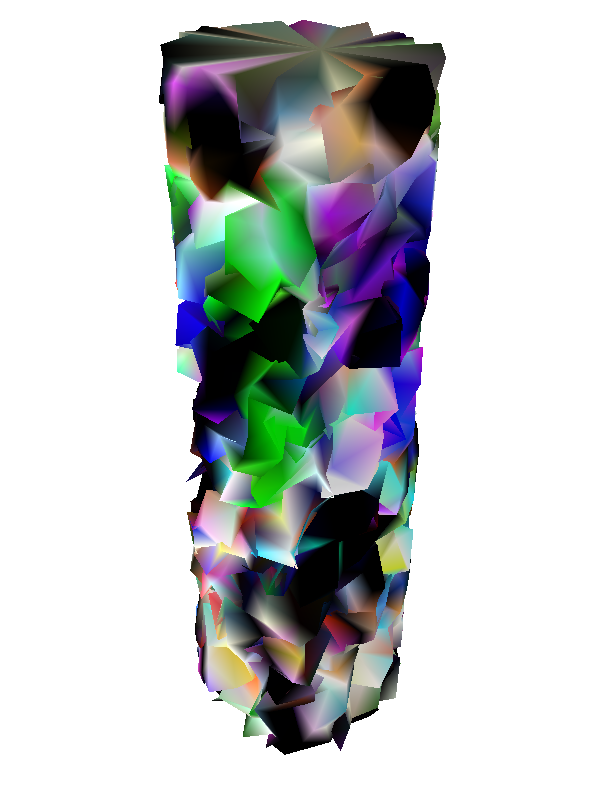} &
		\includegraphics[width=0.186\linewidth,height=0.170\linewidth]{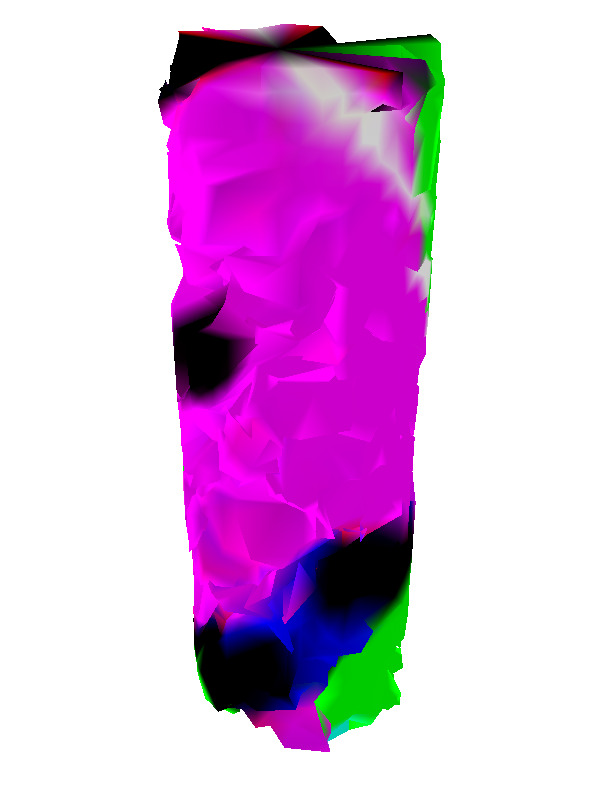} &
		\includegraphics[width=0.186\linewidth,height=0.170\linewidth]{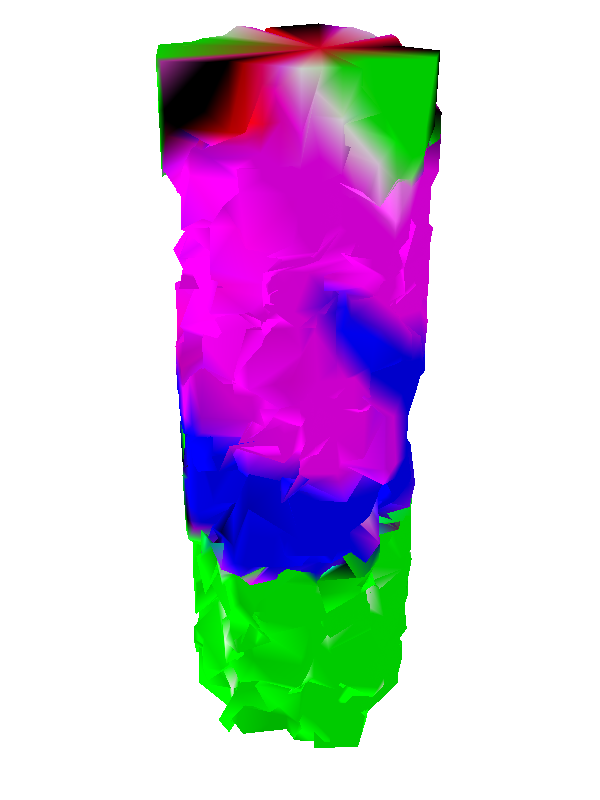} &
		\includegraphics[width=0.186\linewidth,height=0.170\linewidth]{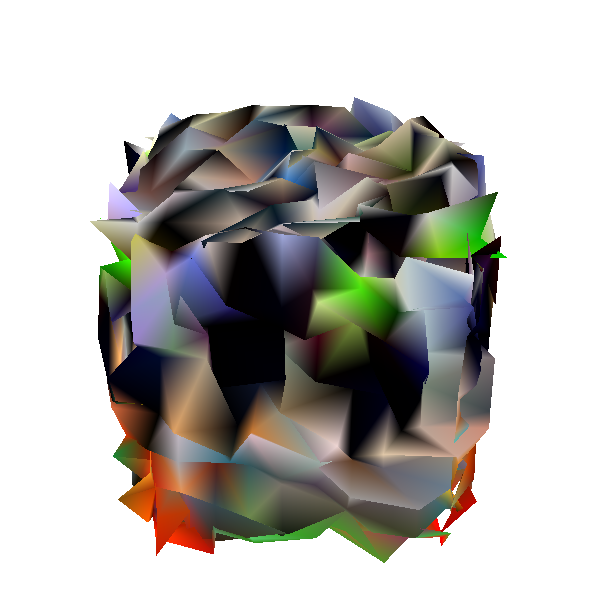} &
		\includegraphics[width=0.186\linewidth,height=0.170\linewidth]{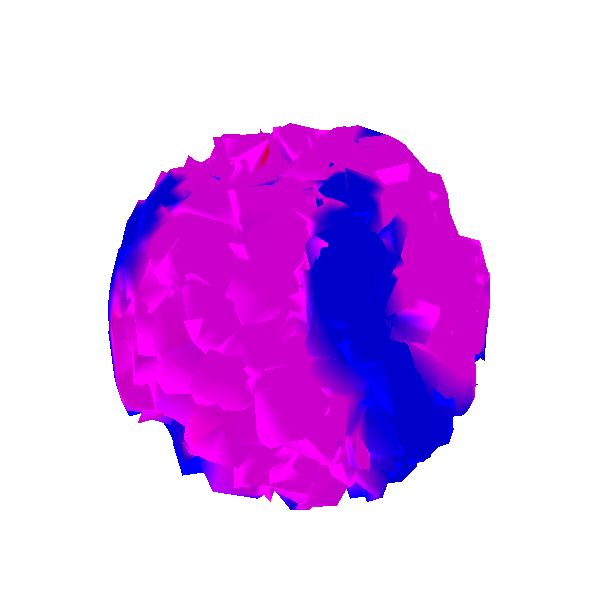} \\
		
		\raisebox{2em}{\rowname{side}} &
		\includegraphics[width=0.186\linewidth,height=0.170\linewidth]{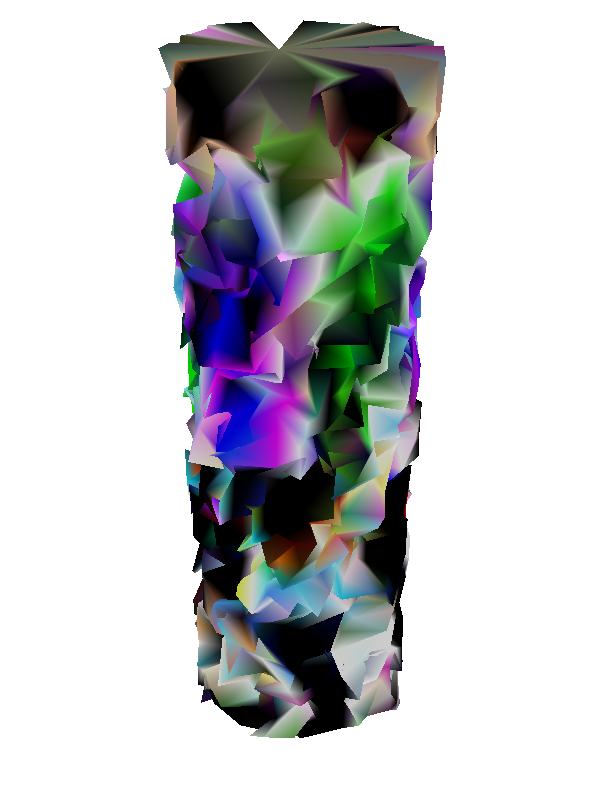} &
		\includegraphics[width=0.186\linewidth,height=0.170\linewidth]{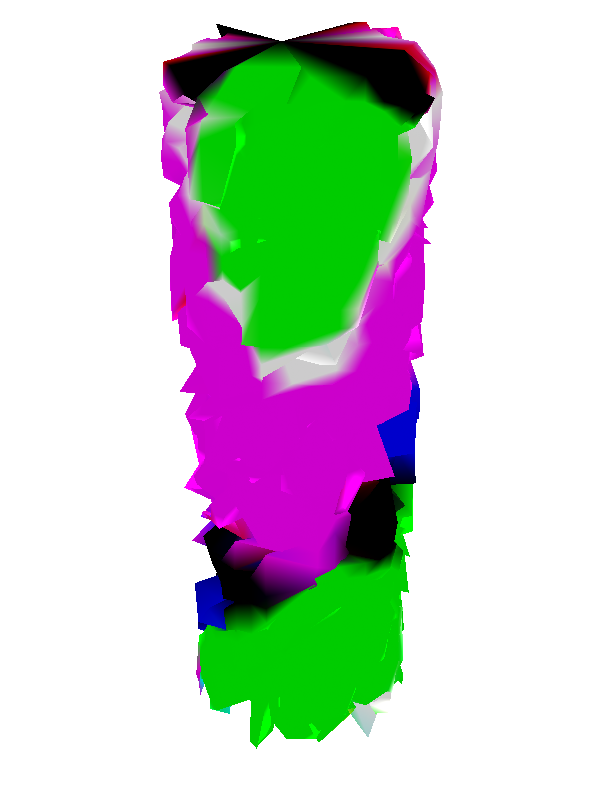} &
		\includegraphics[width=0.186\linewidth,height=0.170\linewidth]{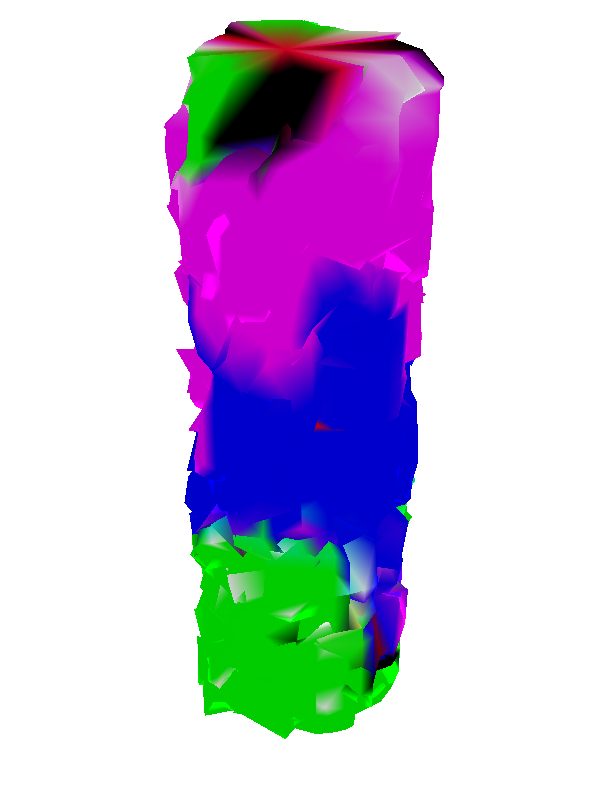} &
		\includegraphics[width=0.186\linewidth,height=0.170\linewidth]{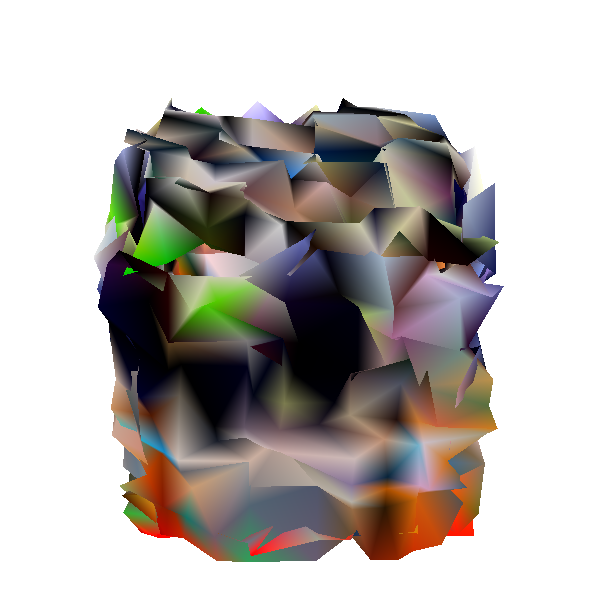} &
		\includegraphics[width=0.186\linewidth,height=0.170\linewidth]{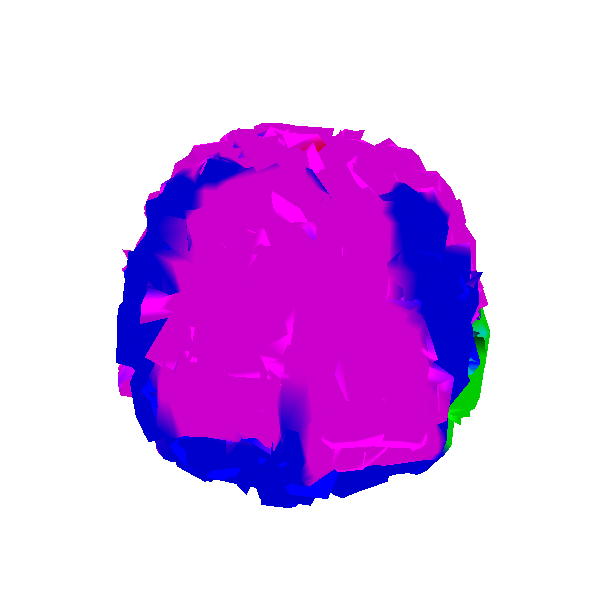} \\
		
		\raisebox{2em}{\rowname{back}} &
		\includegraphics[width=0.186\linewidth,height=0.170\linewidth]{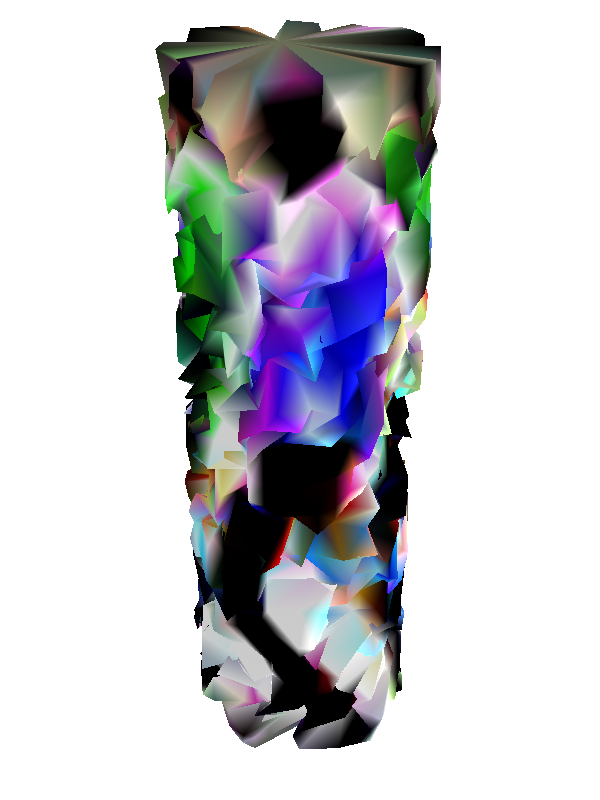} &
		\includegraphics[width=0.186\linewidth,height=0.170\linewidth]{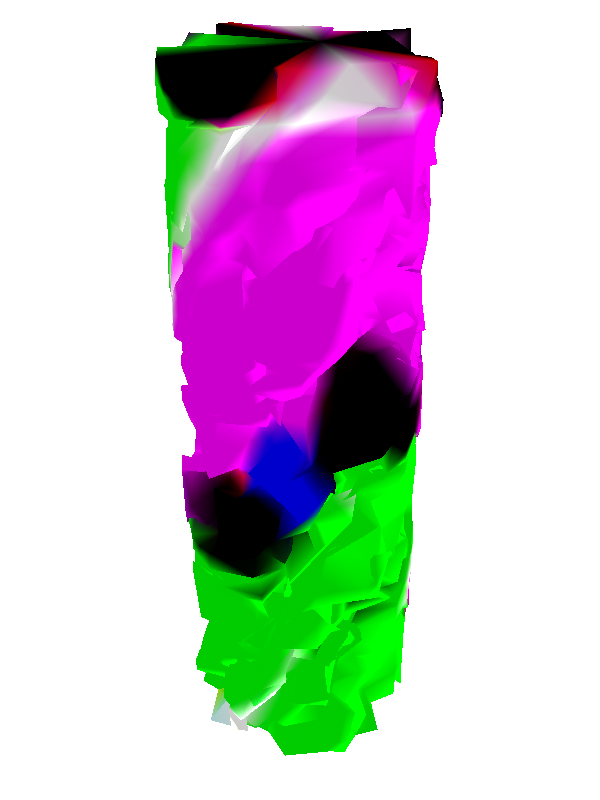} &
		\includegraphics[width=0.186\linewidth,height=0.170\linewidth]{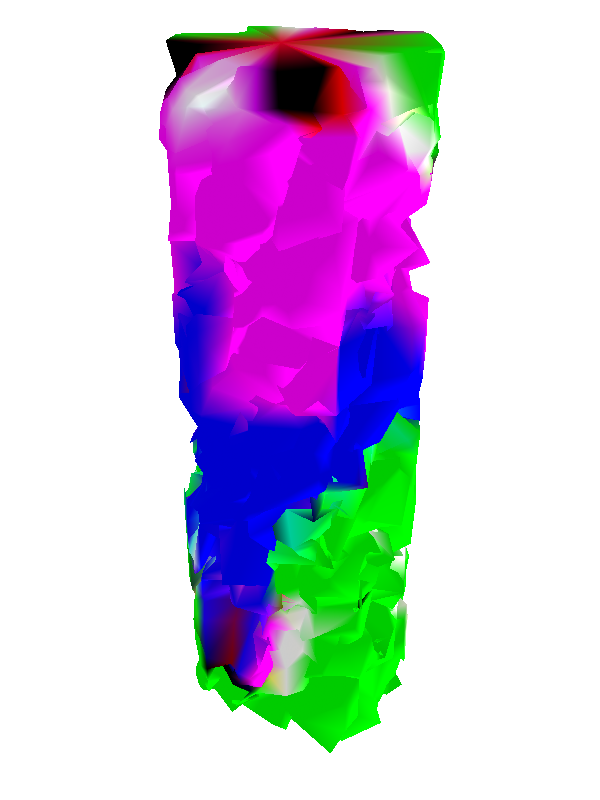} &
		\includegraphics[width=0.186\linewidth,height=0.170\linewidth]{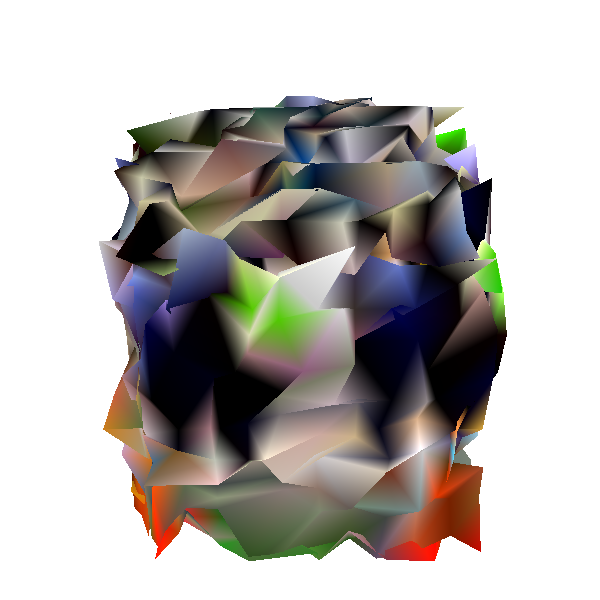} &
		\includegraphics[width=0.186\linewidth,height=0.170\linewidth]{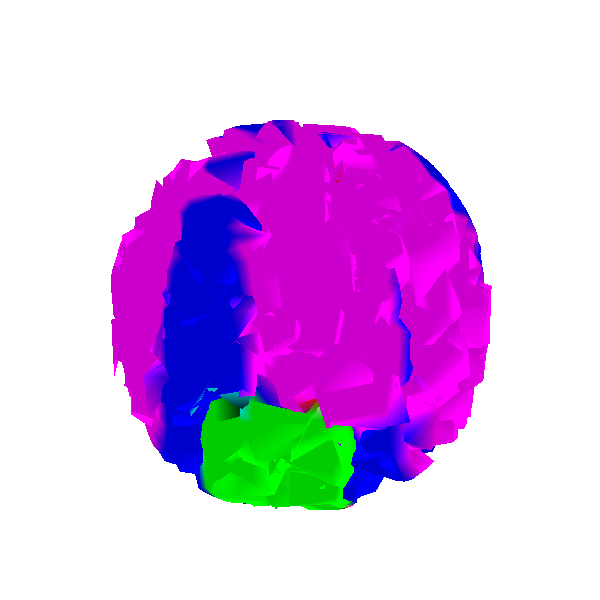} \\
		
	\end{tabular}
	\vspace{-5pt}
	\caption{\textbf{Visualization of adversarial objects}, with each row displaying the front, side, and back views. 
		The first three columns show attacks targeting different models (BEVFormer, BEV4D, and BEVDet, respectively). 
		The last three columns show attacks targeting the BEVDet model, but initialized with different geometric shapes (cylinder, cube, sphere). }
	\label{fig:combined_visuals} 
	\vspace{-15pt}
\end{figure}

\begin{figure*}[tp]
	\centering
	\small
	\begin{tabular}{{c@{ } c@{ } c@{ } c@{ } c@{ } }}
		\multirow{1}*[12mm]{\rotatebox[origin=c]{90}{Init}} & 
		{\includegraphics[width=0.21\linewidth,height=0.16\linewidth]{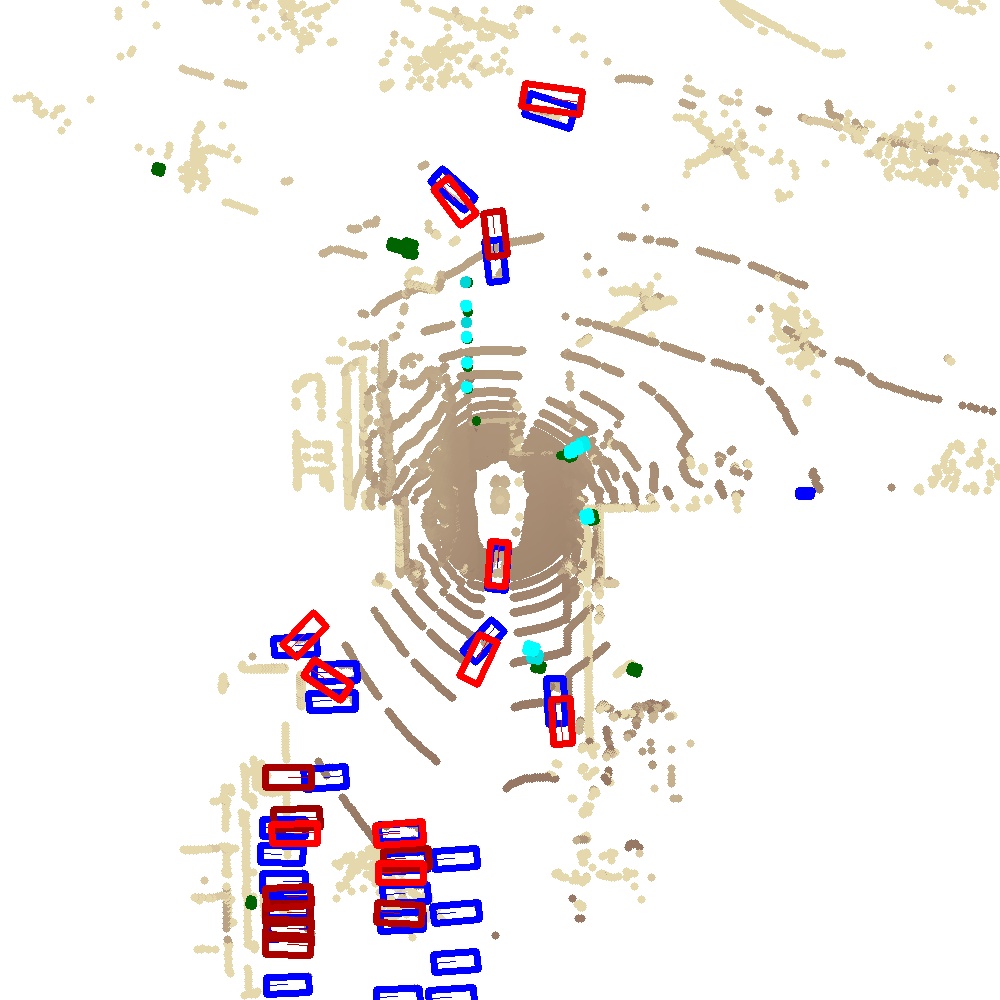}}&
		{\includegraphics[width=0.21\linewidth,height=0.16\linewidth]{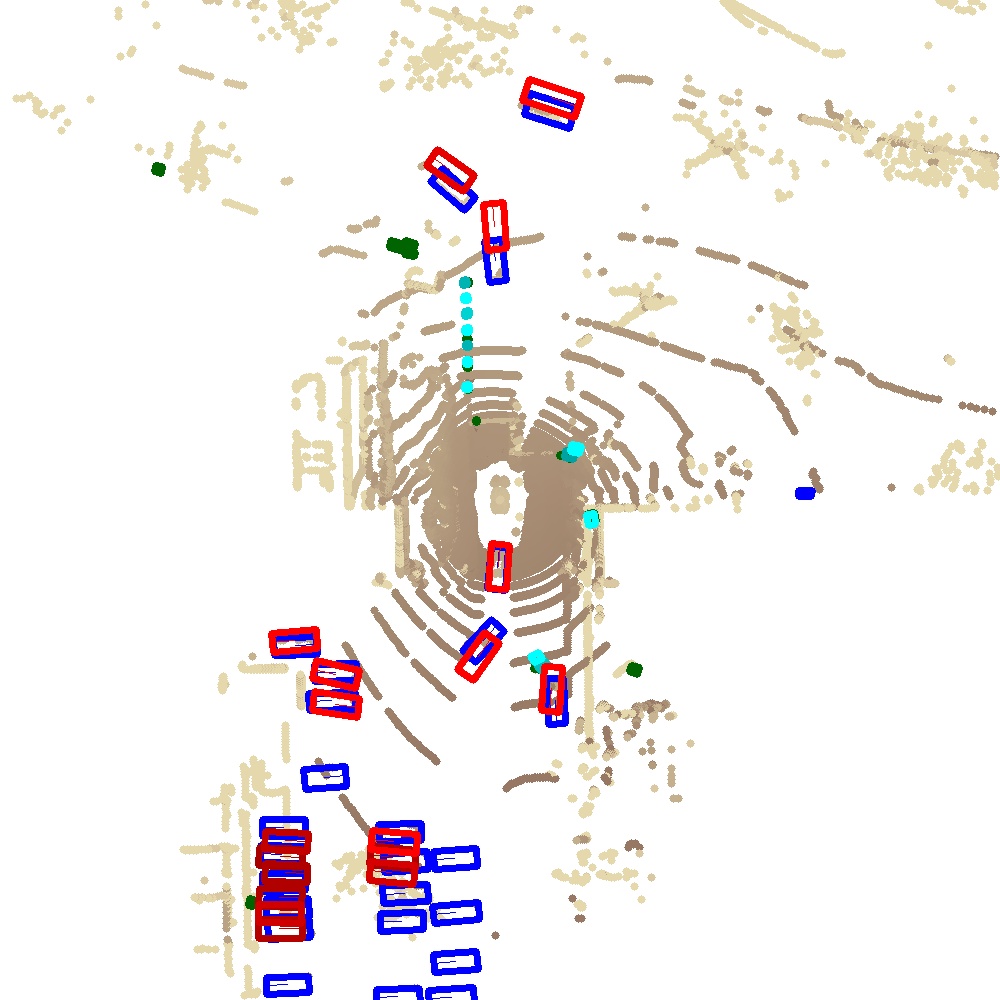}}&
		{\includegraphics[width=0.21\linewidth,height=0.16\linewidth]{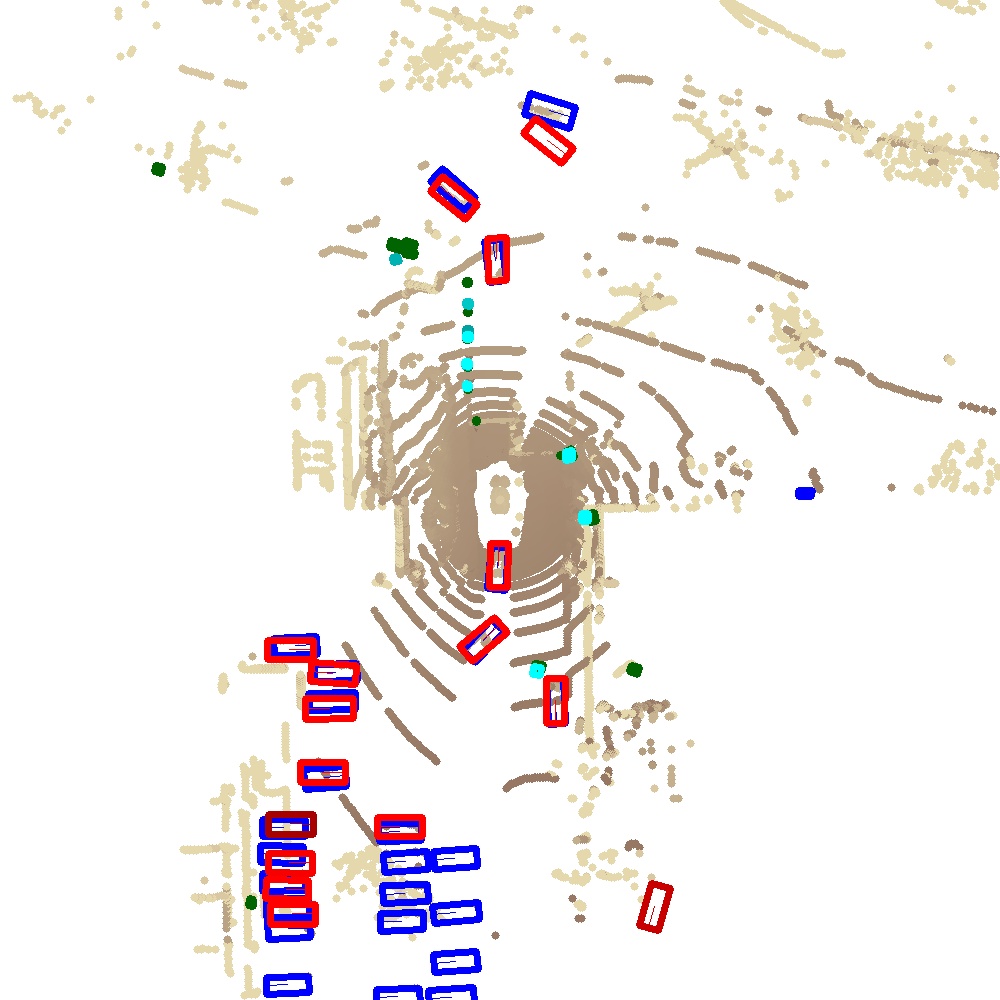}}&
		{\includegraphics[width=0.21\linewidth,height=0.16\linewidth]{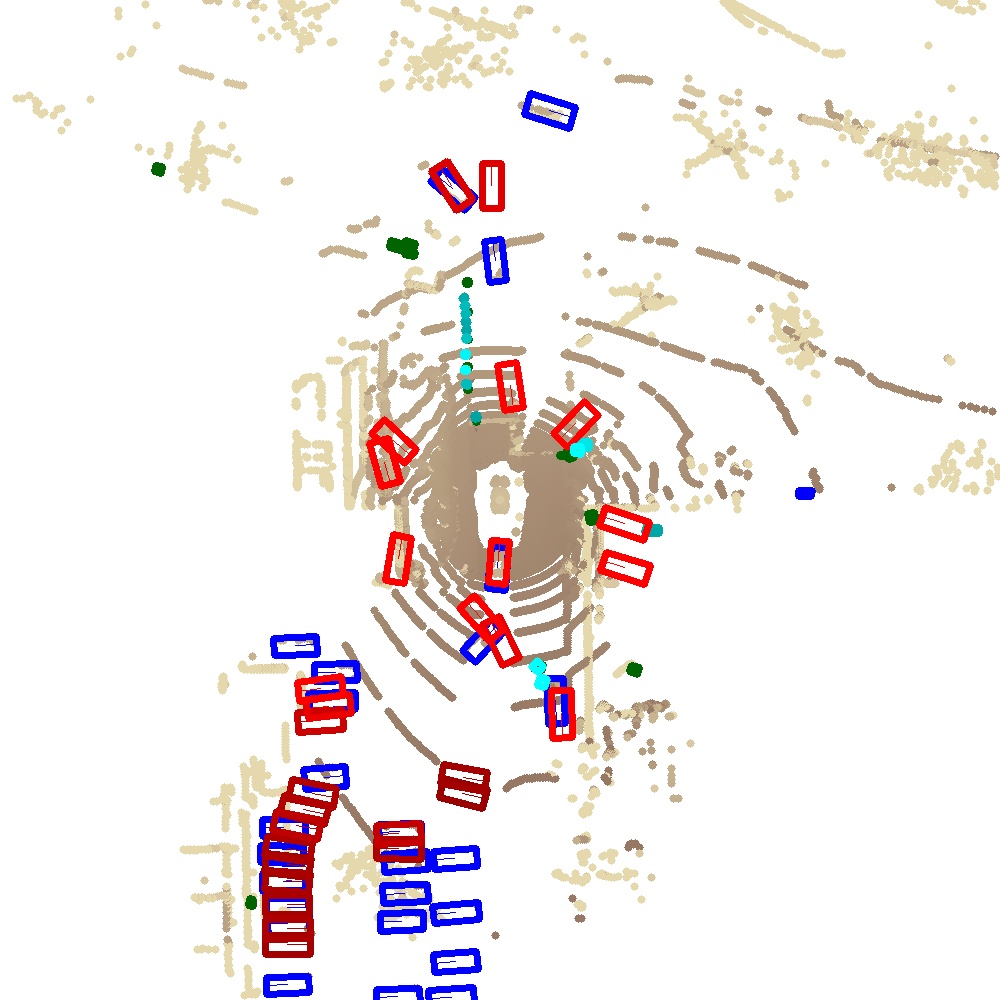}}
		
		\\
		\multirow{1}*[15mm]{\rotatebox[origin=c]{90}{Attack}} & 
		{\includegraphics[width=0.21\linewidth,height=0.16\linewidth]{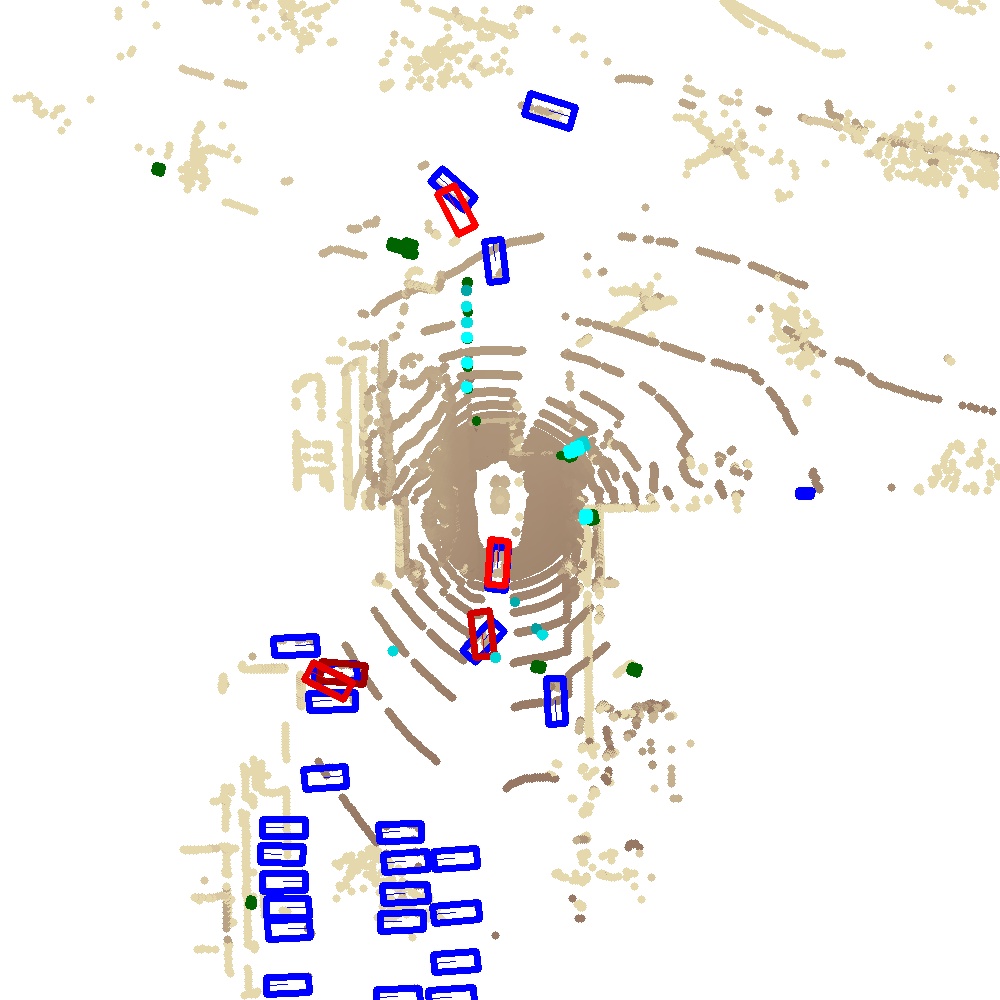}}&
		{\includegraphics[width=0.21\linewidth,height=0.16\linewidth]{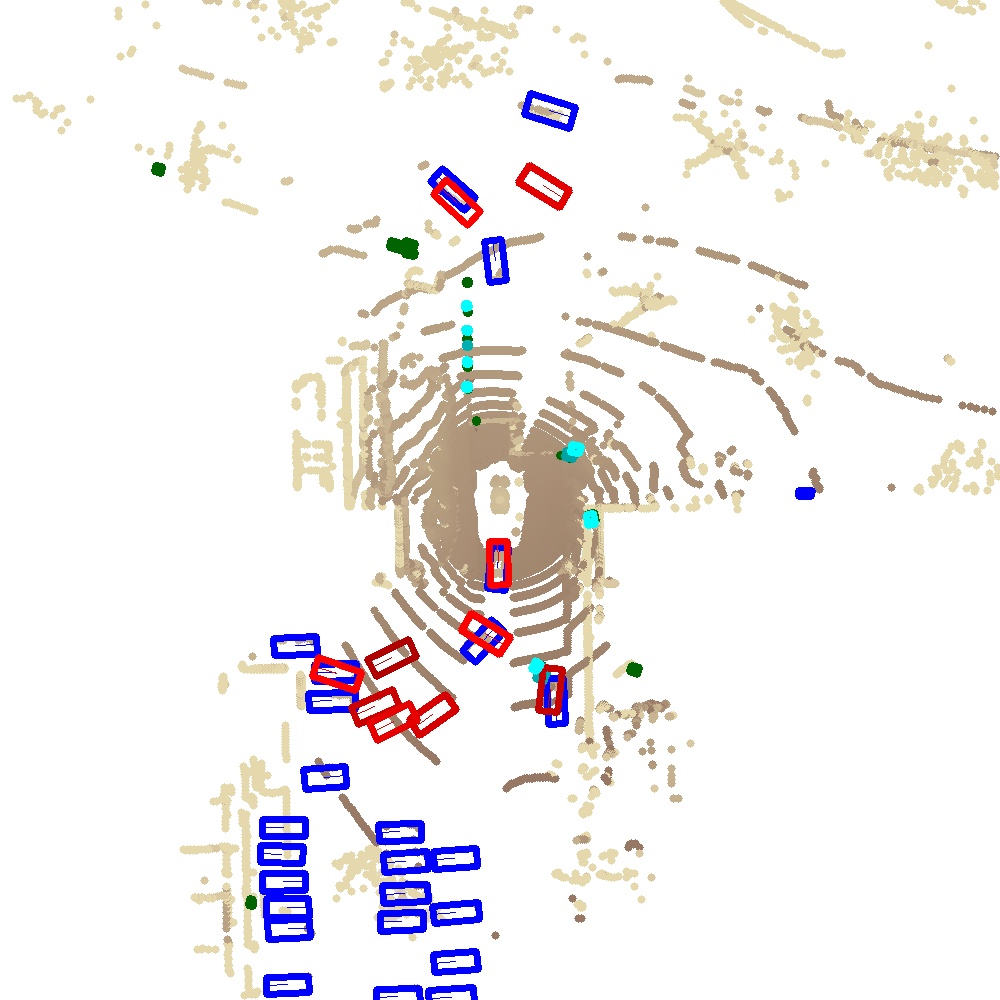}}&
		{\includegraphics[width=0.21\linewidth,height=0.16\linewidth]{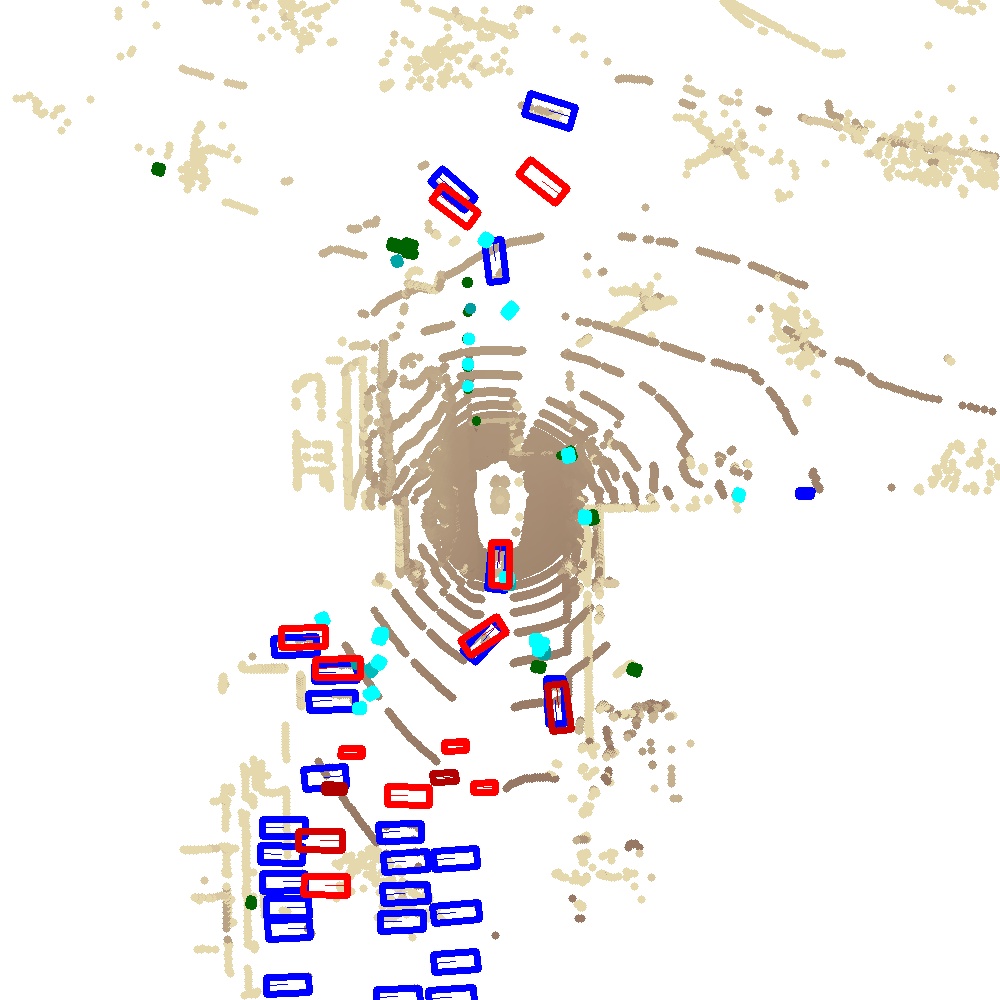}}&
		{\includegraphics[width=0.21\linewidth,height=0.16\linewidth]{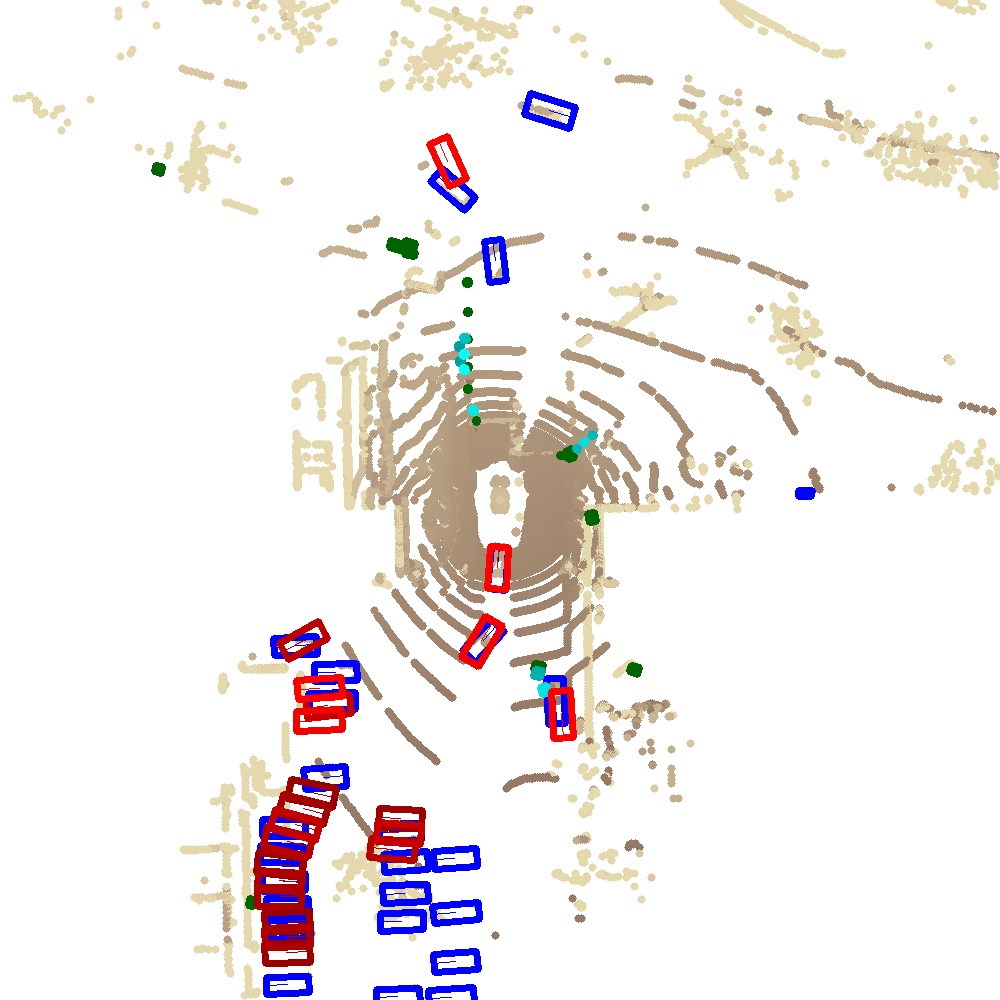}}
		
		\\
		& (a) Ours-BEVDet & (b) Ours-BEVDet4D &(c) Ours-BEVFormer  & (d)  Adv3D-BEVDet  \\
	\end{tabular} 
	\vspace{-2mm}
	\caption{\textbf{Visualizations of attack effects in the BEV.} Top: predictions with initial objects. Bottom: predictions after inserting the adversarial object. Blue/red indicate ground-truth/predicted boxes for vehicles; green/cyan for other object categories.} 
	\label{fig:bev_compare}
	\vspace{-8pt}
\end{figure*}

\cref{fig:2d_compare} and \cref{fig:bev_compare}  illustrate the effectiveness of our attack in both the 2D image and BEV views. 
Unlike the Adv3D method shown in \cref{fig:2d_compare} (d), our attack adheres to the scene's 3D-consistency and does not severely occlude the overall image. Our method can effectively suppress the model's detection of the target vehicle ((a)-(c) in \cref{fig:2d_compare}) or introduce new false positives (as shown in \cref{fig:2d_compare} (b) and (c)).
As clearly shown in the BEV view of \cref{fig:bev_compare} (d), the Init object for Adv3D's vehicle-generation attack already severely impacts the baseline prediction near the ego car. In contrast, our environment-manipulation method, which places a mesh nearby, causes 
minimal %
change to the Init prediction while achieving a more effective final attack.
This attack reveals a profound vulnerability: BEV systems, which should be robust to irrelevant environmental context, are instead over-reliant on learned priors. 
The optimized textures of the adversarial objects offer a window into how this semantic context attack exploits the model's priors.
In the first three columns of \cref{fig:combined_visuals}, we visualize adversarial objects optimized against different detectors. 
In particular, the object optimized for BEVFormer exhibits pedestrian-like textures, suggesting the model may have learned semantically incorrect associations, potentially stemming from significant dataset deficiencies.

\subsection{Ablation Studies}
\label{sec_abu}
\noindent\textbf{Initialization of 3D Objects with Different Shapes.} 
\label{exp_set2}
We evaluate how the initial mesh geometry (cylinder, cube, sphere) influences attack performance on BEVDet. The experimental settings follow those of the main experiment, with specific details provided in \cref{Implem}.
The attack results (\cref{tab:shape_compare}) and visualizations (last three columns of \cref{fig:combined_visuals}) show that when the cube and sphere have comparable volumes, the cube has a clear advantage in scene-wide attack efficacy. We attribute this to the cube's vehicle-like geometry, which facilitates generating semantically-relevant adversarial textures. The cube developed features resembling traffic cones on its vertical edges and pedestrian-like textures on its faces, which also suggests potential dataset biases.
All three shapes in \cref{tab:shape_compare} can effectively attack the detector, demonstrating that our adversarial pipeline is not limited to a specific initial shape. 
We use the cylinder mesh for our main evaluations because it simultaneously satisfies two critical requirements for practical deployment: its smooth, edgeless surface reduces the influence of lighting variations, and its profile causes minimal benign occlusion at a vehicle-comparable height.

\begin{table}[!t]
\centering
\caption{\textbf{Attack performance comparison regarding different shape} as initialization on BEVDet.}
\vspace{-3pt}
\resizebox{\linewidth}{!}{
	\LARGE
	\setlength{\tabcolsep}{2pt}  %
	\begin{tabular}{l|c|cc|cc|cc} 
		\toprule
		Shape & Real Occ & Init NDS & Adv NDS &Init mAP &Adv mAP  &$\text{ASR}_{0.3}$ &$\text{ASR}_{0.5}$\\
		\midrule
		Clean &   -   & 0.3942 & -  & 0.3086 & - & -  & - \\ \midrule
		Sphere & \xmark   & 0.3627 &  0.2402 & 0.2661 &  0.1534  & 0.565  &  0.614  \\  
		Sphere & \cmark & 0.3736 & 0.2948 & 0.2818 & 0.1958  & 0.310  &  0.367  \\  
		
		Cube & \xmark  & 0.3462 &  0.2054 & 0.2479 & 0.0881  &  0.611 &  0.679  \\  
		Cube & \cmark   & 0.3670 &   0.2574 & 0.2726& 0.1357  & 0.322  &  0.395  \\  
		
		Cylinder & \xmark  &  0.3579 & 0.2097 &0.2625 &0.1298  & 0.613  &  0.657  \\  
		Cylinder & \cmark   & 0.3682 & 0.2668& 0.2754 & 0.1597 & 0.454  &  0.515  \\  
		\bottomrule
\end{tabular}}

\label{tab:shape_compare}
\vspace{-5pt}
\end{table}

\begin{table}[!t]
\centering
	\centering
	\caption{\textbf{Performance comparison at varying distances} (in meters) from the target with Real Occ. 
	} 
	\vspace{-3pt}
	
	\resizebox{\linewidth}{!}{
		\setlength{\tabcolsep}{4pt}{
			\renewcommand{\arraystretch}{0.9}
			\begin{tabular}{c|cc|cc|cc} 
				\toprule
				Dist. & Init NDS & Adv NDS &Init mAP &Adv mAP  & $\text{ASR}_{0.3}$ & $\text{ASR}_{0.5}$ \\
				\midrule
				Clean   & 0.3942 & -  & 0.3086 & - & -  & - \\ \midrule
				0.1  &   0.3682  & 0.2668  &  0.2754  &  0.1597  & 0.454   & 0.515    \\
				0.3 &  0.3684  &  0.2629 &  0.2772 &  0.1572  &  0.457  &   0.530  \\
				0.5 & 0.3719  &  0.2710  & 0.2796  & 0.1609   &  0.444  &   0.504  \\
				0.7 & 0.3707  &  0.2698 &  0.2789 & 0.1620   &  0.518  &   0.585  \\
				1.0 &  0.3701 & 0.2760  &  0.2786 & 0.1645  &  0.444  &   0.497  \\
				\bottomrule
			\end{tabular}
	}}
	\label{tab:distance}
	\vspace{-12pt}
\end{table}

\noindent\textbf{Analysis of Training Distance.}
\label{train_dis}
We evaluate the operational range of our pipeline by training and testing the adversarial object at varying offsets from the target.
The results in \cref{tab:distance} 
indicate that our 3D-consistent, non-invasive attack uniformly reduces detection performance across these distances, demonstrating that our pipeline is not limited to a single, specific offset but generalizes within the target's vicinity, with little collateral impact on the scene.
We provide a more in-depth study of distance robustness in \cref{dis_robust}.

\noindent\textbf{Placement Generalization.} 
While our initial experiments use fixed object locations for controlled evaluation, more realistic scenario-level attacks involve objects in multiple random locations, making it hard to pre-deploy the adversarial mesh at its training position, where it may also become occluded. 
We therefore investigate placement generalization by testing whether attack effectiveness increases with the number of visible (unoccluded) meshes. 
To achieve this objective,%
we placing different numbers of guaranteed-visible meshes at random locations 
within a range of 7-27 meters
from the ego-vehicle 
to obtain less-occluded scenes and avoid tiny meshes.
The trend in \cref{tab:placement} shows that the attack performance scales consistently with the number of meshes, which demonstrates the robustness of random placement for scene-level attack. 
This highlights a practical trade-off: increasing object density boosts attack efficacy but also raises deployment cost and complexity.

\begin{table}
	\caption{\textbf{Performance comparison with varying numbers of randomly placed adversarial objects} without Real Occ.}
	\resizebox{\linewidth}{!}{
		\setlength{\tabcolsep}{4pt}{
			\renewcommand{\arraystretch}{0.92}
			\begin{tabular}{c|cc|cc|cc} 
				\toprule
				Num. & Init NDS & Adv NDS &Init mAP &Adv mAP  & $\text{ASR}_{0.3}$ & $\text{ASR}_{0.5}$ \\
				\midrule
				Clean    & 0.3942 & -  & 0.3086 & - & -  & - \\ \midrule
				1   & 0.3871  & 0.3434 & 0.2999 & 0.2433& 0.175  & 0.130  \\ 
				3   &  0.3737 & 0.2735 & 0.2811 & 0.1611  & 0.300 &  0.334 \\
				5    & 0.3599 &  0.2339  &0.2631& 0.1085 & 0.401  & 0.508 \\
				7   &  0.3432 & 0.1781&0.2429 & 0.0711& 0.590  &  0.624 \\
				10  & 0.3203 & 0.1303 &0.2147 & 0.0342& 0.793 &   0.744  \\
				\bottomrule
			\end{tabular}
	}}
	\label{tab:placement}
\vspace{-3pt}
\end{table}

\subsection{Physical Attack}
We conduct a proof-of-concept experiment to validate the physical-world feasibility of our non-invasive attack. 
To facilitate the physical validation, we simplify the attack by constraining the optimization to the mesh texture alone, leaving its geometry fixed.
We first experimentally verify the effectiveness of the adversarial mesh in the digital domain. 
After confirming that this simplification method still retained strong attack effectiveness, we create the real adversarial object physically. 
We validate our attack's effectiveness by separately placing the original mesh (a gray cylinder) and the adversarial object in the same position near the vehicle. The detailed digital validation and physical setup are provided in \cref{physical_setup}.

\begin{figure}[tp]
\centering
\small
\begin{tabular}{{c@{ } c@{ } c@{ } }}
	\multirow{1}*[12mm]{\rotatebox[origin=c]{90}{Init}} & 
	{\includegraphics[width=0.43\linewidth,height=0.23\linewidth]{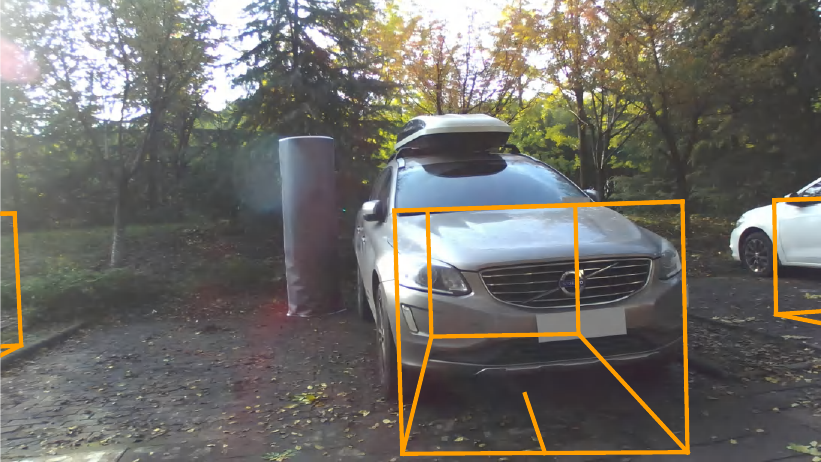}}&
	{\includegraphics[width=0.43\linewidth,height=0.23\linewidth]{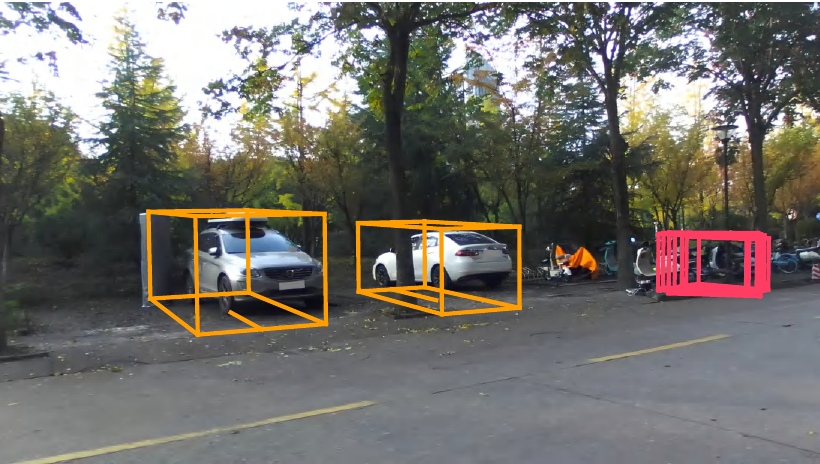}}
	\\
	\multirow{1}*[13mm]{\rotatebox[origin=c]{90}{Attack}} & 
	{\includegraphics[width=0.43\linewidth,height=0.23\linewidth]{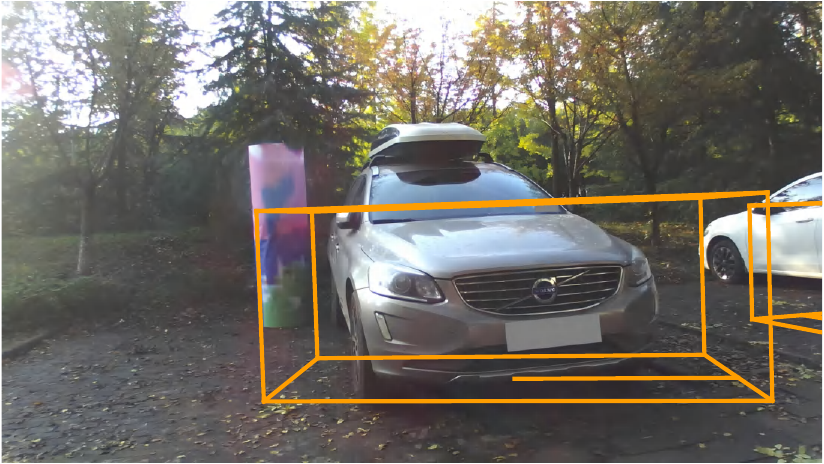}}&
	{\includegraphics[width=0.43\linewidth,height=0.23\linewidth]{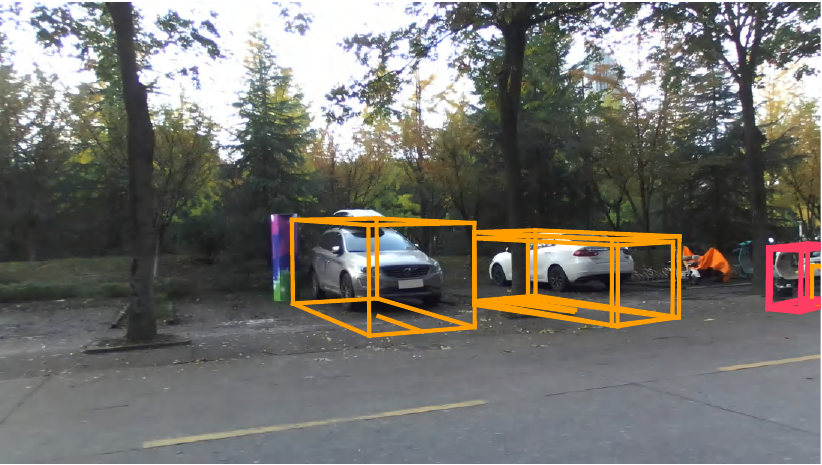}}
	\\
	& (a) Severe localization error & (b) False positive generation \\
\end{tabular} 
\vspace{-2mm}
\caption{\textbf{Visualizations of physical adversarial attack effects.} 
	Top row: predictions on the original mesh. Bottom row: predictions with our adversarial mesh. (a) Bounding box deflection: The target's predicted bounding box noticeably shifts due to the attack. (b) False Positive Generation: 
	The attack induces the model to erroneously detect an additional bounding box (white car) on the right side, predicting two vehicles instead of one. 
} 
\label{fig:physical_vis}
\vspace{-15pt}
\end{figure}

\cref{fig:physical_vis} provides a clear qualitative validation of its physical-world effectiveness. 
When a non-adversarial (gray) mesh is placed near the vehicle, the BEV detector functions correctly. However, when our adversarial mesh is placed in the same position, the detector's prediction for the nearby vehicle is successfully disrupted. This experiment confirms that our non-invasive, 3D-consistent attack framework is physically plausible and effective in real-world deployments. This also demonstrates that BEV models, which should be robust to environmental interference, are vulnerable in scenarios where our adversarial object and the vehicle co-occur.

\section{Conclusion}
In this paper, we proposed a 3D  universal adversarial object attack framework that enforces 3D spatial and inter-frame consistency. 
We leveraged scene geometry to locate candidate regions for non-invasive object insertion and employed an occlusion-aware module to ensure physically plausible occlusions. 
To further enhance attack effectiveness across space and time, we designed a BEV spatial feature-guided optimization strategy. 
Our method generates physically consistent and deployable adversarial objects, validated through digital and real-world experiments.
These results demonstrate that our environment-manipulation attack paradigm can reliably exploit contextual vulnerabilities shared across models, revealing a critical weakness in current BEV perception systems.
Our future work will explore relaxing geometric constraints to develop more flexible and physically credible attack strategies.

\section{Acknowledgments}
This research was supported in part by the National Natural Science Foundation of China (62525115, U2570208, 62271410).

{
\small
\bibliographystyle{ieeenat_fullname}
\bibliography{main}
}

\clearpage
\setcounter{page}{1}
\maketitlesupplementary

In this Supplementary Material, we first describe the implementation details in \cref{Implem} and illustrate the Realistic Occlusion checking process in \cref{sec:real_occ_check}. 
Secondly, we provide extended ablation studies in \cref{sec:extended_ablations}, covering both loss function components and the contributions of color and geometry, to further justify our design choices.
We then provide additional results on black-box transferability in \cref{black_box} and further detail the setup and digital validation of our physical attacks in \cref{physical_setup}. Moreover, we present extended analyses of the distance robustness of our attack mesh in \cref{dis_robust}, as well as results on attacking camera-only models trained with LiDAR supervision in \cref{liadr_attack}.
To examine whether our attack can be mitigated by defenses, we also evaluate its effectiveness against a robust model in \cref{sec:defence}. Finally, we provide additional metric comparisons and visualization results of our proposed non-invasive attack mesh in \cref{more_eval}.

\section{Implementation details}
\label{Implem}
\noindent\textbf{Optimization and Training Setup:}
We keep mesh topology fixed and optimize both vertex positions and texture using Adam (learning rate $0.02$), with per-vertex displacement capped at $0.1$ 
meter. 
To avoid overlapping, the offset is adjusted as the mesh size changes. 
The mesh location is determined from the current frame and kept fixed across temporal inputs. Training uses eight NVIDIA L40 48G GPUs for 10 epochs. 
The training time is 14.1h (BEVDet), 15.1h (BEVDet4D), and 57.4h (BEVFormer).
Due to memory constraints, the number of meshes per image is set to 4 for BEVDet/BEVDet4D and 2 for BEVFormer, whereas 15 meshes are used at test time. We set $\alpha$ and $\beta$ to $10$ in Eq.(9) of main text. 

\noindent\textbf{Details for Shape Initialization Experiment:}
In the experiment regarding the initialization of 3D objects with different shapes, all meshes are placed with a $0.1$m offset from the target's bottom-right corner to avoid invasive. 
The \textit{cylinder} mesh contains $1642$ vertices and $3240$ faces, with a base radius of $0.3$ and a height of $2.0$.  
The \textit{cube} mesh contains $726$ vertices and $1200$ faces, with each edge measuring $0.9$ meter.  
The \textit{sphere} mesh consists of $2562$ vertices and $5120$ faces, with a radius of $0.5$.

\section{Realistic Occlusion check}
\label{sec:real_occ_check}
\begin{figure}[tp]
   \centering
   \small
   \begin{tabular}{{c@{ } c@{ } c@{ }   }}
\multirow{3}*[-5mm]{\rotatebox[origin=c]{90}{2D check}} & 

\multicolumn{2}{c}{\includegraphics[width=0.65\linewidth,height=0.04\linewidth]{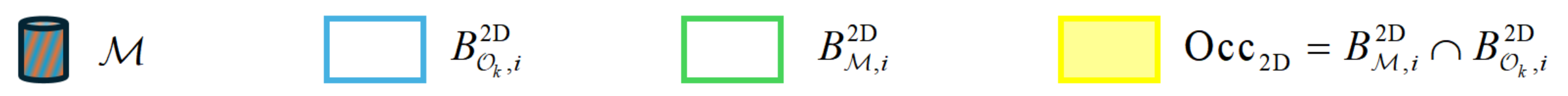}} \\
 &  {\includegraphics[height=0.21\linewidth]{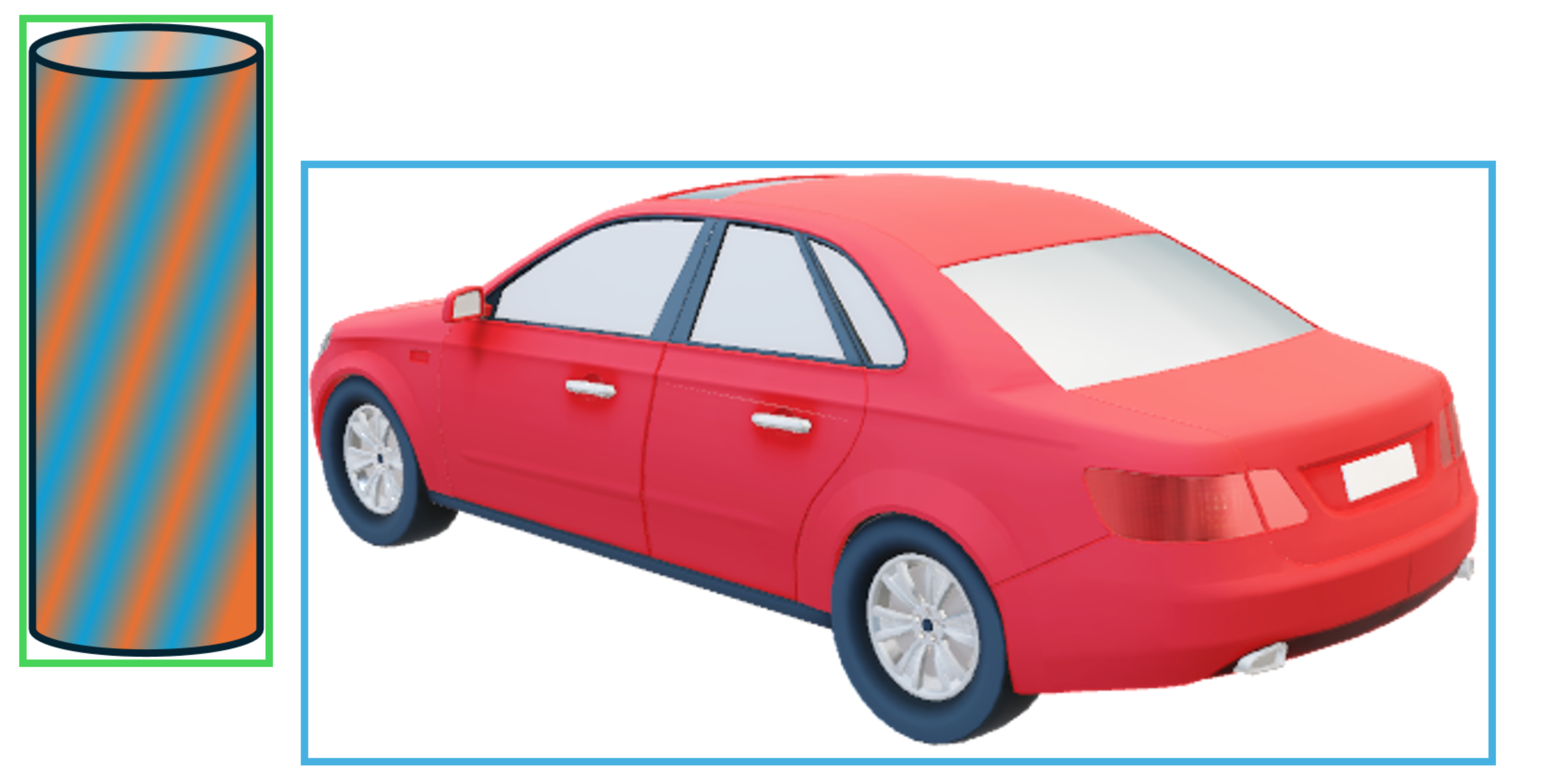}} &
  {\includegraphics[height=0.21\linewidth]{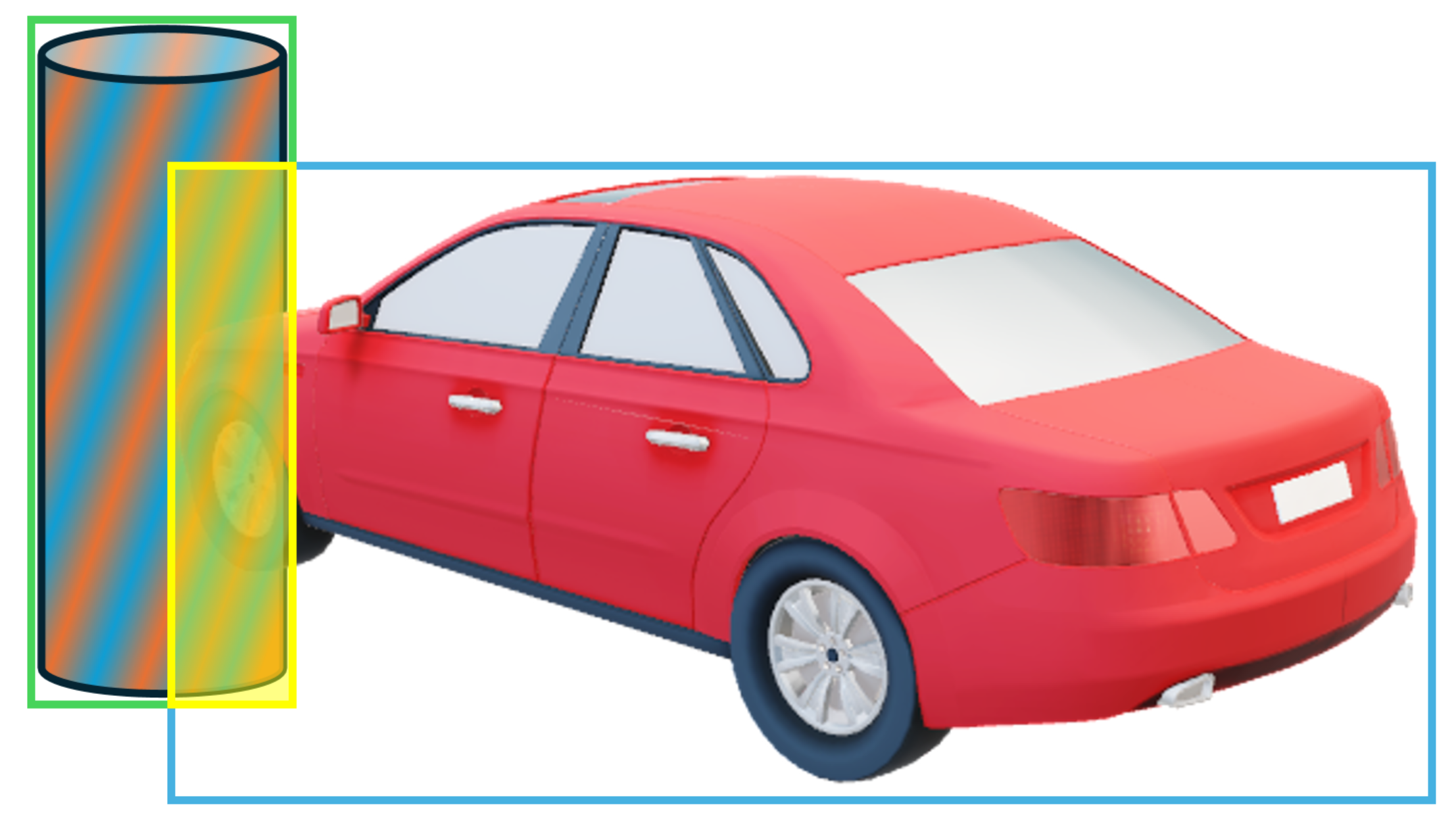}} \\ 
  & (a) \small $\text{Occ}_{\text{2D}}$  \xmark & (b)  \small $\text{Occ}_{\text{2D}}$ \cmark 
 \\
 
 \\
\multirow{3}*[-10mm]{\rotatebox[origin=c]{90}{BEV check}} & 
\multicolumn{2}{c}{\includegraphics[width=0.85\linewidth,height=0.04\linewidth]{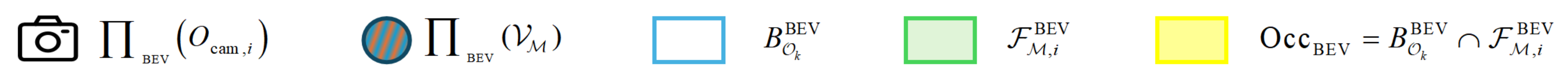}} \\
&  {\includegraphics[height=0.33\linewidth]{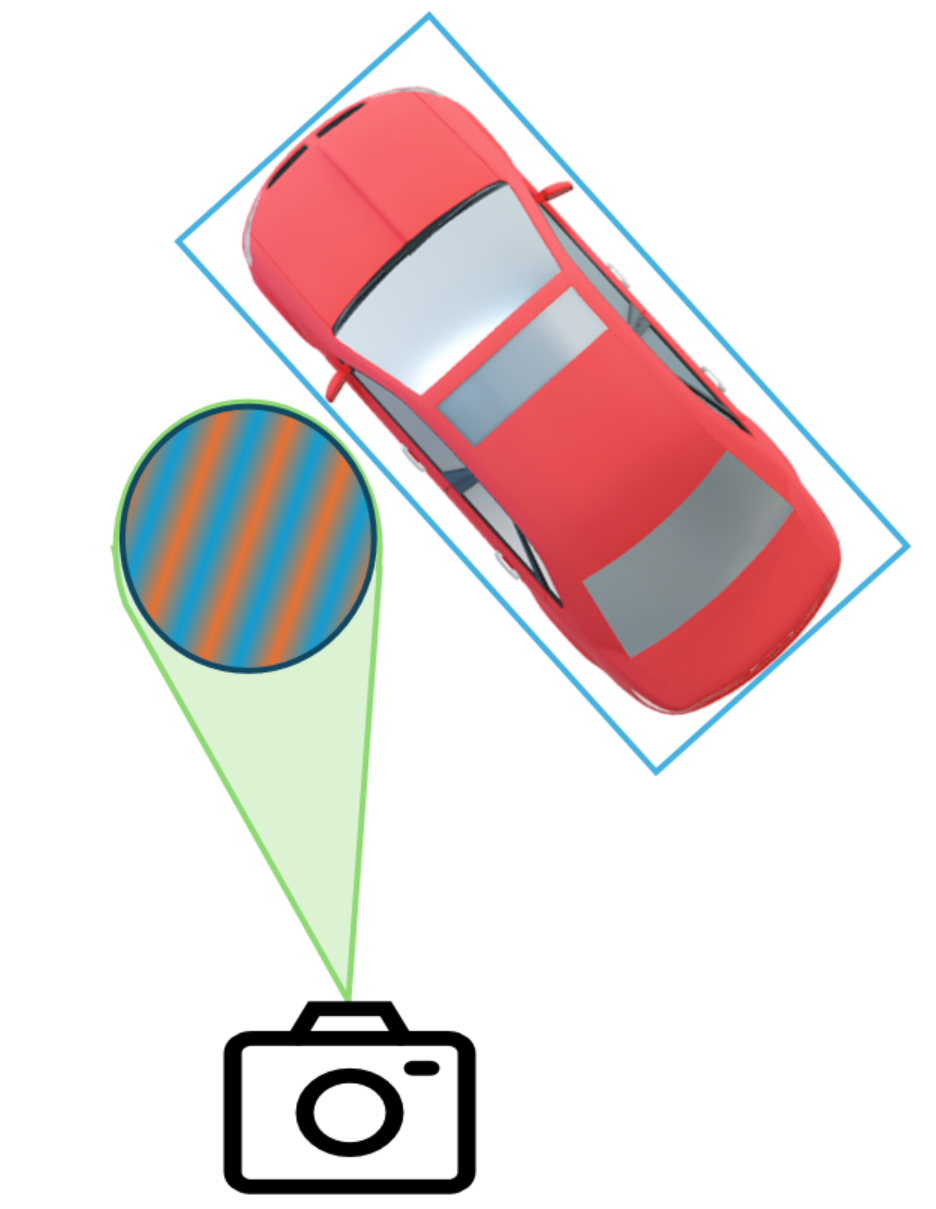}} &
  {\includegraphics[height=0.33\linewidth]{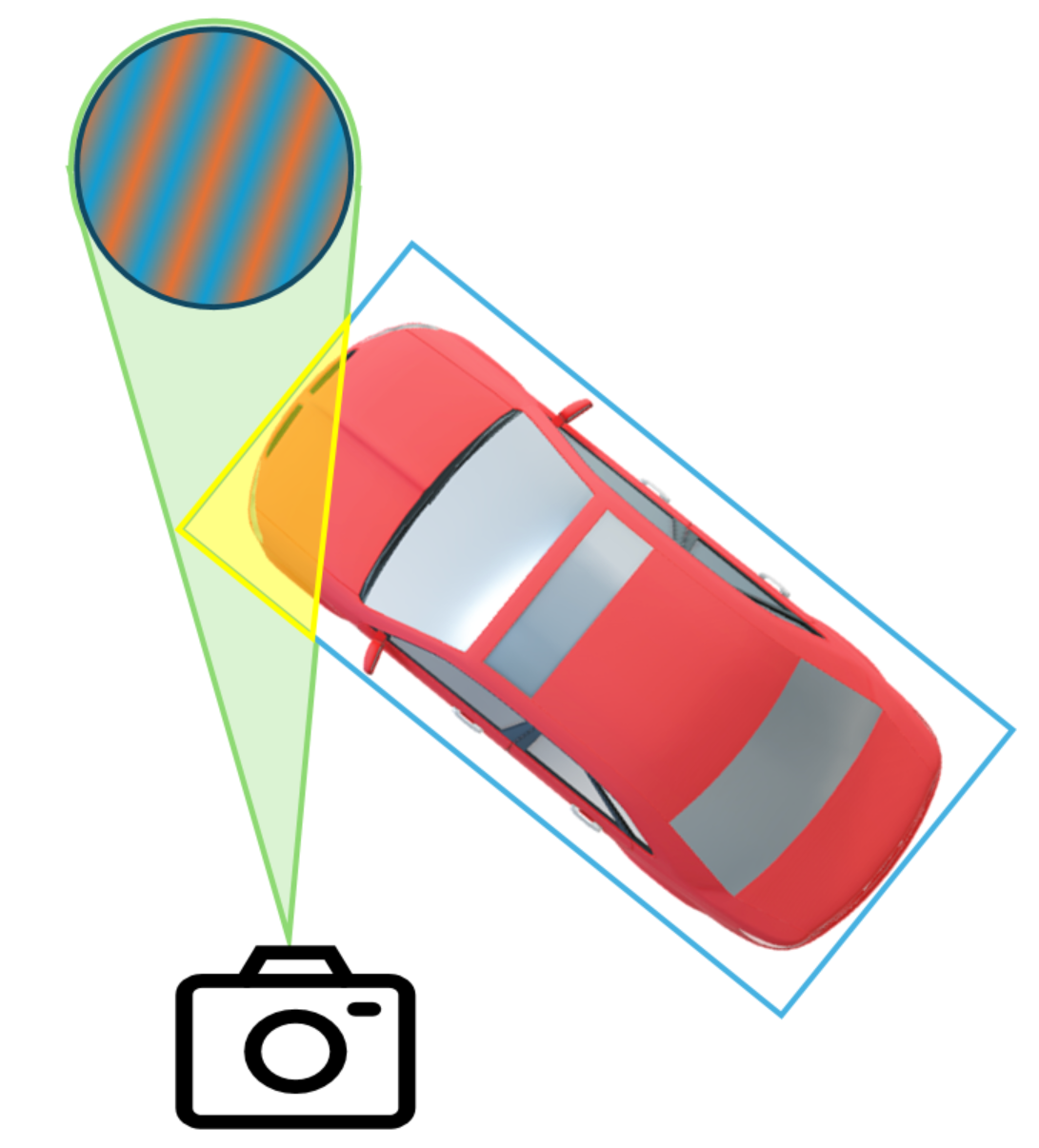}}
 \\
& (c)  \small $\text{Occ}_{\text{BEV}}$ \xmark & (d) \small $\text{Occ}_{\text{BEV}}$ \cmark 
 \\

   \end{tabular} 
    \caption{
   \textbf{Illustration of the Realistic Occlusion checking process.} The top row visualizes the occlusion check in 2D space, while the bottom row depicts the check in BEV space. The yellow regions indicate the overlapping areas. The evaluation of $\text{Occ}_{\text{BEV}}$ is triggered only when $\text{Occ}_{\text{2D}} > 0$. The adversarial mesh is considered occluded by a scene object only when both the $\text{Occ}_{\text{2D}}$ and $\text{Occ}_{\text{BEV}}$ conditions hold.
}

    \label{fig:occ_check}
\vspace{-3mm}
\end{figure}

In Sec. 3.3 of the main paper, the determination of realistic occlusion follows a two-step procedure, which involves checking the overlap of 2D bounding boxes and the overlap of convex hulls in the BEV space. To clearly illustrate this procedure, \cref{fig:occ_check} visualizes the determination scheme for $\text{Occ}_{\text{2D}}$ (Eq. (2) in the main paper) and the rationale behind $\text{Occ}_{\text{BEV}}$ (Eq. (3) in the main paper). Furthermore, \cref{fig:occ_check} (c) highlights the limitations of relying solely on depth for occlusion checking. Specifically, when the vehicle depth is smaller than that of the adversarial mesh, a naive depth-based approach would mistakenly infer that the vehicle occludes the mesh. The proposed convex-hull-based method effectively resolves this ambiguity.

\section{Extended Ablation Studies}
\label{sec:extended_ablations}
\subsection{Ablation for loss function}  
\label{loss_ablation}
To validate if $\mathcal{L}_{\text{sim}}$ effectively encourages false positives in irrelevant regions, we conduct ablation studies on different loss components in \cref{tab:loss_abu}. 
Specifically, we employ mAP for evaluation, as it is the primary metric most sensitive to false detections in nuScenes.
The further decrease in mAP upon adding $\mathcal{L}_{\text{sim}}$ confirms its effectiveness in inducing scene-level confusion and security risks, justifying its necessity.

\begin{table}[h!]
\centering
  \setlength{\tabcolsep}{1.5pt} 
  \caption{\textbf{Ablation study of loss functions on BEVDet.}}
  \vspace{-2mm}
  \label{tab:loss_abu}
    \resizebox{0.65\linewidth}{!}{
  \centering
\begin{tabular}{ccc|ccc} 
 \toprule

$\mathcal{L}_{\text{cls}}$ & $\mathcal{L}_{\text{loc}}$ & $\mathcal{L}_{\text{sim}}$ & Clean mAP & Init mAP  & Adv mAP \\
\midrule
\checkmark & &  & 0.3086 & 0.2625  & 0.1506  \\
\checkmark & \checkmark &  & 0.3086 & 0.2625  &  0.1311   \\
\checkmark & \checkmark & \checkmark & 0.3086 & 0.2625  &  0.1298 \\
\bottomrule
\end{tabular}}
\end{table}

\subsection{Ablation for Color and Geometry}
\label{sec:color_geo_ablation}
We provide additional ablation studies to compare the independent effects of color and geometry optimizations with cylinder initialization in \cref{tab:abu_goe_color}. Results show that color-based optimization is more effective, aligning with \cite{huang2024towards} in main paper that solely optimizing geometry yields limited adversarial effects.

\begin{table}[h!]
\centering
\setlength{\tabcolsep}{2pt} 
\caption{\textbf{Ablation of color and geometry optimization.} }
\vspace{-2mm}
\label{tab:abu_goe_color}
\resizebox{\linewidth}{!}{
\renewcommand{\arraystretch}{0.8}
\centering
\begin{tabular}{cc|ccc|ccc} 
\toprule
Geometry & Color& Clean NDS  & Init NDS & Adv NDS  & Clean mAP &Init mAP &Adv mAP    \\
\midrule
\checkmark &  -   & 0.3942 &  0.3579 & 0.3505 & 0.3086 & 0.2625 & 0.2519     \\
- & \checkmark  & 0.3942 &  0.3579 & 0.2136 & 0.3086 & 0.2625 & 0.1357     \\
\checkmark & \checkmark   & 0.3942 &  0.3579 & 0.2097 & 0.3086 & 0.2625 &0.1298     \\
\bottomrule
\end{tabular}}
\end{table}

\section{Results of 3D Adversarial Objects under Transfer Attacks}
\label{black_box}
To validate the cross-model generalization capability of our 3D adversarial mesh, we conduct transfer-attack evaluations on both full scenes and the ``vehicle'' category using BEVDet, BEVDet4D, and BEVFormer (\cref{tab:blackbox1} and \cref{tab:blackbox2}). The results indicate that the 3D adversarial objects consistently exploit common weaknesses across these BEV models, independent of the scenario or target category.

\begin{table}[ht!]
\centering
  \setlength{\tabcolsep}{1.5pt} 
  \caption{\textbf{Performance comparison of attack on the full scene by 3D adversarial objects under transfer attacks.}  The first row shows the source mesh model, the first column shows the victim model.}
  \label{tab:blackbox1}
    \resizebox{\linewidth}{!}{
  \centering
  
\begin{tabular}{l|cc|cc|cc|cc} 
 \toprule
 
\multirow{2}{*}{\diagbox{victim}{source}}&  \multicolumn{2}{c|}{Init}   & \multicolumn{2}{c}{BEVDet }  & \multicolumn{2}{c}{BEVDet4D}  & \multicolumn{2}{c}{BEVFormer}  \\
 & NDS & mAP & \multicolumn{2}{c|}{NDS / mAP}  & \multicolumn{2}{c|}{NDS / mAP}& \multicolumn{2}{c}{NDS / mAP} \\
\midrule
BEVDet    & 0.3579  &  0.2625 & 0.2097 & 0.1298 &  0.2306 &  0.1422  &  0.2869 & 0.1856\\
BEVDet4D   & 0.4158  & 0.2734  & 0.3229  &  0.1716 & 0.2762  &  0.1564 & 0.3309  &  0.1868 \\
BEVFormer  &  0.4592  & 0.3402  & 0.4049  & 0.2829  & 0.4039  & 0.2821  & 0.2876  &  0.1652   \\
\bottomrule
     \end{tabular}}
\vspace{-8pt}
\end{table}

\begin{table}[ht!]
\centering
  \setlength{\tabcolsep}{1pt} 
  \caption{\textbf{Performance comparison of attack on the ``Vehicle'' category by 3D adversarial objects under transfer attacks.}  The first row shows the source mesh model, and the first column shows the victim model. }
  \label{tab:blackbox2}
    \resizebox{1.0\linewidth}{!}{
     \LARGE
  \centering
 \begin{tabular}{l|ccc|ccc|ccc|ccc} 
 \toprule
 
\multirow{2}{*}{\diagbox{victim}{source}} &  \multicolumn{3}{c|}{Init}   & \multicolumn{3}{c|}{BEVDet}  & \multicolumn{3}{c|}{BEVDet4D}  & \multicolumn{3}{c}{BEVFormer}  \\
 & $\text{mAP}_{vel}$ & $\text{AP}_{car0.5}$ & $\text{AP}_{car2.0}$ & $\text{mAP}_{vel}$ & $\text{AP}_{car0.5}$ & $\text{AP}_{car2.0}$& $\text{mAP}_{vel}$ & $\text{AP}_{car0.5}$ & $\text{AP}_{car2.0}$& $\text{mAP}_{vel}$ & $\text{AP}_{car0.5}$ & $\text{AP}_{car2.0}$ \\
\midrule
BEVDet     &  0.187 &  0.1489 & 0.5835 &  0.038  &  0.0141 &  0.2048 &0.050  & 0.0342  & 0.2872 & 0.101  & 0.0629  & 0.4366  \\
BEVDet4D   &  0.195  &   0.1513 &  0.5936 & 0.068  & 0.0422 &  0.3214 &  0.051 & 0.0226  &  0.2461 &  0.095  &  0.0739 & 0.4581 \\
BEVFormer  & 0.280  & 0.1607  & 0.6890 & 0.206  &  0.1190  & 0.5627  &  0.204 &  0.1190 & 0.5578  &  0.101 & 0.0474  & 0.3590 \\
\bottomrule
     \end{tabular}}
\vspace{-8pt}
\end{table}

\section{Physical Attack Setup and Digital Validation}
\label{physical_setup}

\begin{figure}[tp]
   \centering
   \small
   \begin{tabular}{{c@{ } c@{ } }}

{\includegraphics[width=0.43\linewidth,height=0.50\linewidth]{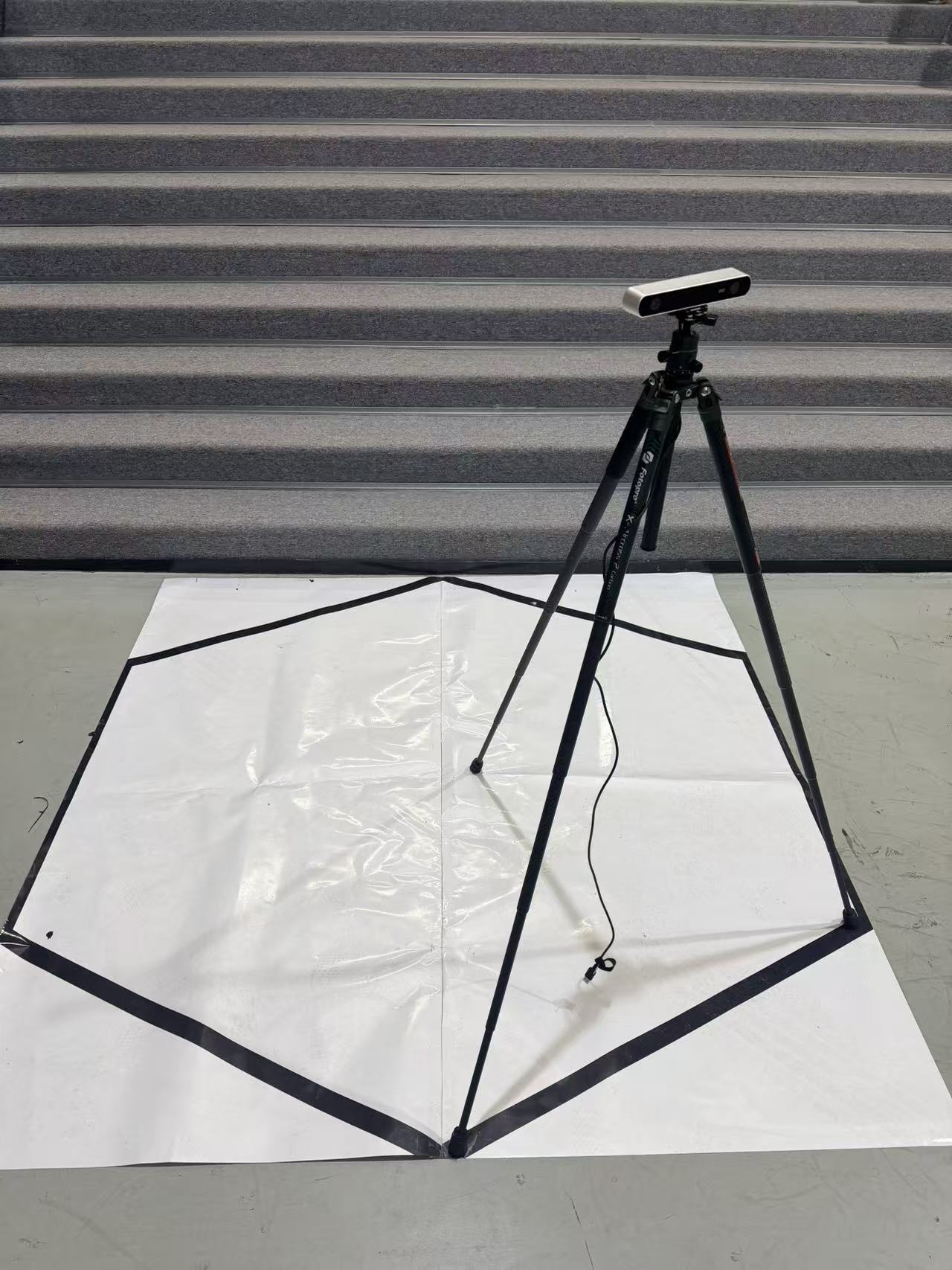}}&
  {\raisebox{.0\height}{\includegraphics[width=0.43\linewidth,keepaspectratio]{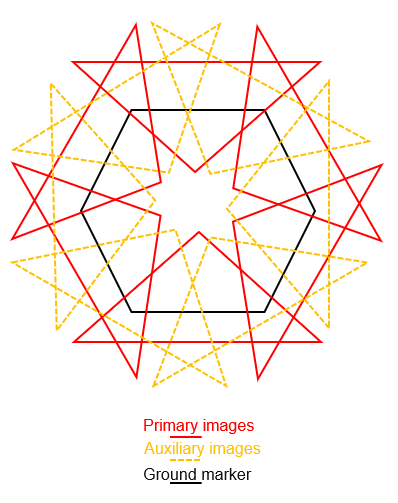}}}  
 \\

  (a) & (b) \\
   \end{tabular} 
    \caption{\textbf{Overview of our data acquisition setup.} 
    (a) Hardware setup: a camera mounted on a tripod used for image capture.(b) Diagram illustrating the 12 camera poses: 6 primary images for model input and 6 auxiliary images. The auxiliary images are captured from slightly shifted viewpoints to assist in accurate extrinsic reconstruction.} 
    \label{fig:physical_hardware}
\vspace{-8pt}
\end{figure}

To evaluate the proposed adversarial attack method under real-world physical conditions, we construct a compact physical BEV data acquisition setup.
In principle, a BEV perception system employs six synchronized cameras mounted around a vehicle to provide 360° surround perception~\cite{huang2021bevdet}.
Due to hardware limitations, we emulate this configuration using a single ZED2i stereo camera. Specifically, we capture six views sequentially by placing the camera at fixed positions corresponding to the vertices of a regular hexagon marked on the ground, each separated by 60°.
In addition, six auxiliary images are captured from slightly shifted viewpoints to assist in accurate extrinsic reconstruction, although these auxiliary images are not used for detection experiments. Our data acquisition setup and the camera pose diagram are illustrated in \cref{fig:physical_hardware}
Camera intrinsics are obtained through standard chessboard calibration~\cite{cameraCalibration}. The initial extrinsics are estimated using VGGT~\cite{wang2025vggt} from the multi-view images and subsequently refined based on the physically measured camera positions and manual inspection to recover the real-world scale.
This physical data-capture procedure effectively emulates a fixed six-camera surround-view configuration and produces geometrically consistent inputs for BEV-based 3D object detection.

To transfer a learned digital adversarial mesh into the real world by printing, we apply two pragmatic simplifications during mesh optimization. 
First, we only optimize the mesh \textbf{texture}, not the mesh geometry, which eases fabrication. 
Secondly, we adopt a \textbf{two-stage training approach}. The initial stage consists of 10 epochs under a standard digital setting. The second stage, lasting 5 epochs, introduces physically-aware rendering: we modify the rendering light based on scene weather (night, clear, rain) and add random perturbations. 
The base light intensities for night, clear, and rain are set to [0.45, 0.8, 0.5], with corresponding perturbation ranges of [0.1, 0.2, 0.15], respectively. 
During this second stage, we also apply a masked Total Variation (TV) loss \cite{mahendran2015understanding} to the mesh's texture. This loss, denoted as $\mathcal{L}_{\text{TV}}(I, M)$, is guided by the rendered image $I$ (corresponding to $I_s^{\text{rgb}}$) and the mask $M$ (corresponding to $I_s^{\text{mask}}$) from Eq.(5) in main text, and is defined as:
\begin{equation}
\label{eq:tv_loss} 
\begin{split}
    \mathcal{L}_{\text{TV}}(I, M) = & \lambda \Biggl( \frac{\sum_{i,j} \left[ (I_{i+1,j} - I_{i,j})^2 \cdot M'_{h} \right]}{\sum_{i,j} M'_{h} + \epsilon} \\
    &\quad + \frac{\sum_{i,j} \left[ (I_{i,j+1} - I_{i,j})^2 \cdot M'_{w} \right]}{\sum_{i,j} M'_{w} + \epsilon} \Biggr),
\end{split}
\end{equation}
where $M'_{h} = M_{i+1,j} \cdot M_{i,j}$ and $M'_{w} = M_{i,j+1} \cdot M_{i,j}$.

Using the calibrated multi-view captures and the print-aware mesh optimization, we perform physical experiments on driving scenes with the BEVDet model~\cite{huang2021bevdet}. 
As summarized in \cref{tab:phy1}, we first confirm in a controlled, digital setting that the simplified geometry and print-aware color modeling still produce strong adversarial effects.

\begin{figure*}[tp]
   \centering
   \footnotesize

\begin{tabular}{{c@{ } c@{ } c@{ }  c@{ } c@{ }  c@{ } }}
\multirow{1}*[10mm]{\rotatebox[origin=c]{90}{Init}} & 

  {\includegraphics[width=0.18\linewidth,height=0.11\linewidth]{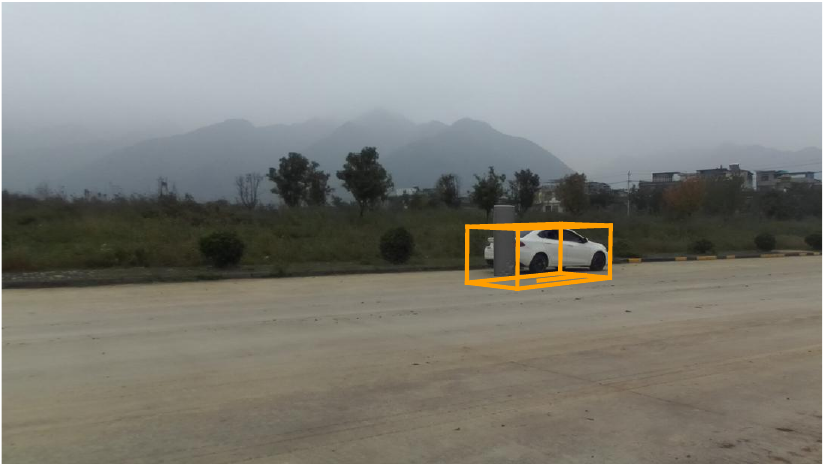}} &
   {\includegraphics[width=0.18\linewidth,height=0.11\linewidth]{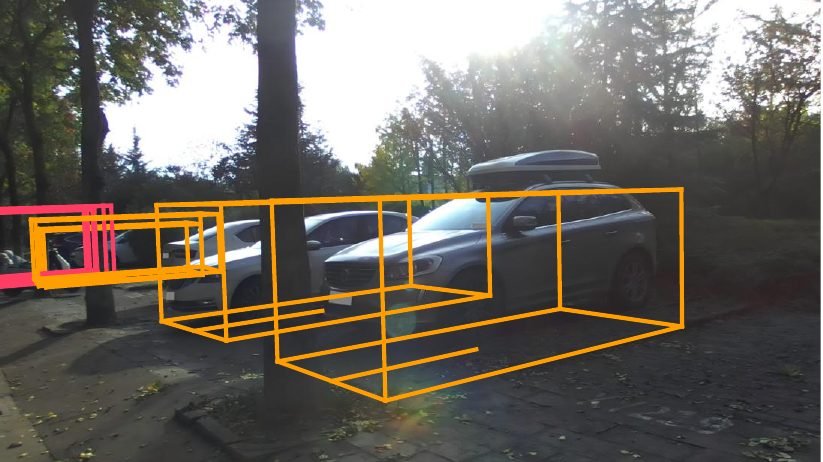}} &
  {\includegraphics[width=0.18\linewidth,height=0.11\linewidth]{fig/physical/3_grey_far.pdf}} & 
  {\includegraphics[width=0.18\linewidth,height=0.11\linewidth]{fig/physical/3_grey_far.pdf}} 
& 
  {\includegraphics[width=0.18\linewidth,height=0.11\linewidth]{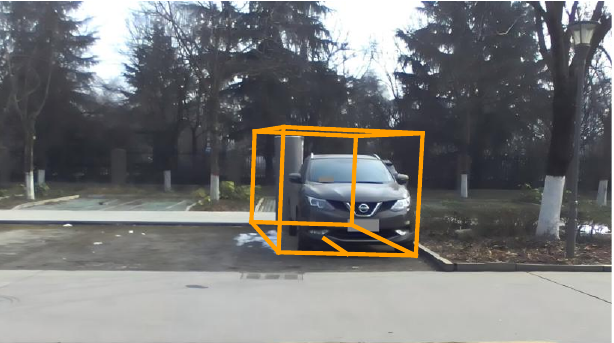}} 

 \\
 \multirow{1}*[11mm]{\rotatebox[origin=c]{90}{Attack}} & 
 {\includegraphics[width=0.18\linewidth,height=0.11\linewidth]{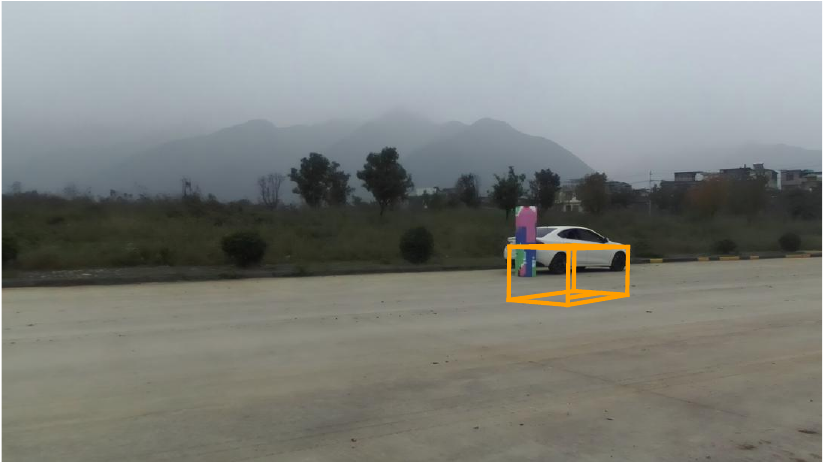}} &
   {\includegraphics[width=0.18\linewidth,height=0.11\linewidth]{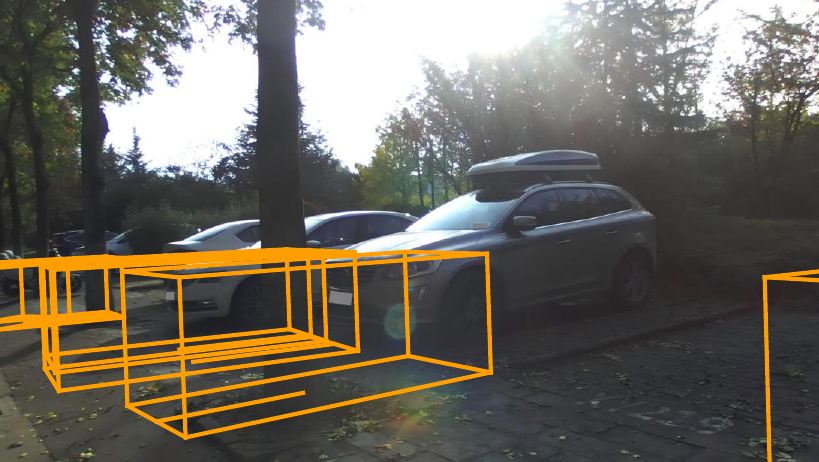}} &
   {\includegraphics[width=0.18\linewidth,height=0.11\linewidth]{fig/physical/5_newcolor_far42.pdf}}  & 
    {\includegraphics[width=0.18\linewidth,height=0.11\linewidth]{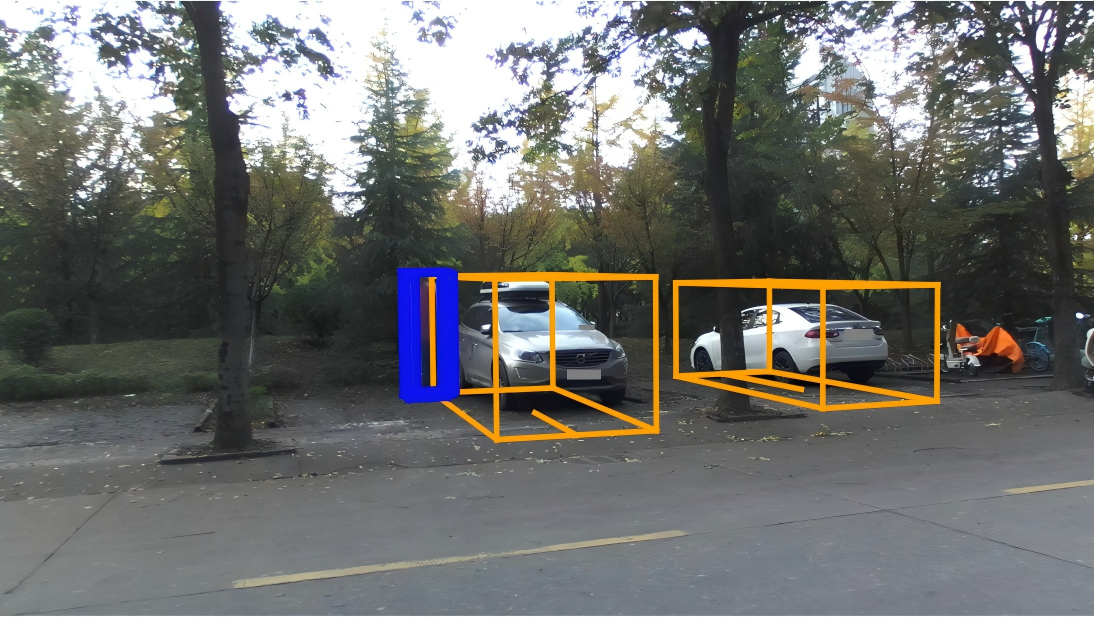}}  & 
    {\includegraphics[width=0.18\linewidth,height=0.11\linewidth]{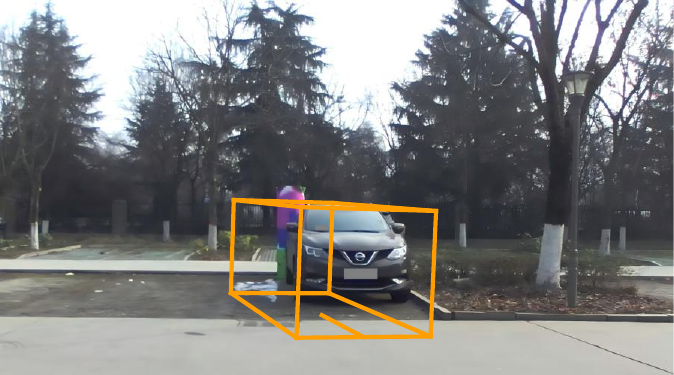}} 
\\
 & (a) Distant mis-localization & (b) Cross-view corruption  & (c) False positive generation & (d) Misclassification & (e) Occlusion Robustness  \\
   \end{tabular} 
    \caption{\textbf{Visualization of our adversarial attack's effectiveness and properties.} (a) Distant mis-localization: The attack induces significant localization errors (bounding box shifts) for vehicles at a distance. (b) Cross-view corruption: A gray/adversarial mesh is placed in the right-side camera view. This attack successfully transfers to the left-side view, where the mesh is not visible, corrupting its prediction and demonstrating the attack's cross-view capability. 
    (c) False Positive Generation: The attack induces the model to erroneously detect an additional bounding box (white car) on the right side, predicting two vehicles instead of one. 
(d) Misclassification: The model erroneously detects the barrier (physical adversarial mesh itself) as two pedestrians. 
(c) and (d) demonstrate different failure modes in the same scene and location but with varying mesh orientations.
(e) Occlusion Robustness: the attack remains effective even when the adversarial mesh is partially occluded by vehicle.
}

    \label{fig:physical_vis2}
\end{figure*}

Extending this to the real world, \cref{fig:physical_vis2} presents visual examples of our physical deployment, where the printed meshes cause consistent degradation in BEV detection across multiple scenes, viewpoints, and visibility conditions.
We observe the following primary effects of our physical attack:
(1) Localization Errors: The attack causes severe bounding box shifts (mis-localization) and, in some cases, can even lead to complete detection suppression (\ie the target vehicle's prediction box vanishes).
(2) Cross-view Corruption: The attack successfully transfers to views where the attack mesh is not directly visible, corrupting the predictions in these unobserved views.
(3) False Positives: The attack generates multiple spurious bounding boxes in the target region.
(4) Misclassification: The model predicts an incorrect category for the target object. 
(5) Occlusion Robustness: The attack remains effective even when the adversarial mesh is partially occluded by environmental obstacles (\eg vehicle).

Another interesting finding is that, under similar viewing angles, different orientations of our mesh can produce varying attack effects, as shown in \cref{fig:physical_vis2} (c) and (d). 
This also highlights the advantage of using an adversarial object, as its inherent 3D nature allows it to be consistently observed and optimized across various views and partial occlusions during training.

Through physical attack experiments, we verify that our printed adversarial mesh successfully mislead BEV detection in real-world scenes. 
This demonstrates that our approach not only works in digital simulation but also transfers effectively to physical deployment, highlighting the potential real-world security risks of BEV perception systems. 

\textbf{Physical-to-digital gap.} 
While our physical attack remains effective, we observe a performance gap compared to digital simulations. This discrepancy primarily stems from color fidelity degradation caused by the gamut misalignment between digital RGB textures and ink-printed CMYK outputs.
To mitigate this, potential future directions include: (1) establishing a differentiable color mapping model for local printers to compensate for printing distortions during the optimization process, (2) using cylindrical electronic displays instead of static ink printing. 

\begin{table}[h]
\centering
  \setlength{\tabcolsep}{1pt} 
  \caption{\textbf{Digital-domain evaluation of our physical-ready mesh against the purely digital baseline, without Real Occ.} }
  \label{tab:phy1}
    \resizebox{\linewidth}{!}{
  \centering
 \begin{tabular}{l|ccc|ccc} 
 \toprule
  & Clean NDS &  Init NDS & Adv NDS & Clean mAP &Init mAP &Adv mAP  \\
\midrule
BEVDet\_dig   & 0.3942 &  0.3579 & 0.2097 & 0.3086 &0.2625 &0.1298   \\
BEVDet\_phy   & 0.3942 &  0.3579 & 0.2139 & 0.3086 &0.2625 &0.1368   \\
\bottomrule
     \end{tabular}}
\vspace{-15pt}
\end{table}

\section{Distance Robustness Analysis}
\label{dis_robust}
\subsection{Generalization to untrained target-to-object distances}
\label{dis_robust1}
In \textbf{Attack distance} of the main text, we compare the results of training 3D adversarial objects at different locations. To further assess spatial robustness, we place a mesh trained at 0.5 units across varying distances (\cref{tab:distance_0.5}). 
The adversarial object successfully attacks the target category across all test locations, indicating robustness over distance. While the attack is more effective near the training position, ASR gradually declines as the placement moves farther away.
This also explains why the ASR in the Sec.4.4 \textbf{Placement Generalization}  of the main text is lower when the number of meshes is small: 
insufficient adversarial object coverage causes the adversarial mesh to be distant from the target vehicle, thereby compromising the attack performance.

\begin{table}[h]
  \centering
    \caption{
    \textbf{Comparison of attack performance across different adversarial object distances from the target vehicle}, using a 3D adversarial object trained at a fixed distance of 0.5 m with Real Occ.
    }
     \resizebox{\linewidth}{!}{
 \begin{tabular}{l|cc|cc|cc} 
 \toprule
Distance  & Init NDS & Adv NDS &Init mAP &Adv mAP  & $\text{ASR}_{0.3}$ & $\text{ASR}_{0.5}$\\
\midrule
Clean   & 0.3942 & -  & 0.3086 & - & -  & - \\ \midrule
0.1  &  0.3682 & 0.2682  &  0.2754 & 0.1610  &  0.452 &   0.510  \\
0.3  & 0.3684  & 0.2708  &  0.2772 &  0.1612 & 0.446  &  0.511   \\
0.5  &  0.3719 &   0.2710 &   0.2796 & 0.1609 & 0.444 &   0.511  \\
0.7  &  0.3707 &  0.2729  &  0.2789 & 0.1622  & 0.439  &  0.504  \\
1.0  & 0.3700  & 0.2782  &  0.2786 & 0.1667  &  0.448 &  0.494   \\
1.5  &  0.3739 &  0.2807  &  0.2812 & 0.1689  &  0.445 &  0.488   \\
2.0  &  0.3722  & 0.2796  & 0.2791  & 0.1711  &  0.436 &  0.489   \\
2.5  &  0.3736 &   0.2888 & 0.2819  &  0.1778 & 0.422  &  0.473   \\
3.0  & 0.3749  &  0.2944  & 0.2855  & 0.1839  & 0.404  &  0.462   \\
\bottomrule
     \end{tabular}}
\vspace{-3mm}
\label{tab:distance_0.5}
\end{table}

\subsection{Attacks on vehicles at varying target-to-ego distances}
To investigate the attack performance relative to the distance between the target and ego vehicles, we stratified the evaluation results into discrete distance intervals. As detailed in \cref{tab:distance_robust}, our proposed universal mesh attack maintains a high Attack Success Rate against the BEVDet model across all evaluated ranges, from near to far. 
This result demonstrates that our method exhibits robust and consistent efficacy across diverse distance scales.

\begin{table}[h]
\centering
\caption{\textbf{Comparison of attack performance at varying distances between the target vehicle and the ego vehicle,} using $\text{ASR}_{0.3}$ as the evaluation metric.}
\resizebox{0.6\linewidth}{!}{
\begin{tabular}{l|ccc} 
\toprule
 & 0-20 m &  20-40 m & 40-60 m \\
\midrule
Ours   & 0.366 & 0.781 & 0.929 \\ \midrule
Adv3D & 0.300 & 0.490 & 0.536  \\
\bottomrule
\end{tabular}
}
\vspace{-3mm}
\label{tab:distance_robust}
\end{table}

\section{Attacking Camera-Only Models Trained with LiDAR Supervision}
\label{liadr_attack}

While our analysis in the main text primarily focuses on attacks against purely camera-based BEV models, some existing methods \cite{zhang2025geobev,wang2025hybridbev, li2023bevdepth, li2024bevnext} utilize LiDAR information as supervision during the training phase while operating as camera-only models at inference. We selected \cite{zhang2025geobev} as a representative method from this category to validate the effectiveness of our attack, as shown in \cref{tab:geobev}.
We observe a more pronounced decline in mAP compared to NDS. This suggests that our attack method is more adept at inducing missed detections rather than degrading the attribute predictions, such as position, size, and orientation, for True Positives.
 This is further corroborated by the high Attack Success Rate (ASR) maintained across various IoU thresholds.

\begin{table}[h]
\centering
  \setlength{\tabcolsep}{1pt} 
  \caption{\textbf{Performance comparison of our adversarial attack on GeoBEV \cite{zhang2025geobev}}, a representative camera-only model trained with LiDAR supervision.}
  \label{tab:geobev}
    \resizebox{0.98\linewidth}{!}{
  \centering
 \begin{tabular}{l|c|ccc|ccc|ccc} 
 \toprule
& Real Occ & Clean NDS &  Init NDS & Adv NDS & Clean mAP &Init mAP &Adv mAP    & $\text{ASR}_{0.3}$ & $\text{ASR}_{0.5}$ &  $\text{ASR}_{0.7}$  \\
\midrule
GeoBEV~\cite{huang2021bevdet} &   \xmark  & 0.5459 &  0.5058 &  0.3493 & 0.4296 &0.3732 & 0.1640   &  0.704   &   0.767 & 0.808   \\
GeoBEV~\cite{huang2021bevdet} & \cmark  & 0.5459 &   0.5174 & 0.4068 & 0.4296 & 0.3906 &  0.2259 &  0.447    &   0.512 &  0.579  \\

\bottomrule
     \end{tabular}}
\vspace{-4mm}

\end{table}

\section{Attack Against Robust Model}
\label{sec:defence}
With the efficacy of our method on standard models previously established, we proceed to investigate whether our attack remains effective against models fortified with defense strategies.
To this end, we performed experiments on BEVDet by incorporating adversarial training to train a robust model. Specifically, we implemented data augmentation by replacing 10\% of the total training data with adversarial examples generated via Projected Gradient Descent (PGD) \cite{madry2017towards}. We then fine-tuned the standard BEVDet model on this augmented dataset for two epochs to obtain the robust BEVDet model. Subsequently, following the same pipeline as described in the main text, we inserted the adversarial object optimized on the standard BEVDet model into the scenes to evaluate the performance of this robust model. For the PGD configuration, we set the step size to $1/255$, the maximum perturbation budget to $8/255$, and the number of steps to $20$.  As shown in \cref{tab:attack_robust}, despite the defense strategies, the robust model experiences a substantial decline in both NDS and mAP, accompanied by a high ASR. This demonstrates that our proposed method is capable of bypassing simple defense measures.

\begin{table}[h]
  \centering
        \caption{\textbf{Evaluating attack effectiveness against a defense-enhanced (robust) BEVDet model.} ``Clean'' refers to the baseline performance on the test set, while   ``Ours'' reports the results of attacking this robust model using our adversarial mesh optimized on the standard (non-robust) BEVDet model. 
        }  
     \resizebox{\linewidth}{!}{
     \setlength{\tabcolsep}{2pt}
     \LARGE
 \begin{tabular}{c|c|cc|cc|cc} 
 \toprule
& Real Occ  & Init NDS & Adv NDS &Init mAP &Adv mAP  & $\text{ASR}_{0.3}$ & $\text{ASR}_{0.5}$  \\
\midrule
Clean&    &  0.3779 & -  &  0.2835 & - & -  & - \\ \midrule
Ours & \xmark &  0.3418 &  0.2339  & 0.2372 & 0.1472    & 0.511 &  0.547  \\
Ours & \cmark &  0.3518  &  0.2791  &  0.2514  &   0.1764   & 0.346 &  0.380  \\
\bottomrule
     \end{tabular}}
\vspace{-3mm}

\label{tab:attack_robust}
\end{table}

\section{Additional Results for White-Box Attacks}
\label{more_eval}
\begin{figure*}[h]
   \centering
   \small
   \begin{tabular}{{c@{ } c@{ } c@{ } c@{ } c@{ }c@{ }  }}
 {\includegraphics[width=0.225\linewidth, height=0.13\linewidth]{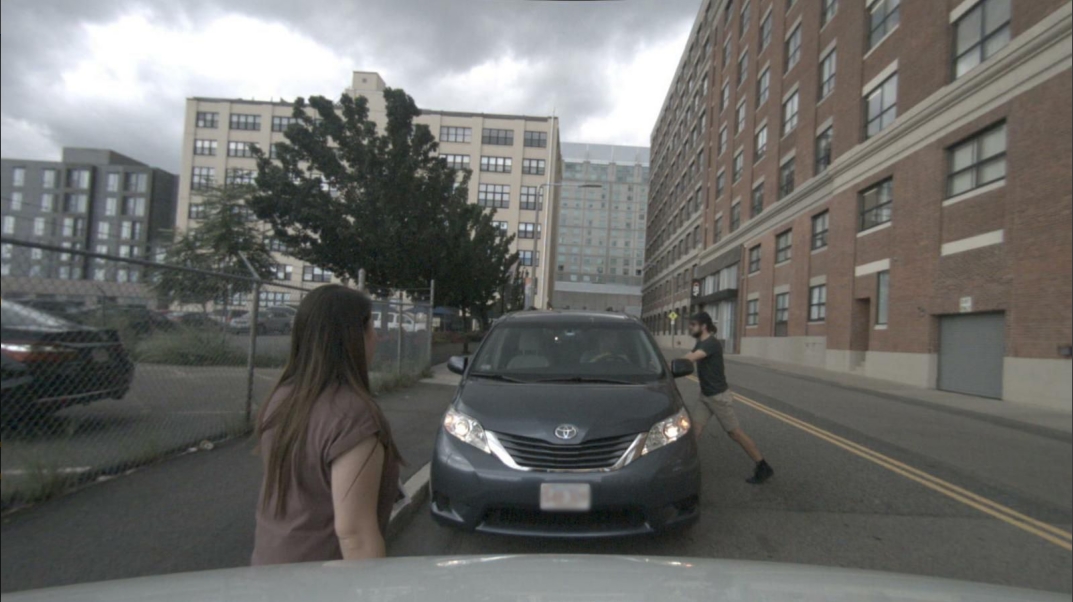}}&
 \multirow{1}*[11mm]{\rotatebox[origin=c]{90}{Init}} & 
 {\includegraphics[width=0.225\linewidth, height=0.13\linewidth]{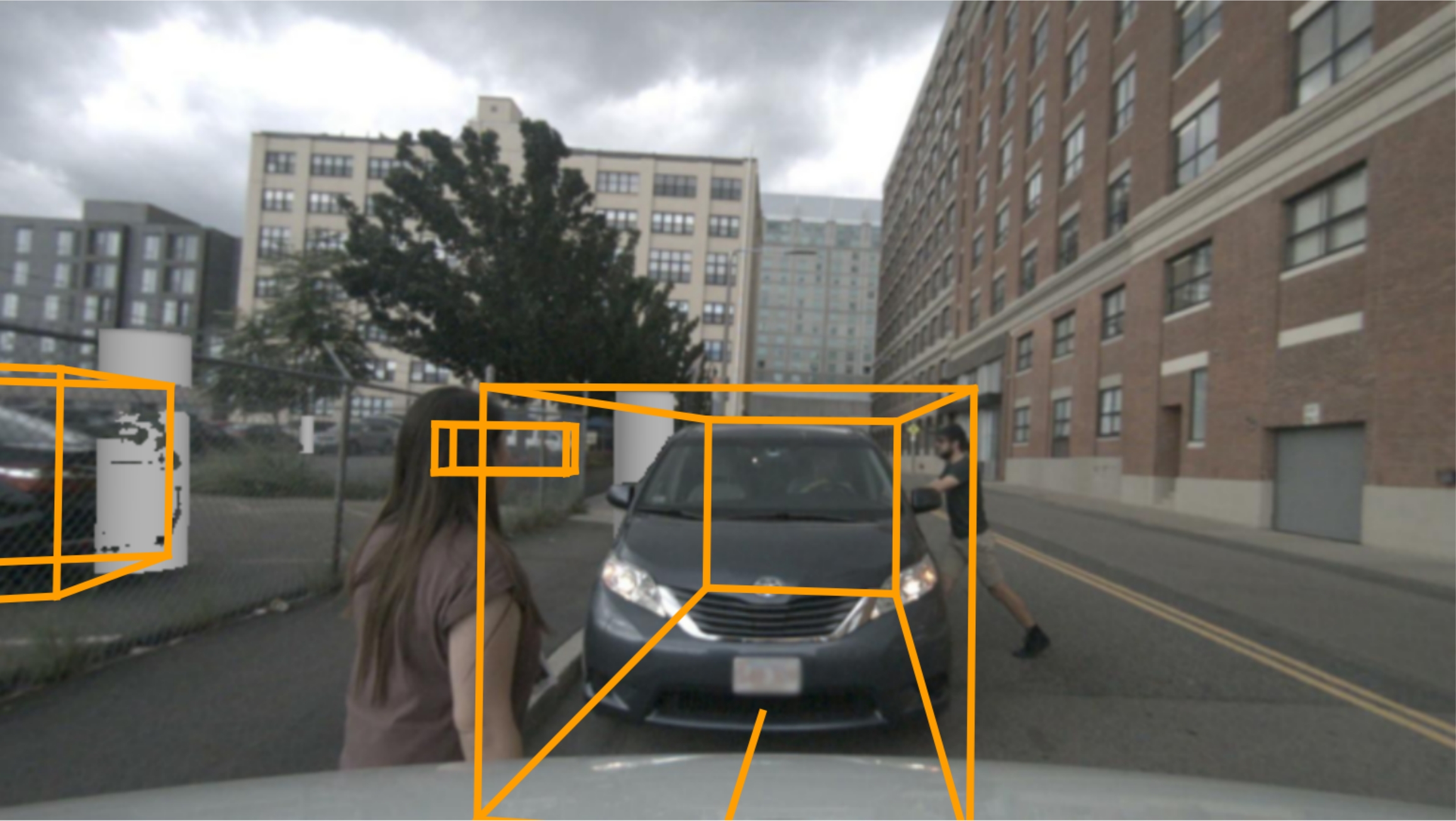}}&
 {\includegraphics[width=0.225\linewidth, height=0.13\linewidth]{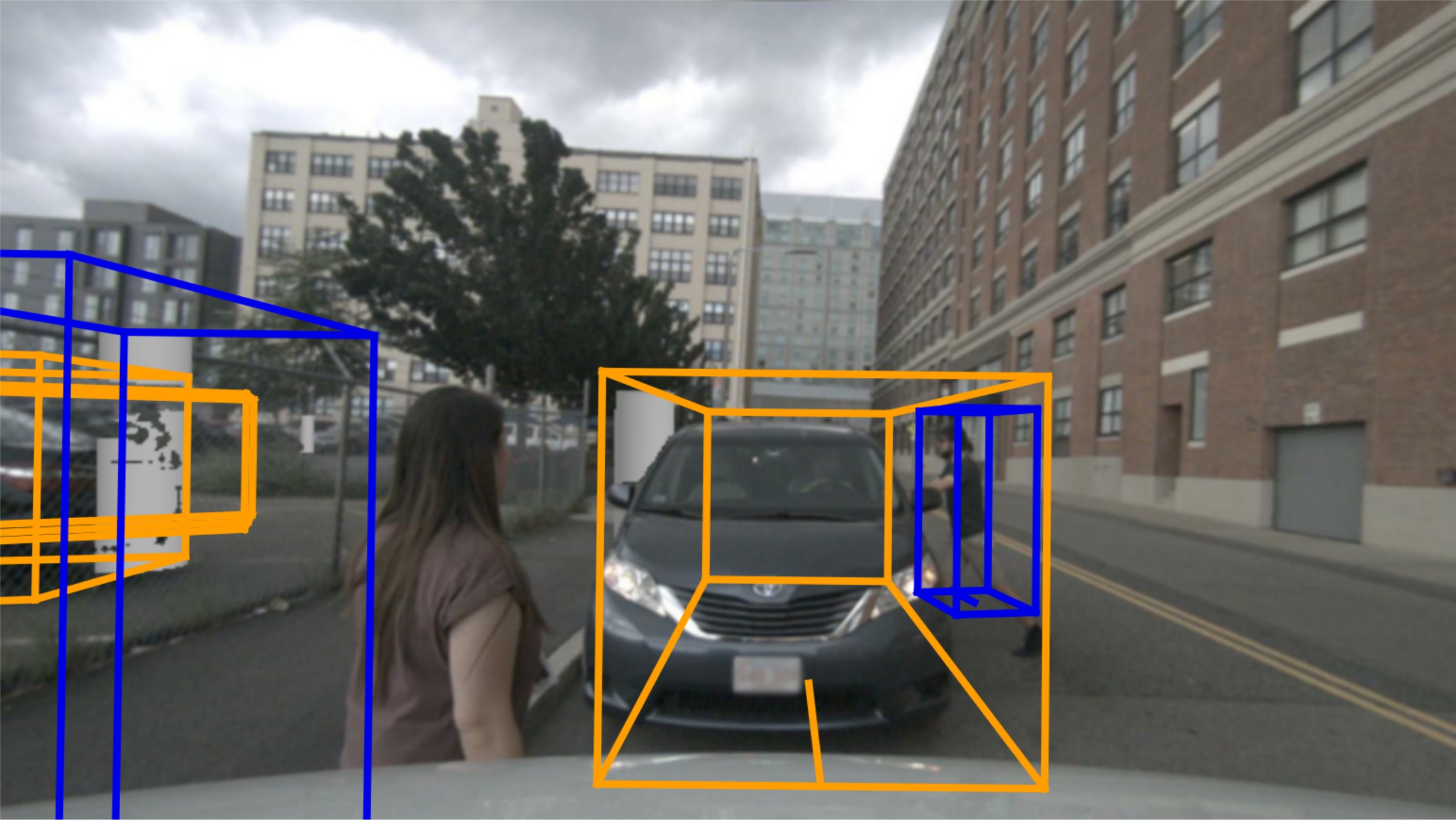}}&
 {\includegraphics[width=0.225\linewidth, height=0.13\linewidth]{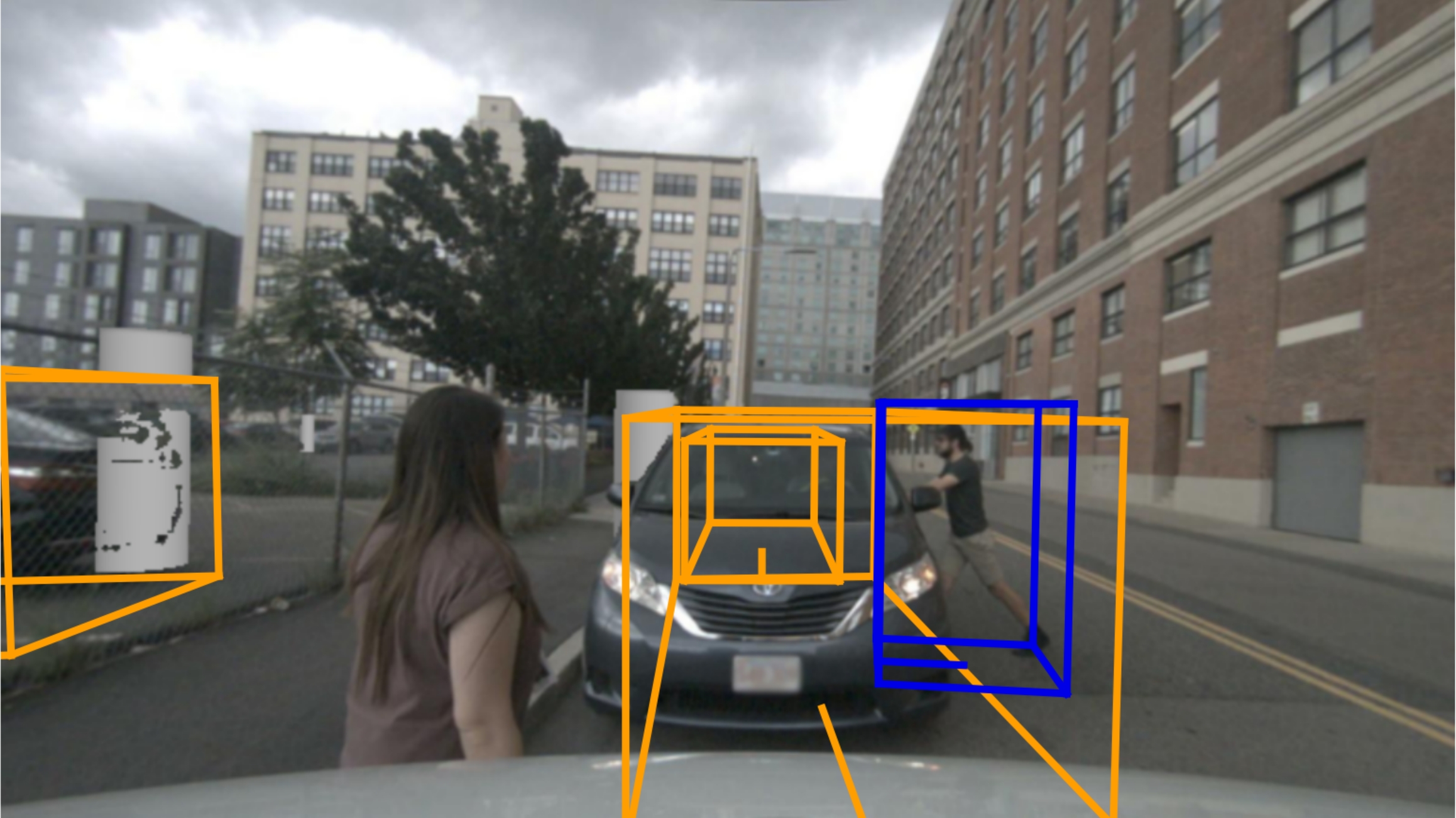}}
 \\

 {\includegraphics[width=0.225\linewidth, height=0.13\linewidth]{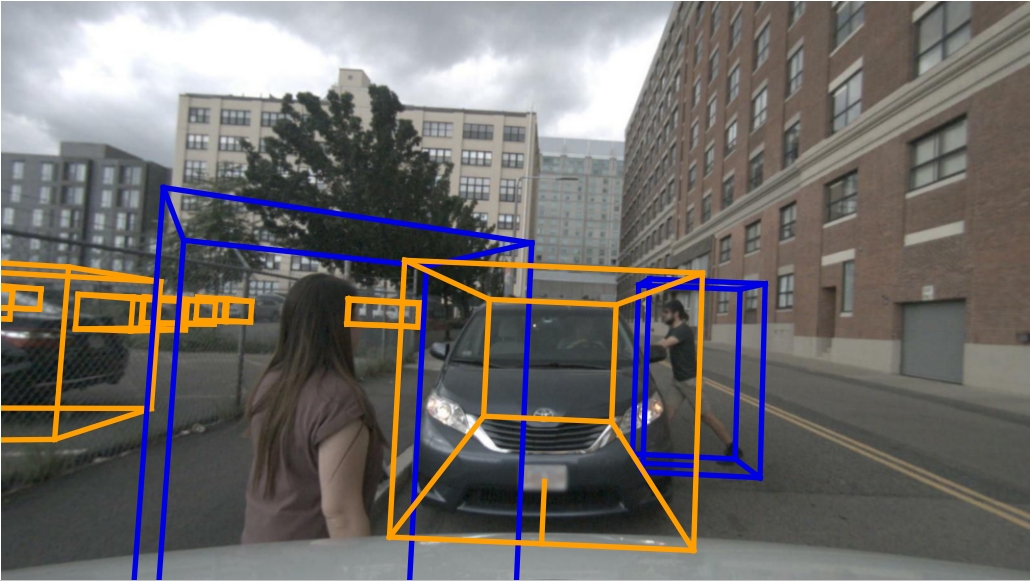}}&
  \multirow{1}*[13mm]{\rotatebox[origin=c]{90}{Attack}} & 
 {\includegraphics[width=0.225\linewidth, height=0.13\linewidth]{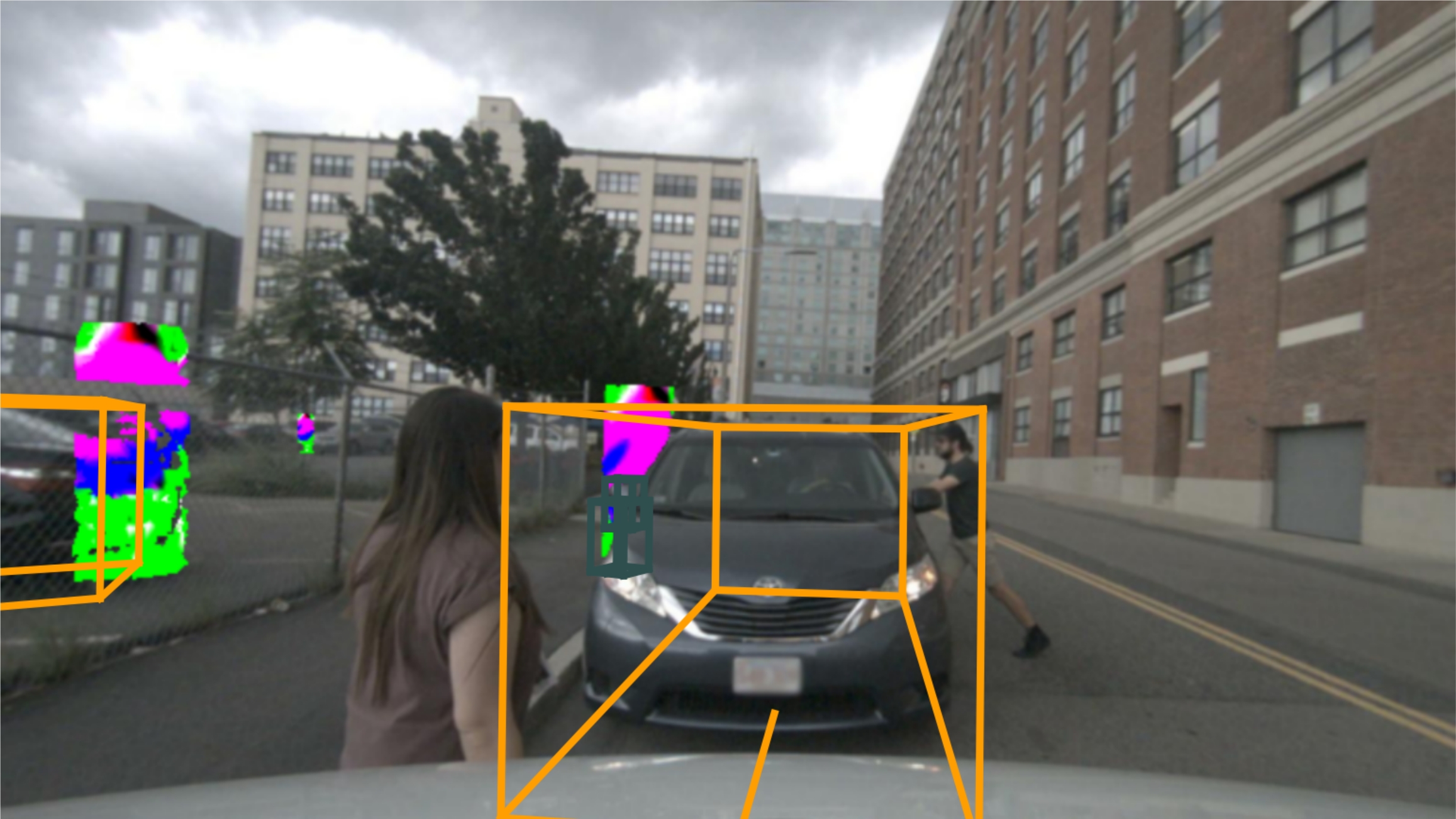}}&
 {\includegraphics[width=0.225\linewidth, height=0.13\linewidth]{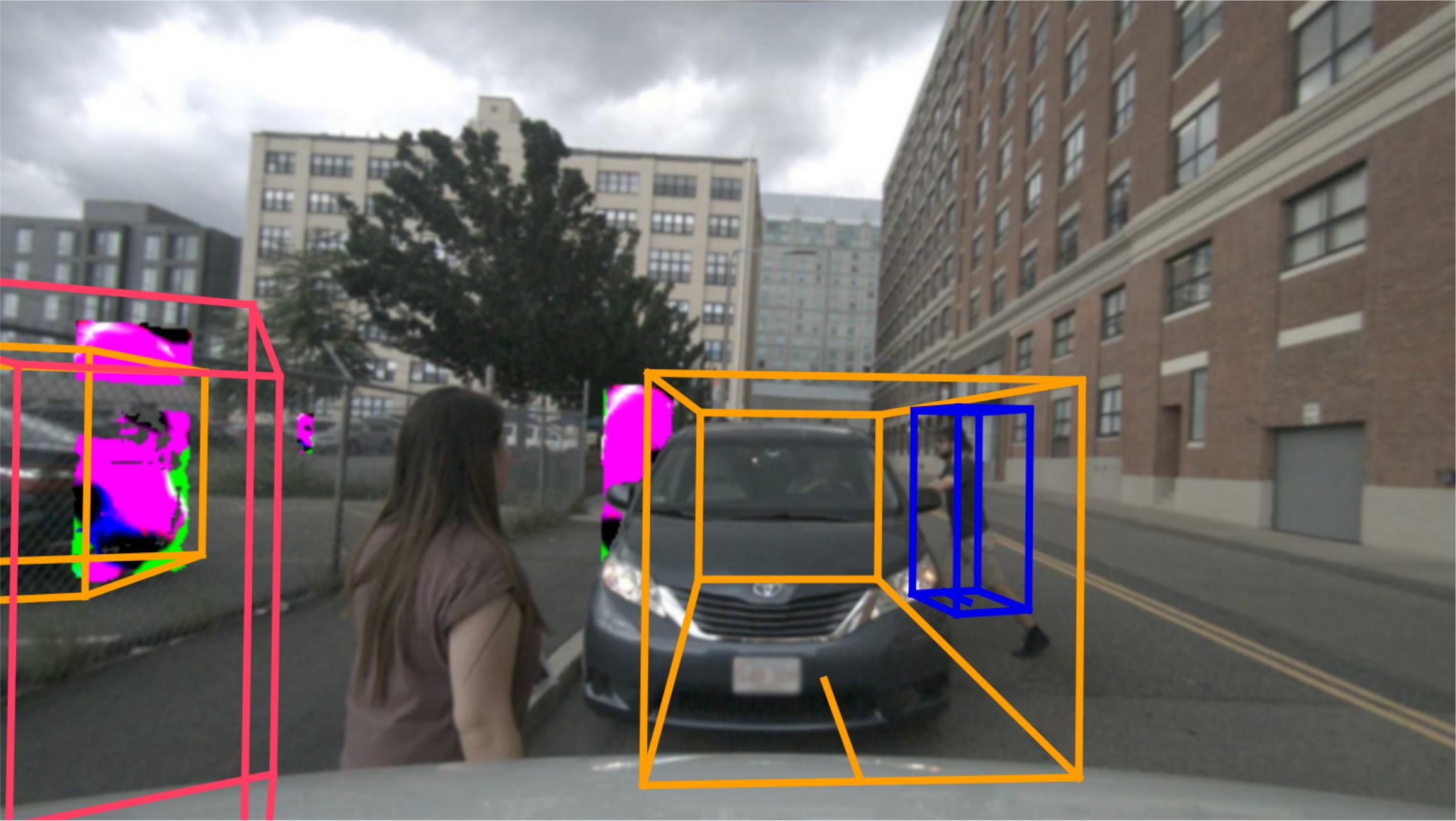}}&
 {\includegraphics[width=0.225\linewidth, height=0.13\linewidth]{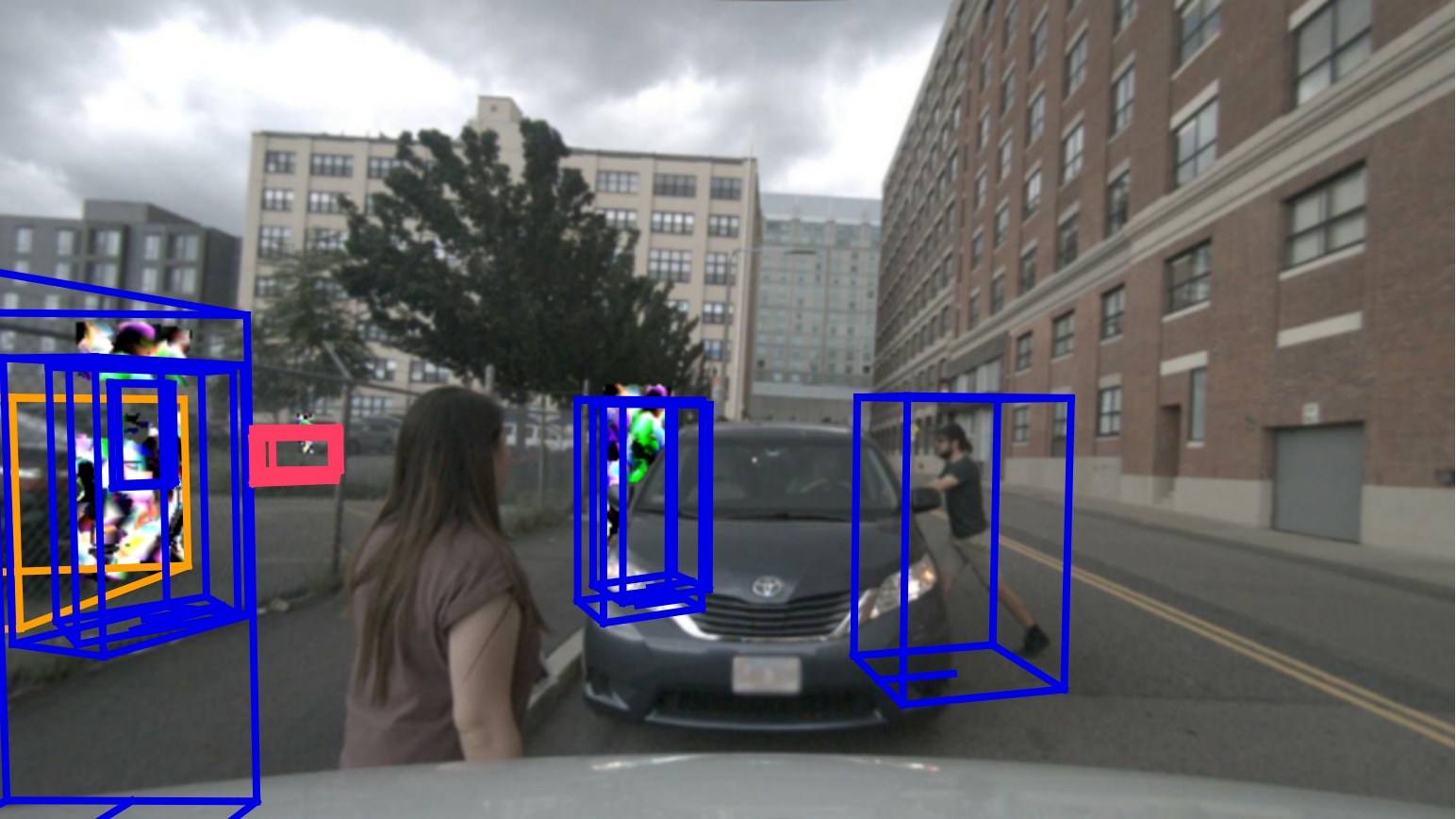}}&
   \\

 {\includegraphics[width=0.225\linewidth, height=0.13\linewidth]{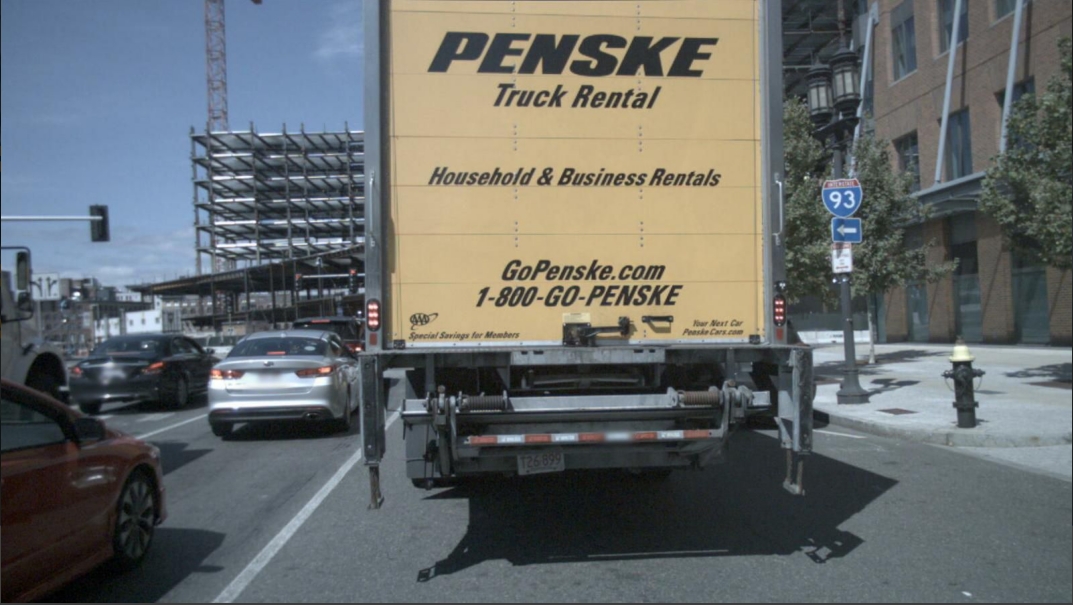}}&
 \multirow{1}*[11mm]{\rotatebox[origin=c]{90}{Init}} & 
 {\includegraphics[width=0.225\linewidth, height=0.13\linewidth]{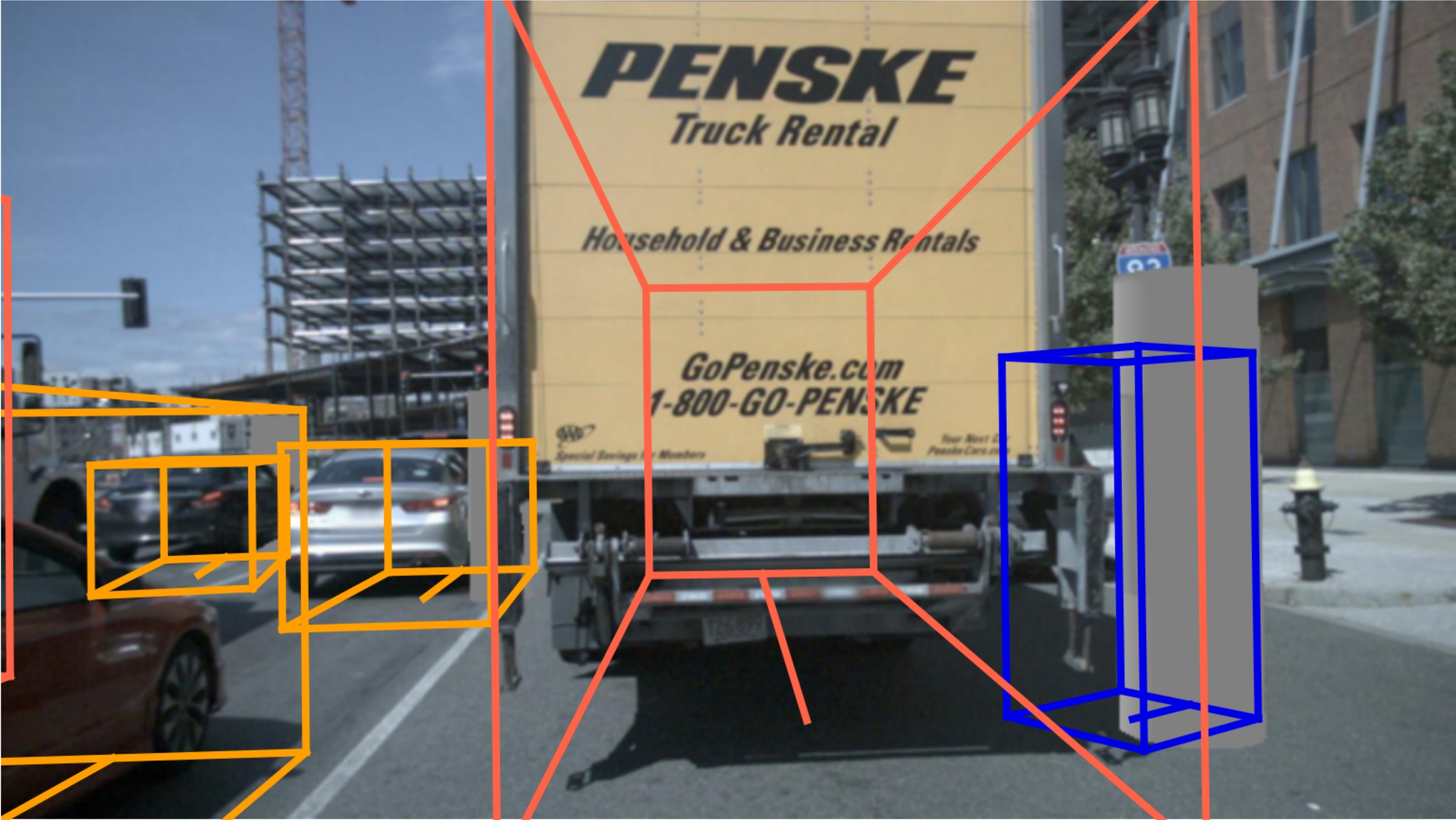}}&
 {\includegraphics[width=0.225\linewidth, height=0.13\linewidth]{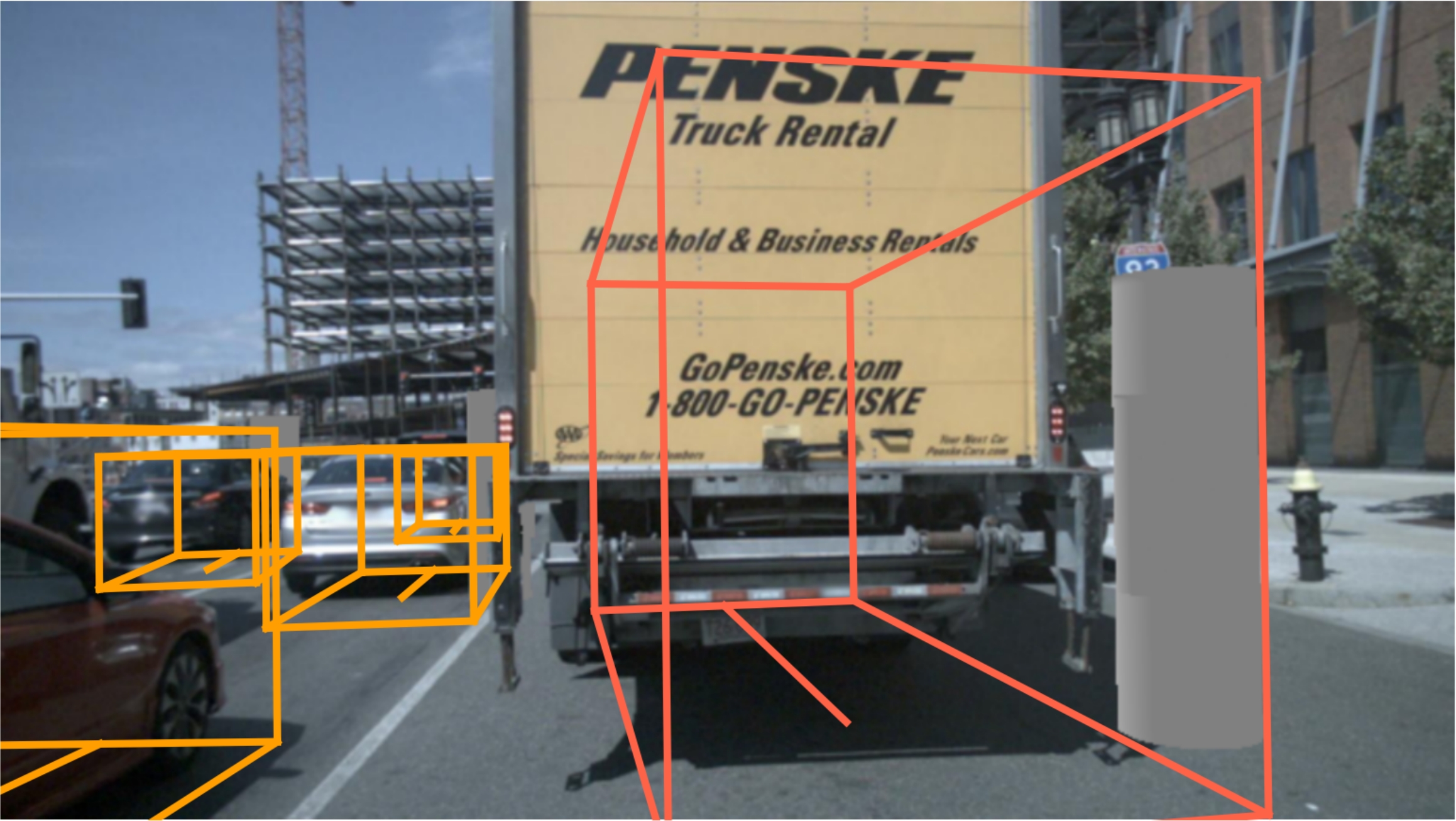}}&
 {\includegraphics[width=0.225\linewidth, height=0.13\linewidth]{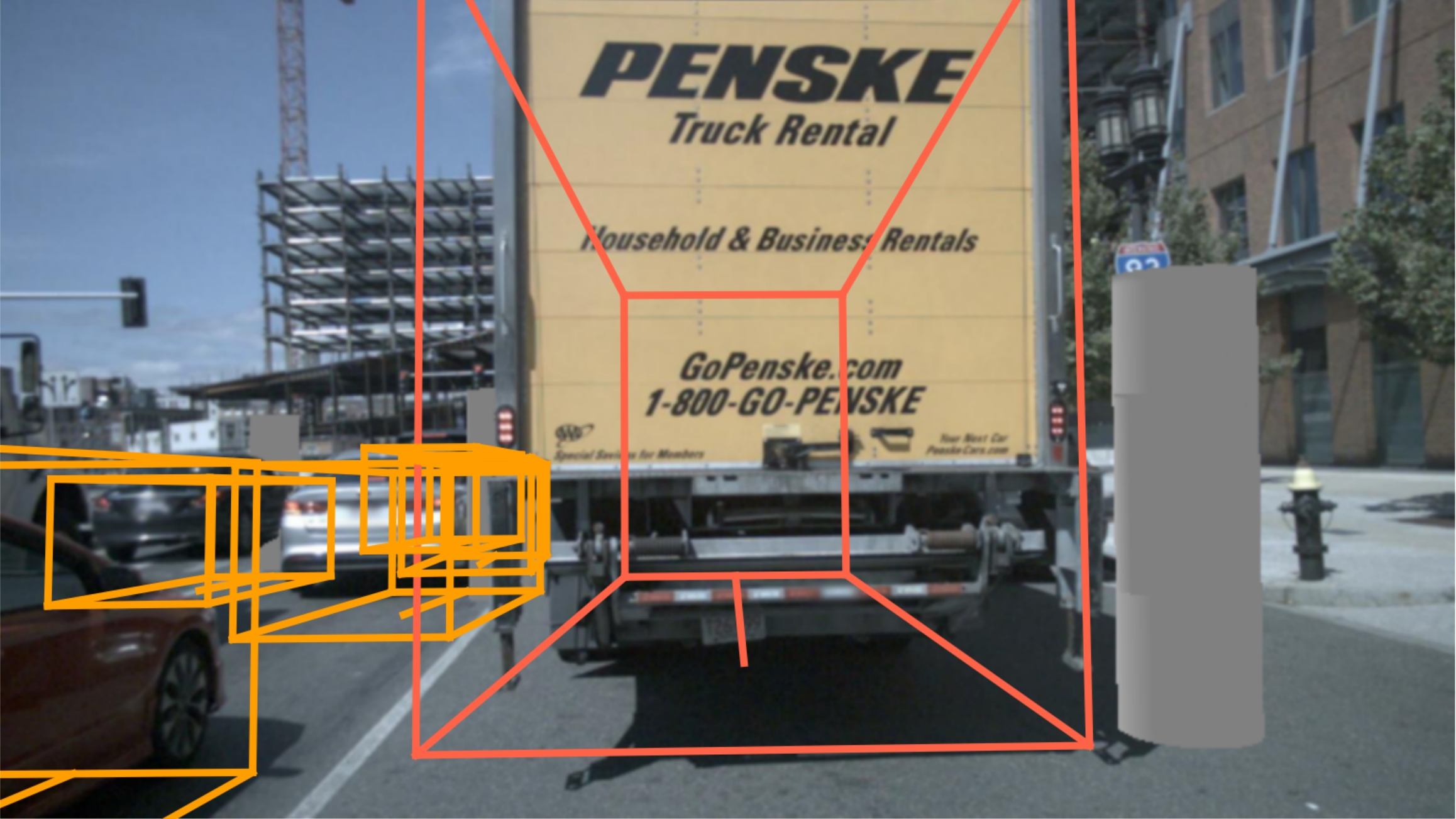}}
 \\

 {\includegraphics[width=0.225\linewidth, height=0.13\linewidth]{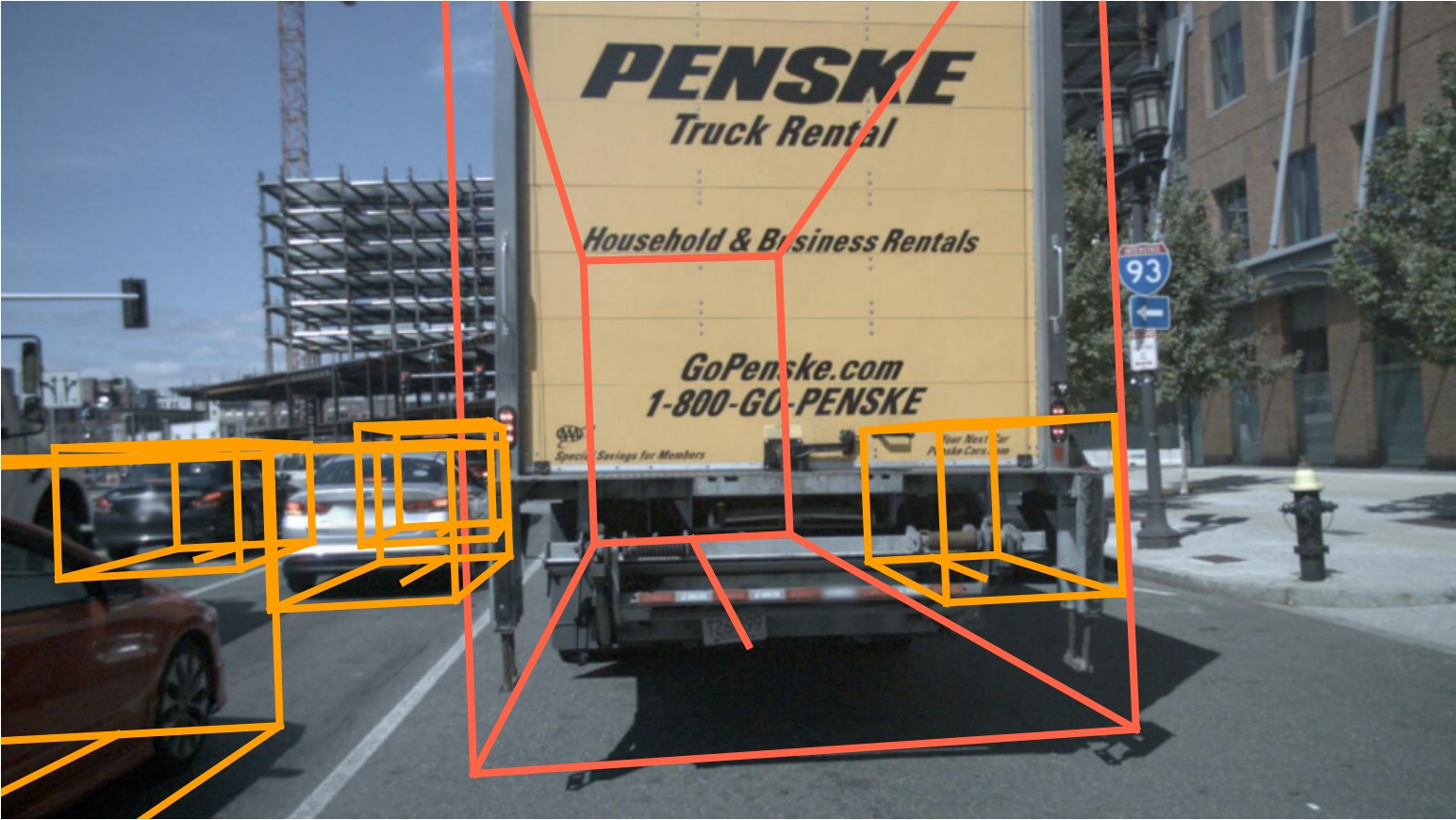}}&
  \multirow{1}*[13mm]{\rotatebox[origin=c]{90}{Attack}} & 
 {\includegraphics[width=0.225\linewidth, height=0.13\linewidth]{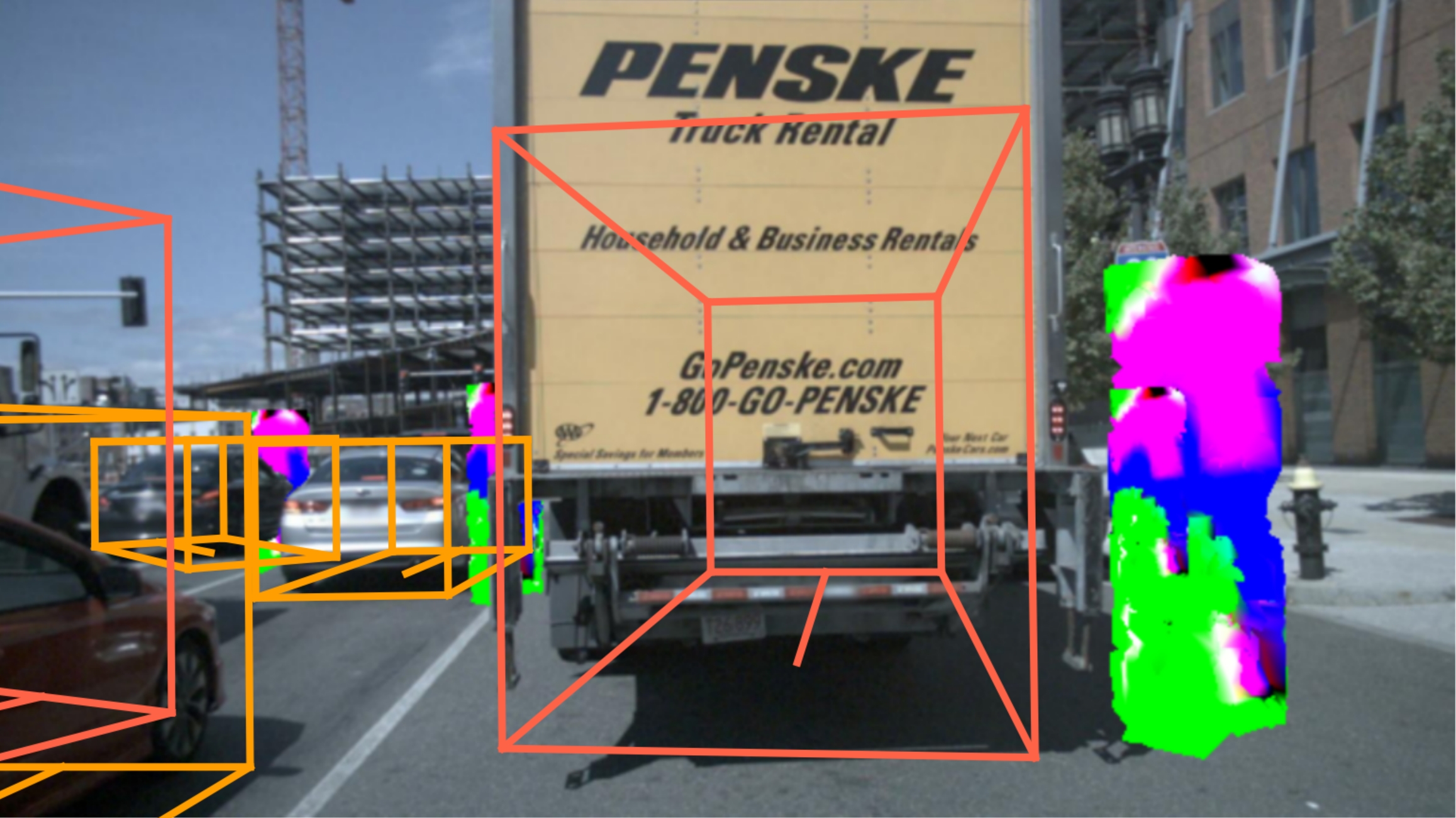}}&
 {\includegraphics[width=0.225\linewidth, height=0.13\linewidth]{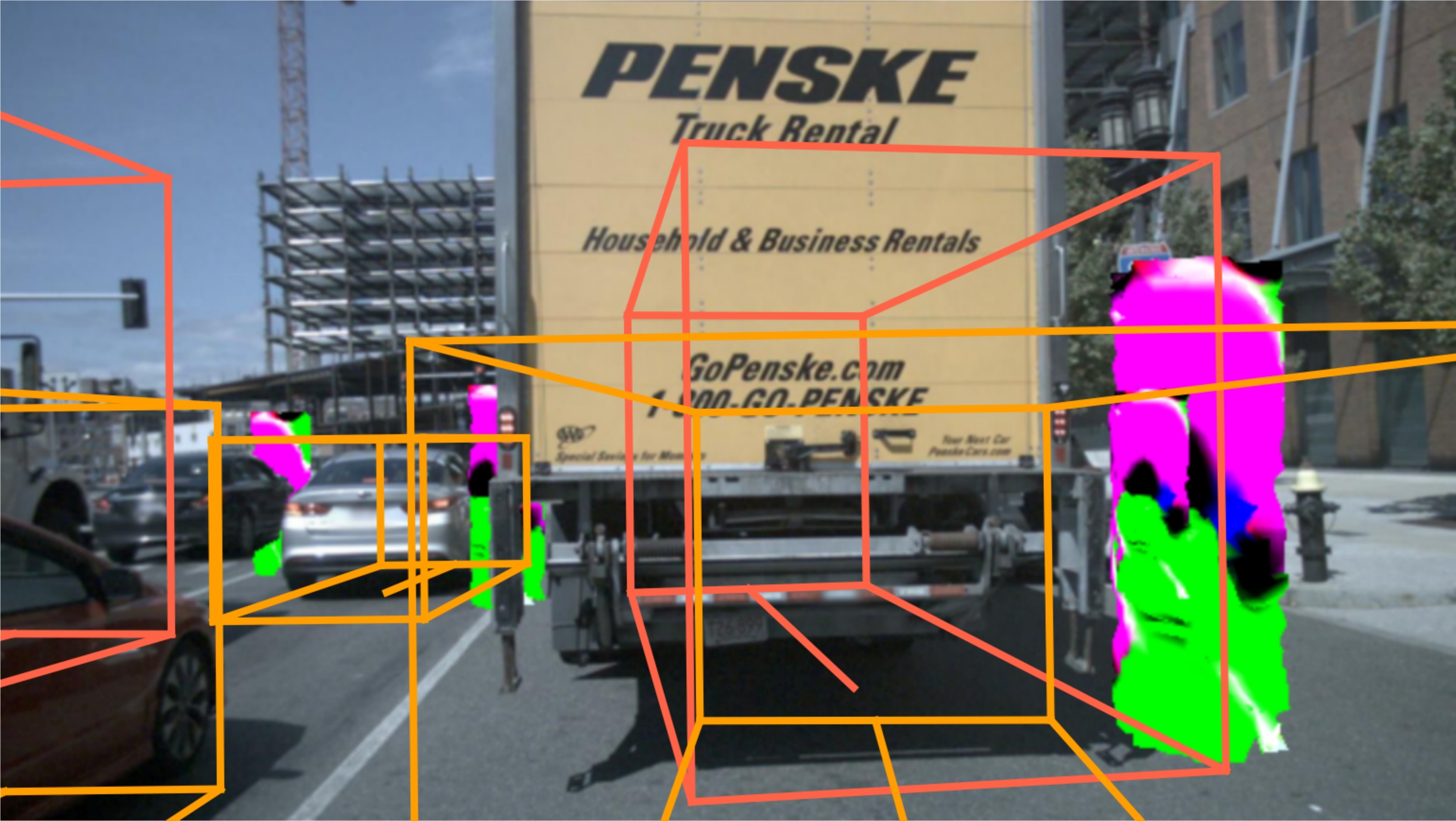}}&
 {\includegraphics[width=0.225\linewidth, height=0.13\linewidth]{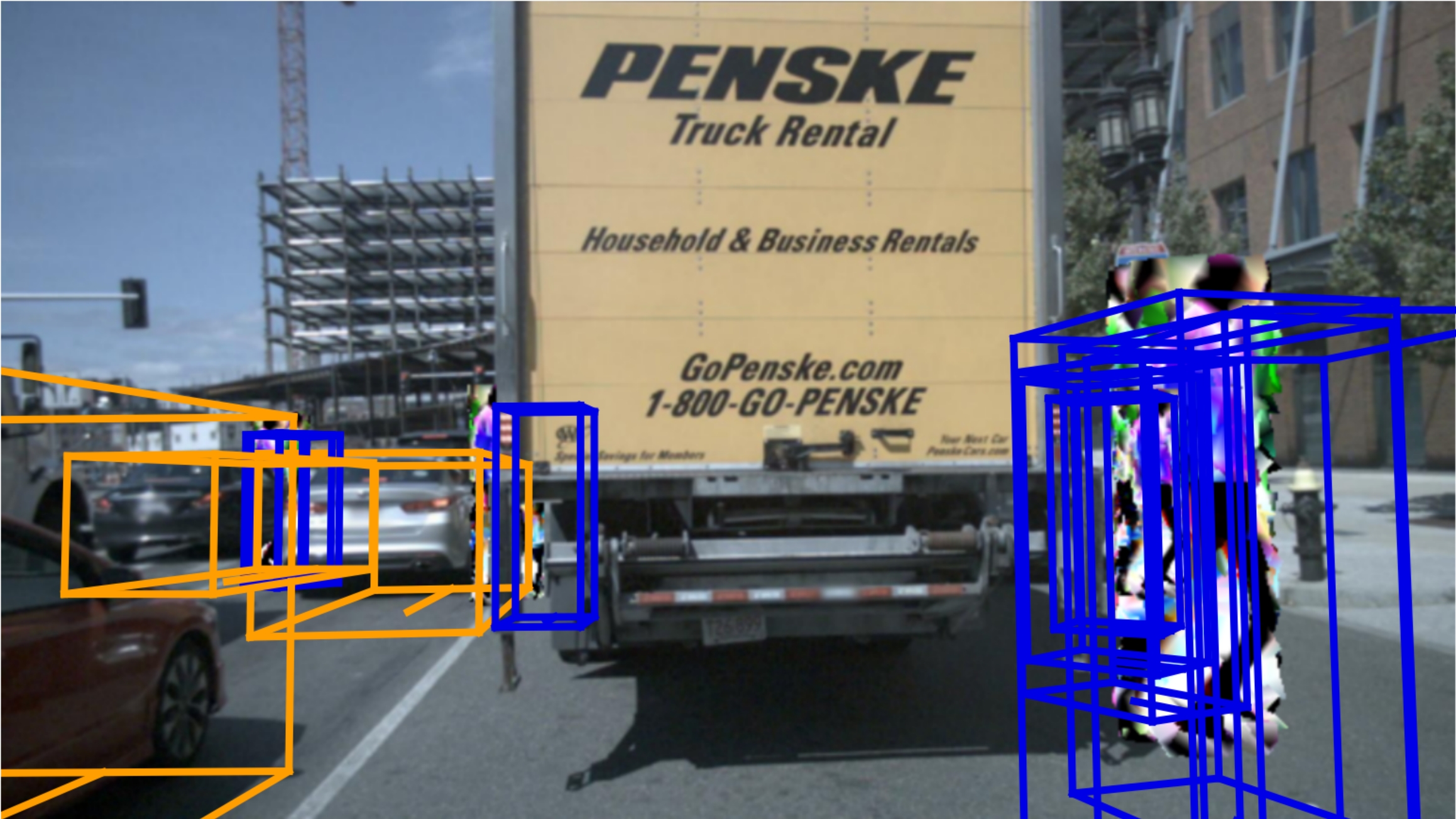}}&
   \\

 {\includegraphics[width=0.225\linewidth, height=0.13\linewidth]{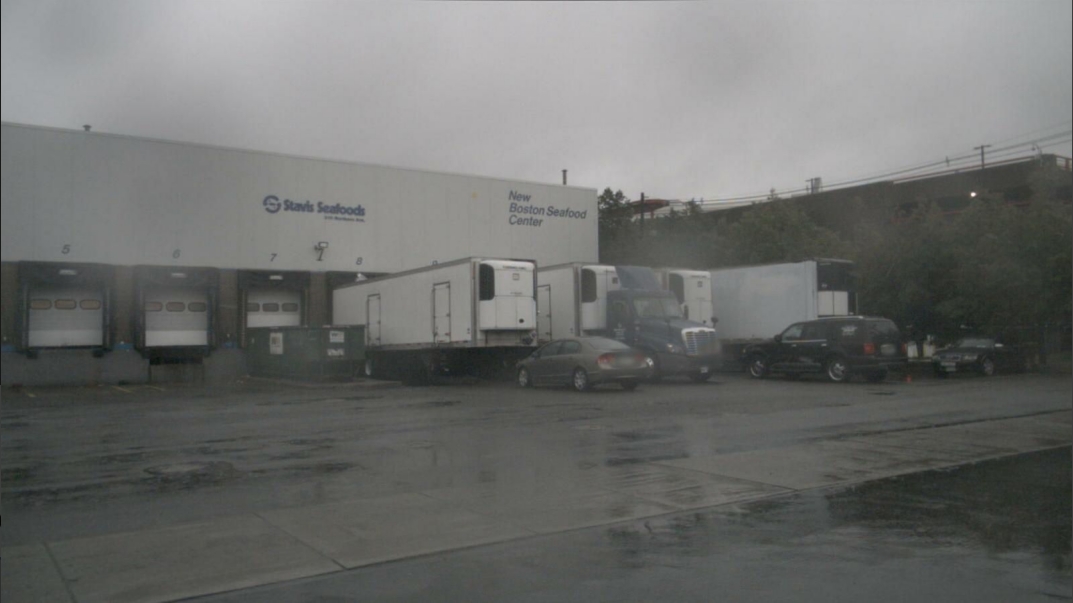}}&
 \multirow{1}*[11mm]{\rotatebox[origin=c]{90}{Init}} & 
 {\includegraphics[width=0.225\linewidth, height=0.13\linewidth]{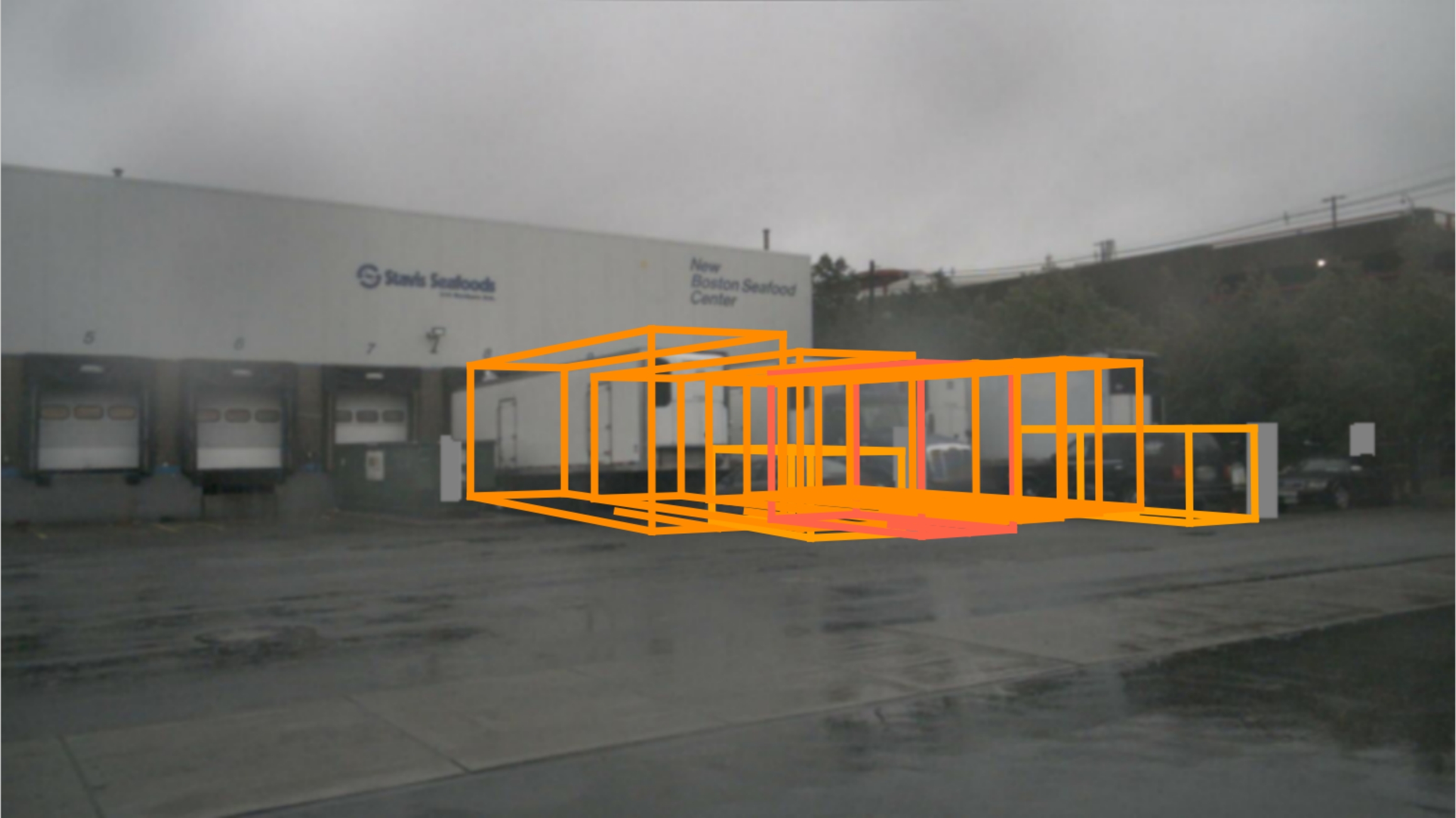}}&
 {\includegraphics[width=0.225\linewidth, height=0.13\linewidth]{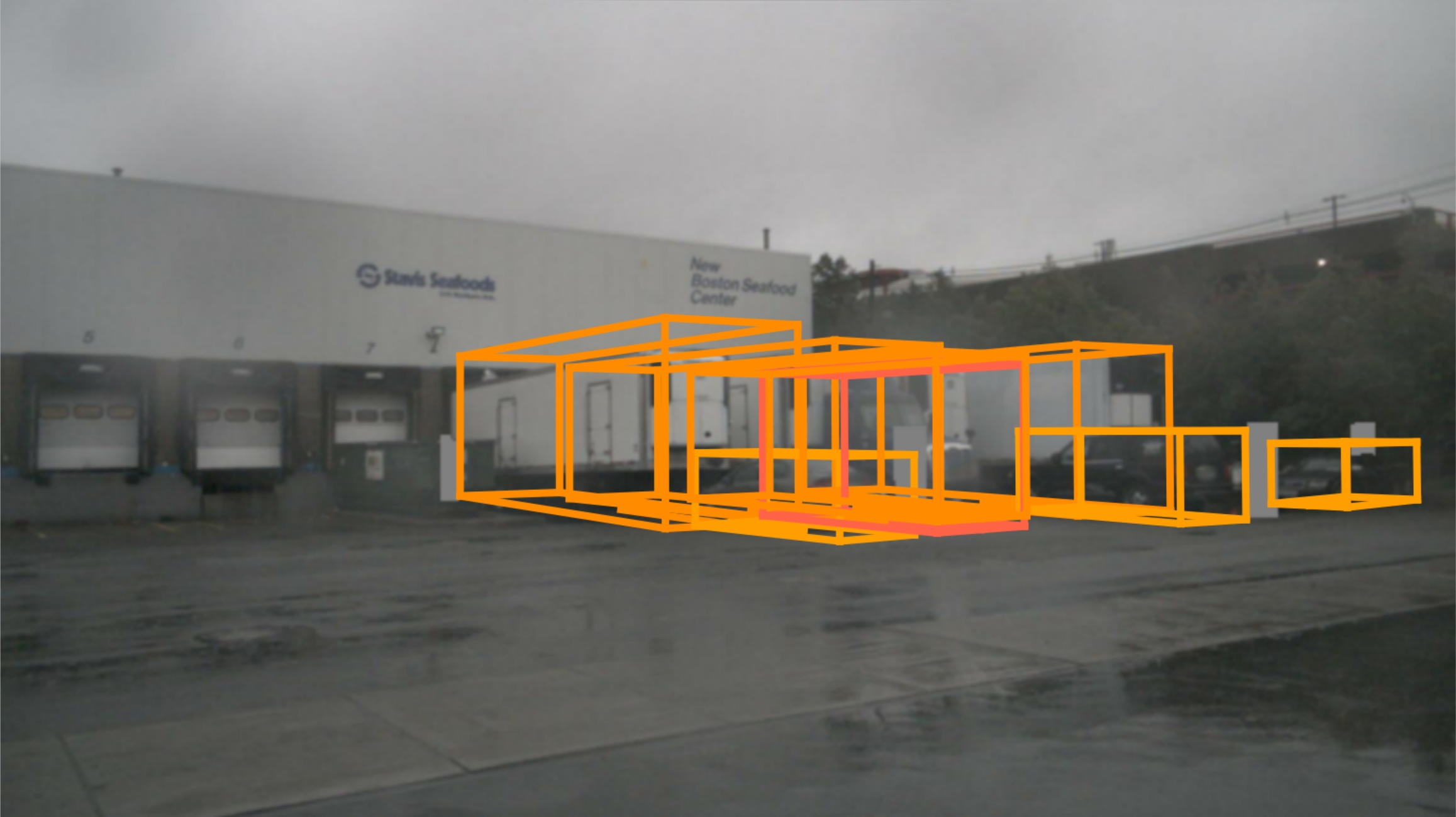}}&
 {\includegraphics[width=0.225\linewidth, height=0.13\linewidth]{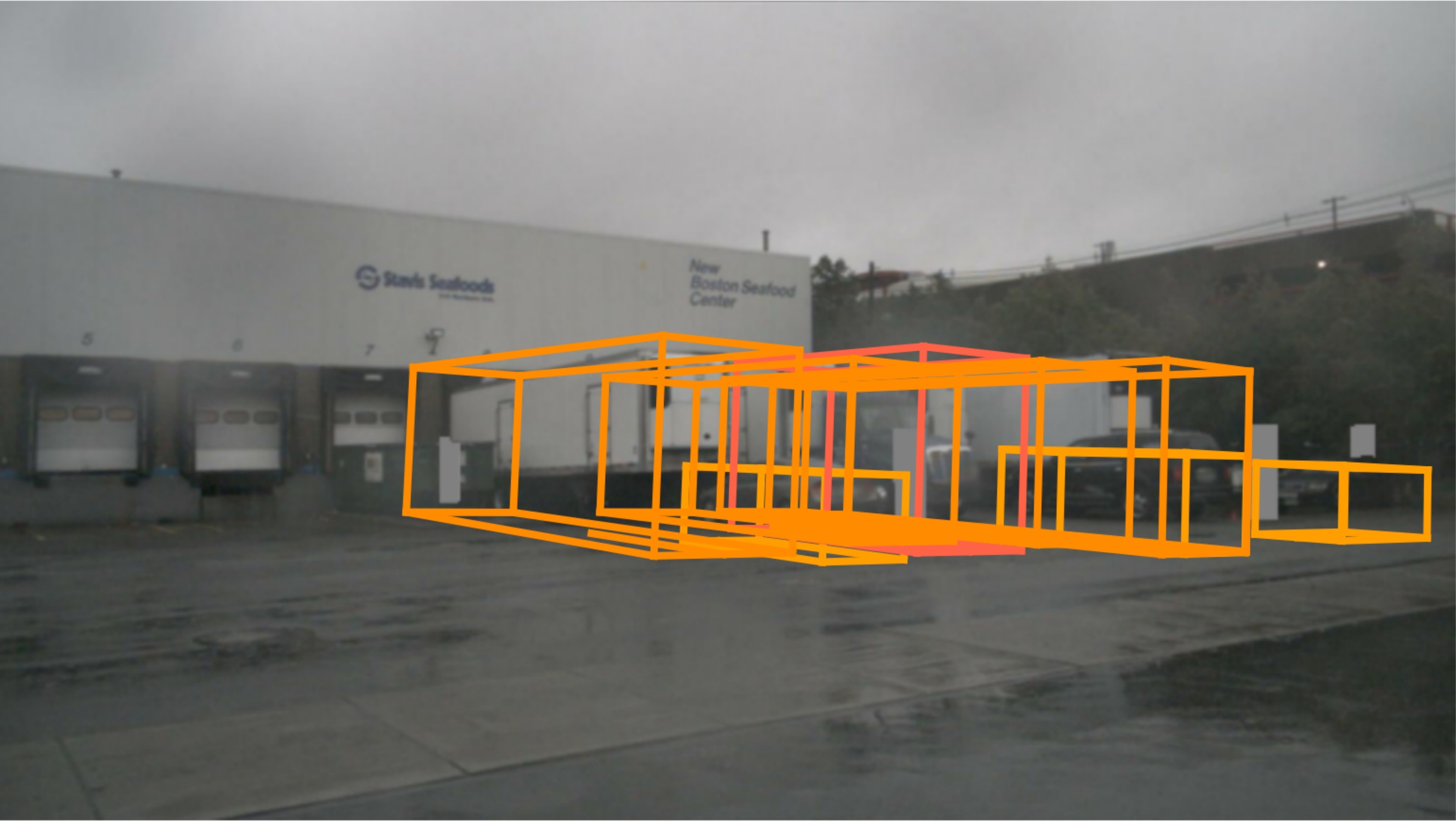}}
 \\

 {\includegraphics[width=0.225\linewidth, height=0.13\linewidth]{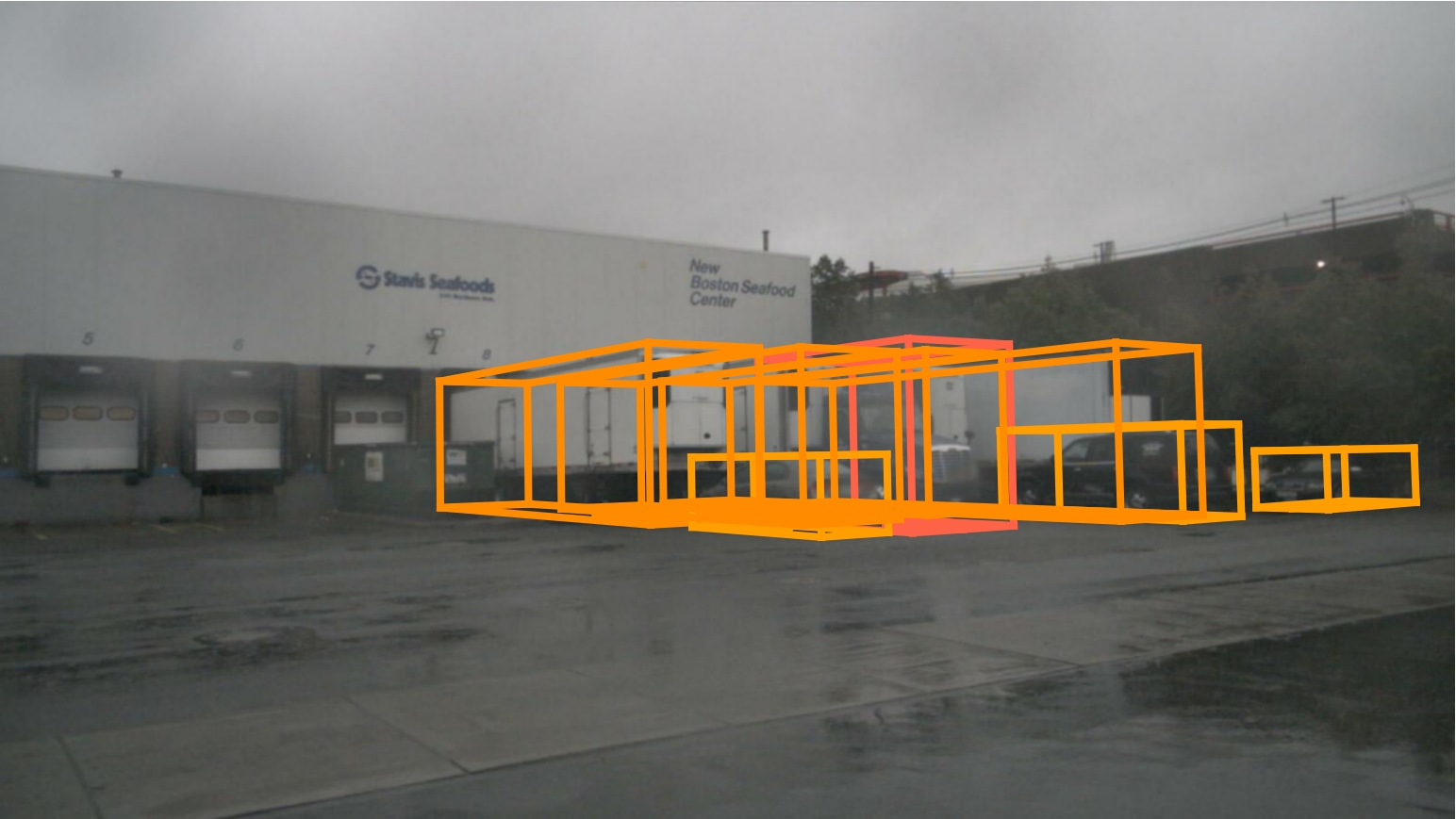}}&
  \multirow{1}*[13mm]{\rotatebox[origin=c]{90}{Attack}} & 
 {\includegraphics[width=0.225\linewidth, height=0.13\linewidth]{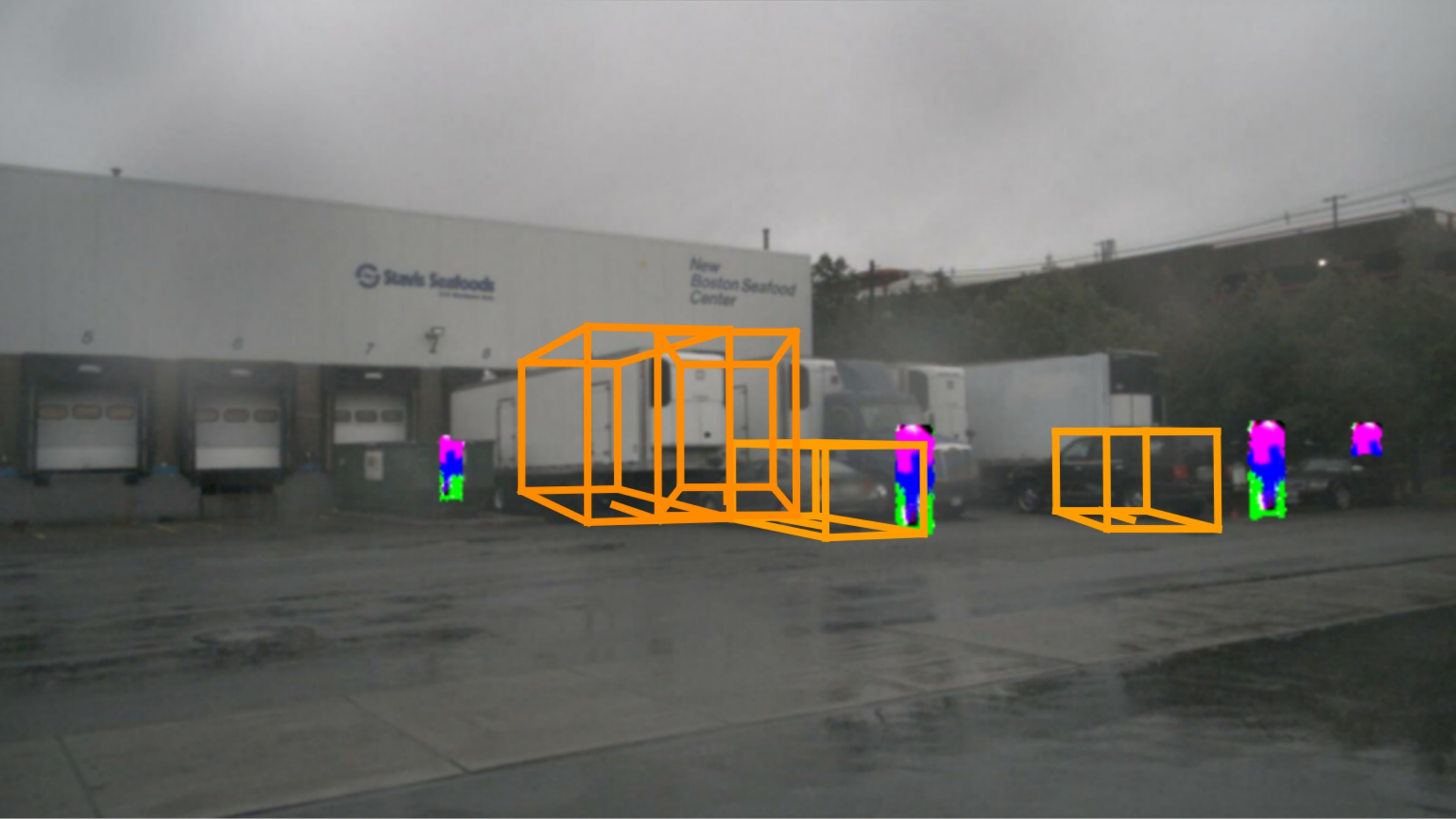}}&
 {\includegraphics[width=0.225\linewidth, height=0.13\linewidth]{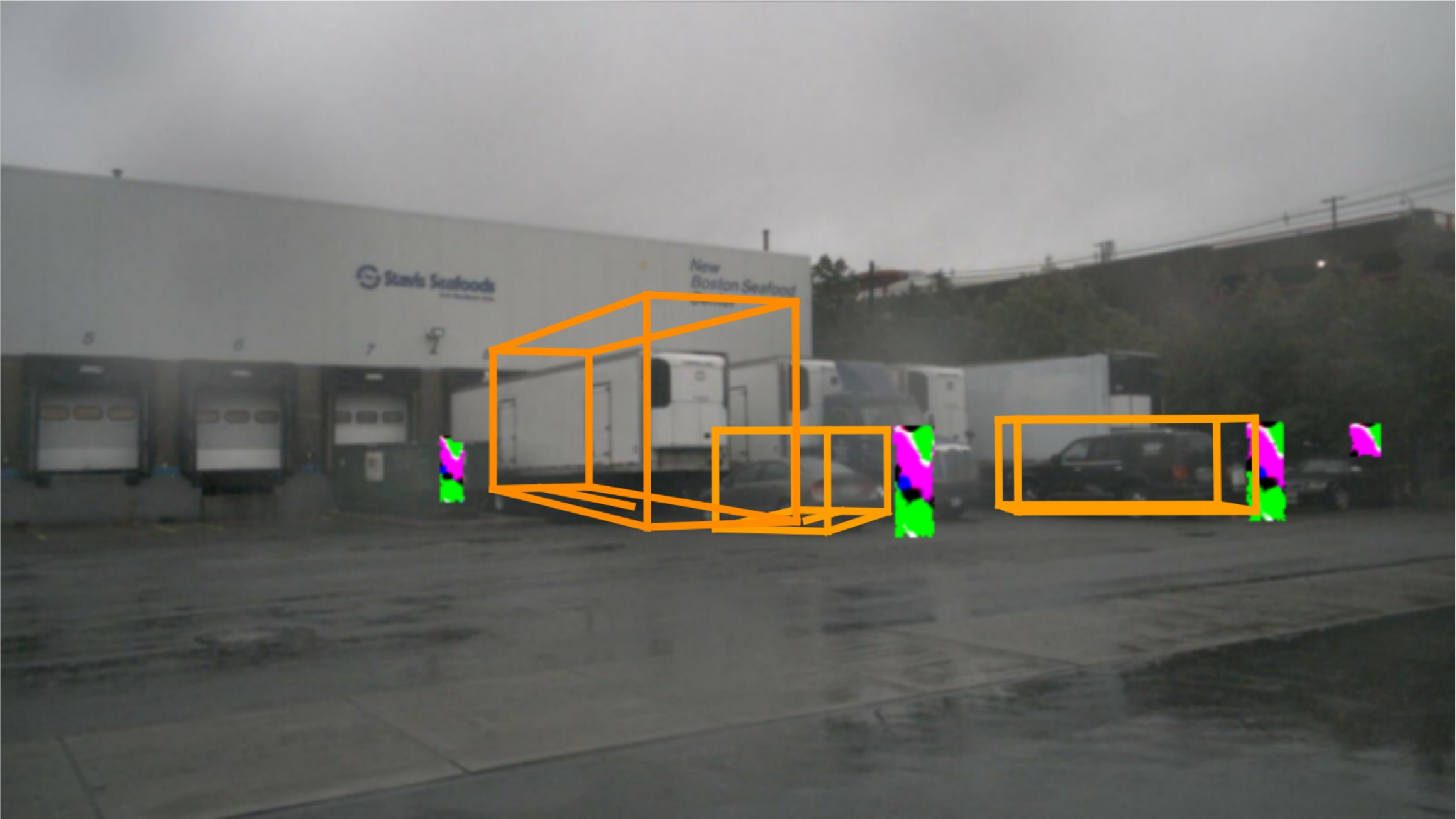}}&
 {\includegraphics[width=0.225\linewidth, height=0.13\linewidth]{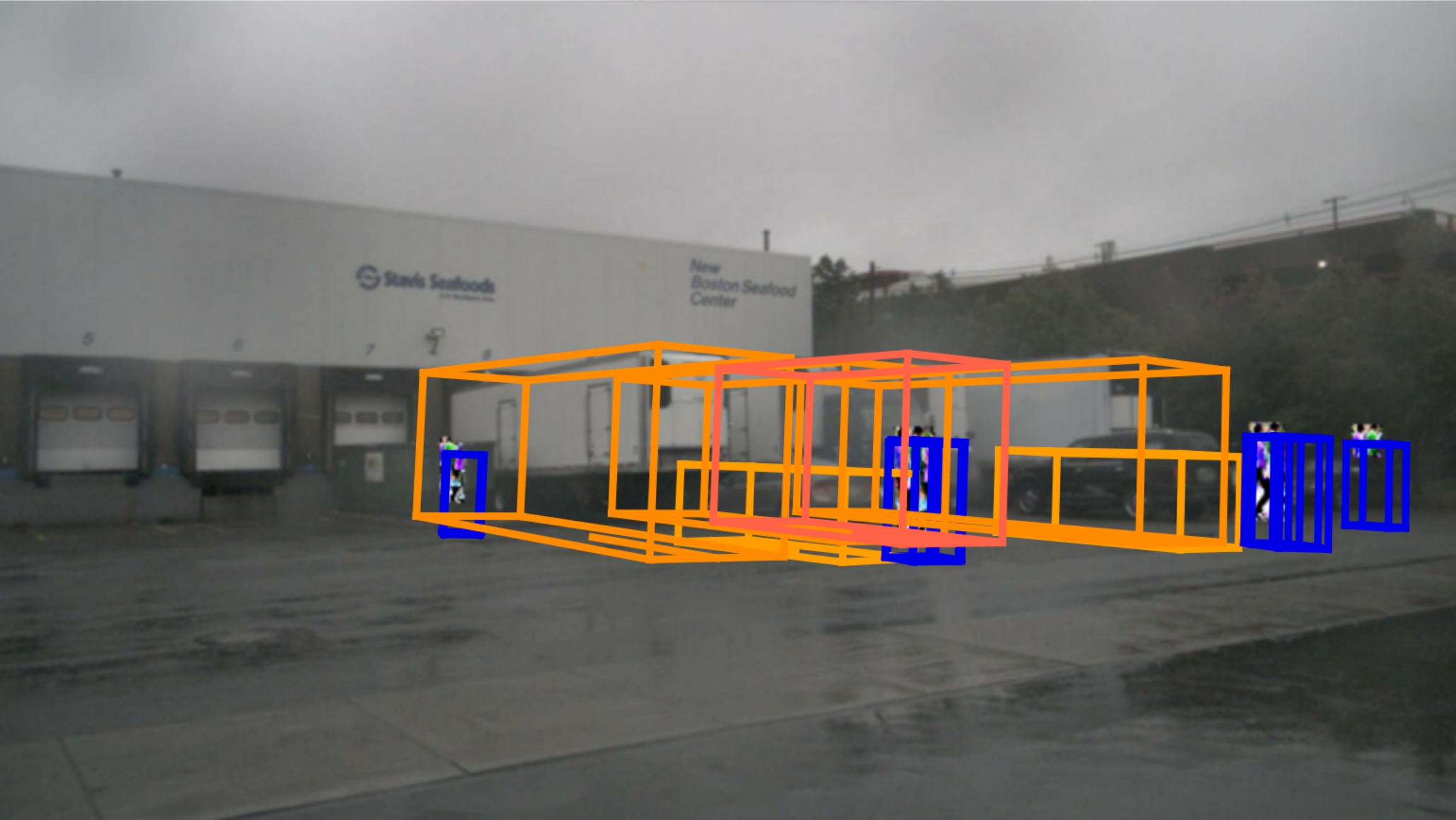}}&
   \\
   
 {GT} & 
   & Ours-BEVDet &  Ours-BEVDet4D &  Ours-BEVFormer  \\
   \end{tabular} 
    \caption{\textbf{Visualizations of attack effects in image view.}} 
    \label{fig:2d_compare1}
\end{figure*}

\begin{figure*}[h]
   \centering
   \small
   \begin{tabular}{{c@{ } c@{ } c@{ } c@{ }  }}

\multirow{1}*[12mm]{\rotatebox[origin=c]{90}{Init}} & 
 {\includegraphics[width=0.32\linewidth,height=0.22\linewidth]{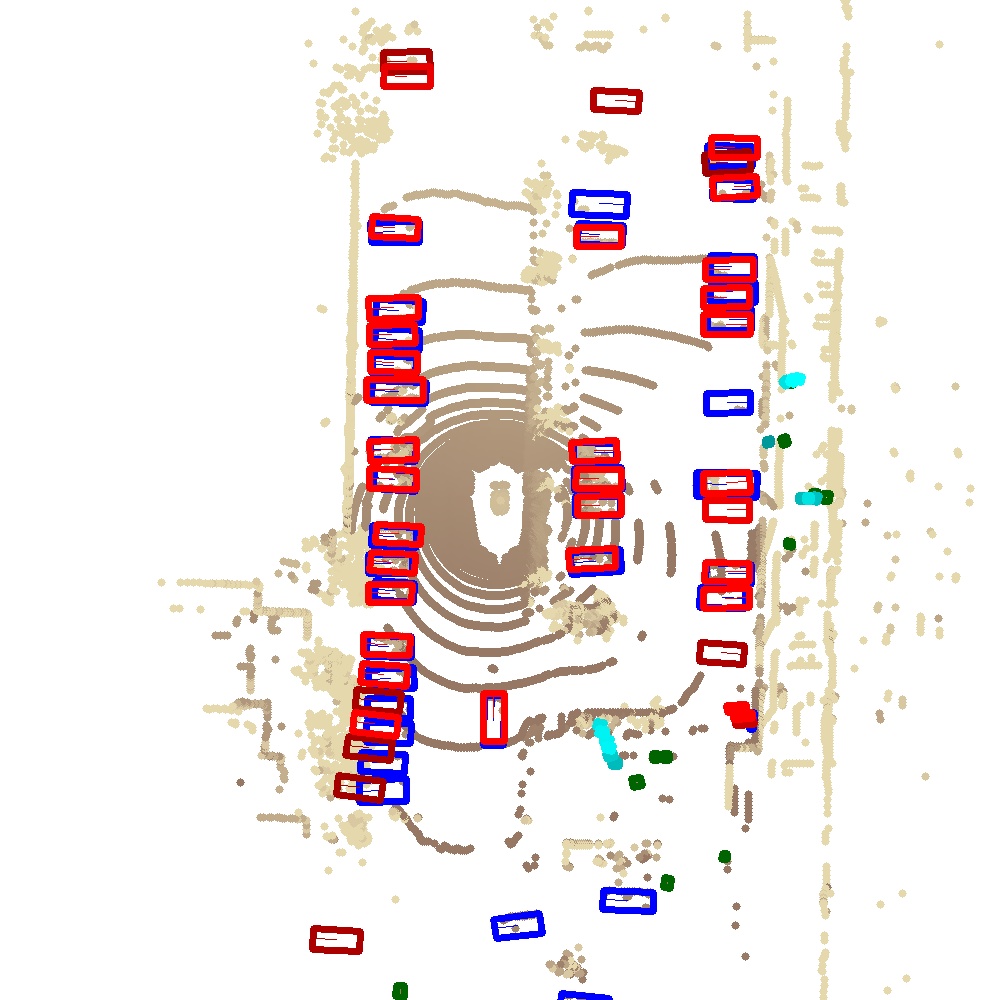}}&
 {\includegraphics[width=0.32\linewidth,height=0.22\linewidth]{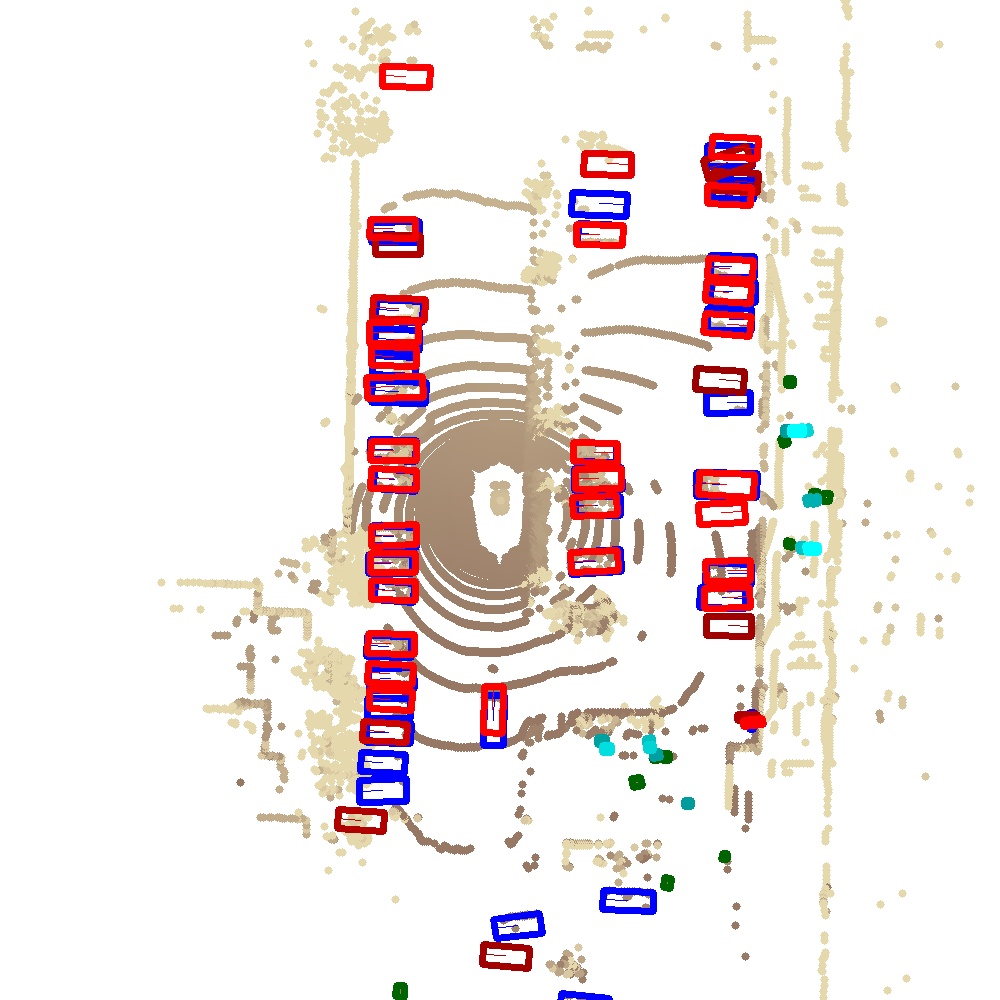}}&
 {\includegraphics[width=0.32\linewidth,height=0.22\linewidth]{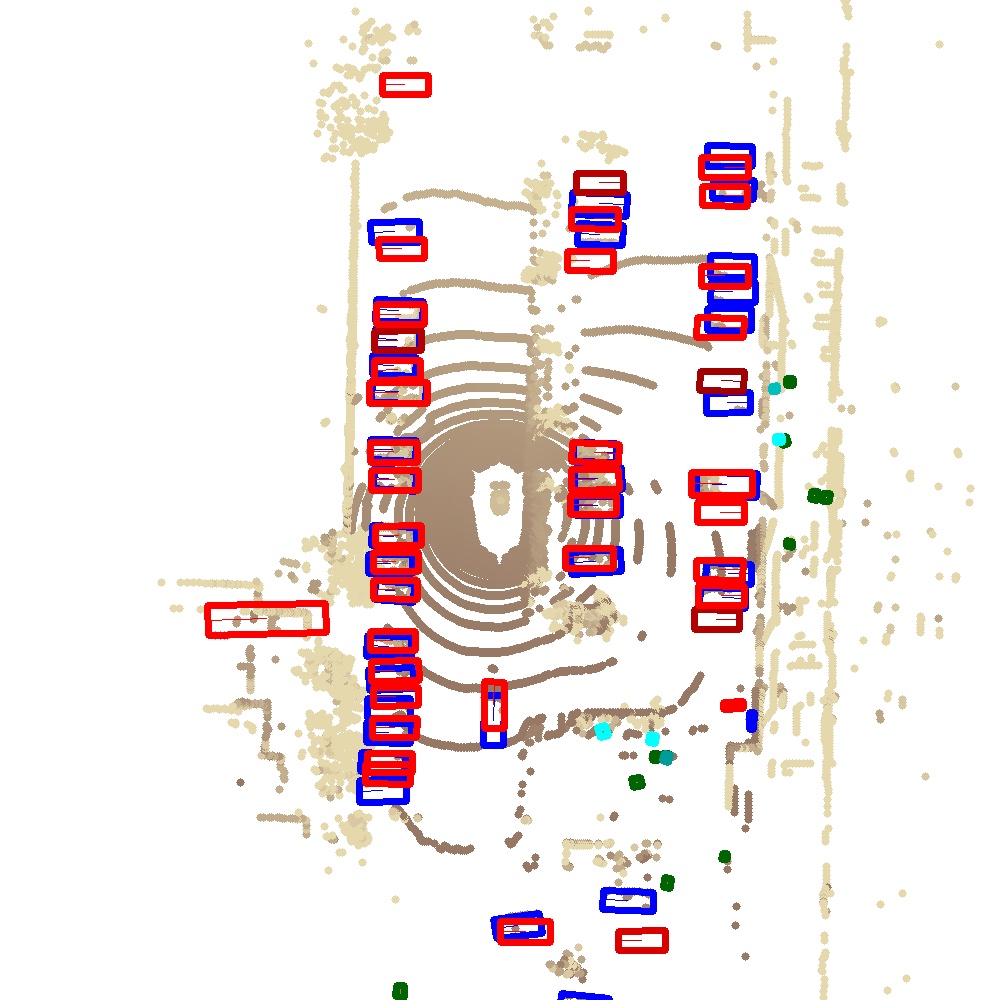}}
 \\
 \multirow{1}*[15mm]{\rotatebox[origin=c]{90}{Attack}} & 
 {\includegraphics[width=0.32\linewidth,height=0.22\linewidth]{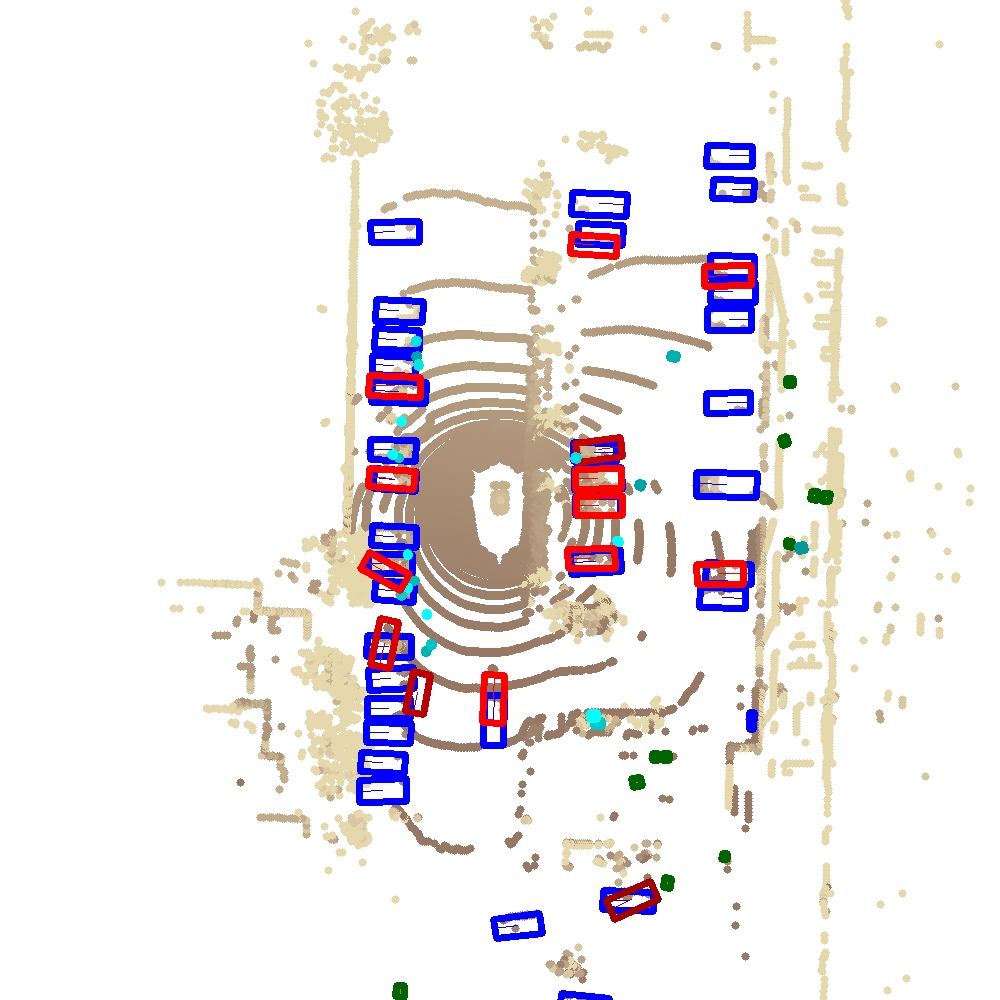}}&
 {\includegraphics[width=0.32\linewidth,height=0.22\linewidth]{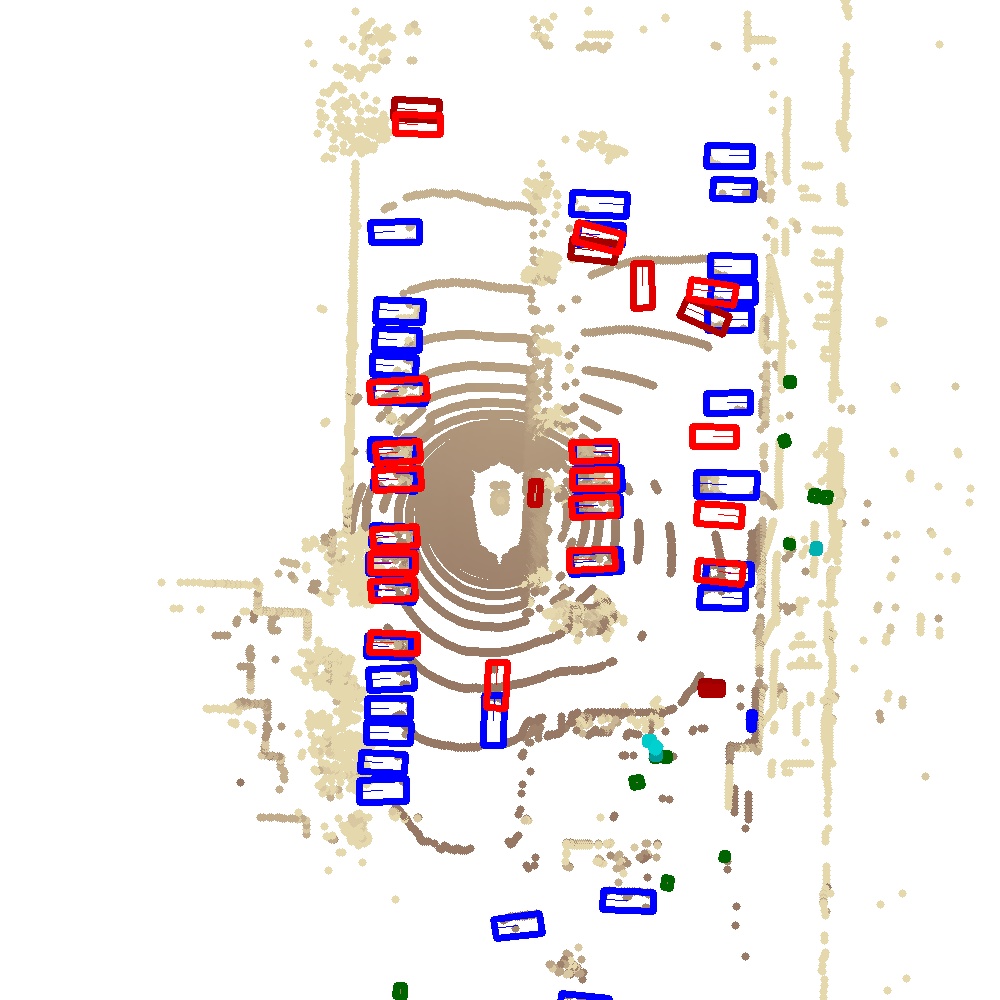}}&
 {\includegraphics[width=0.32\linewidth,height=0.22\linewidth]{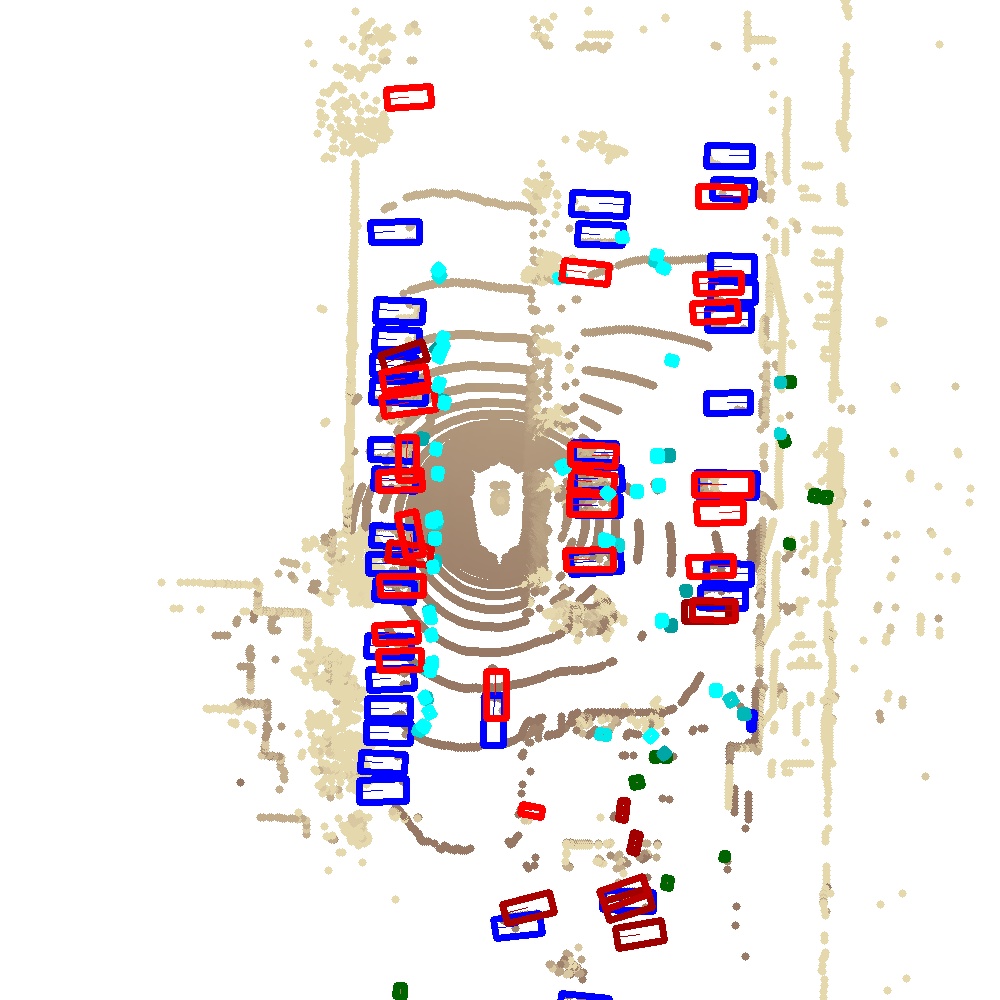}}

\\

\multirow{1}*[12mm]{\rotatebox[origin=c]{90}{Init}} & 
 {\includegraphics[width=0.32\linewidth,height=0.22\linewidth]{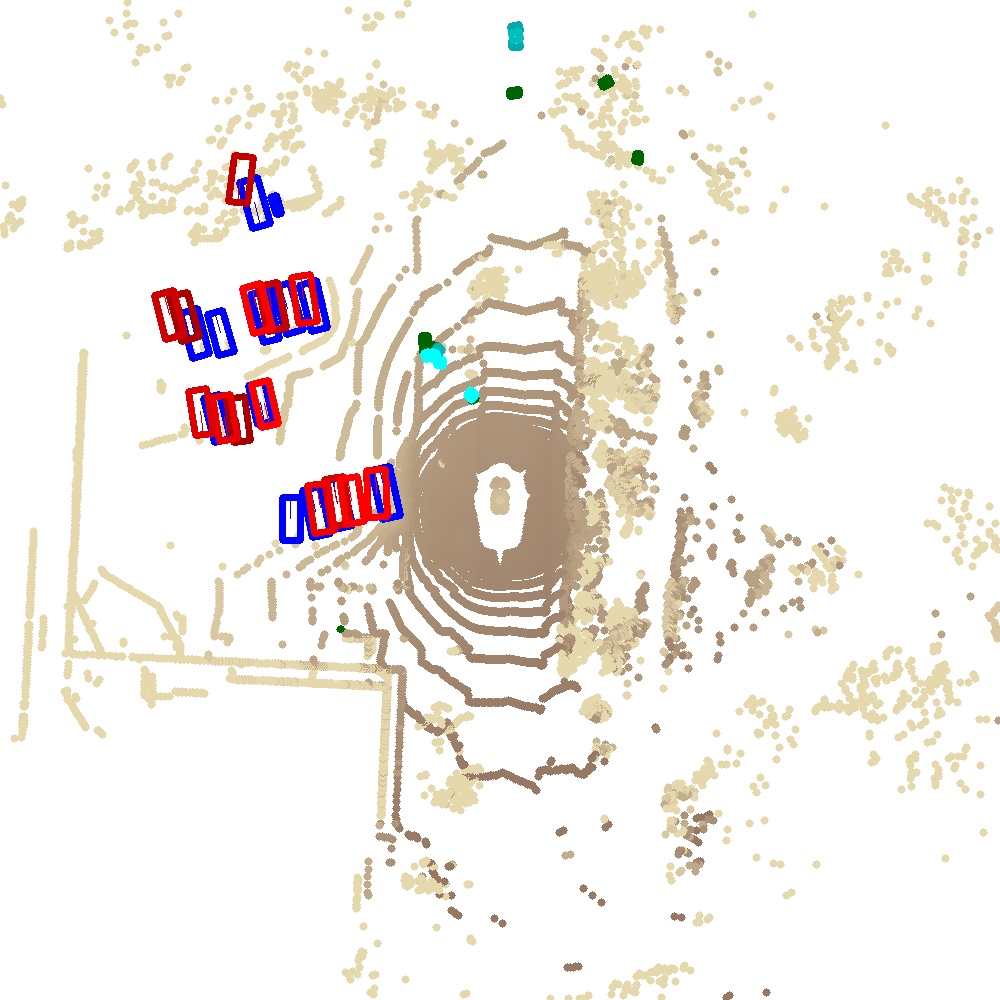}}&
 {\includegraphics[width=0.32\linewidth,height=0.22\linewidth]{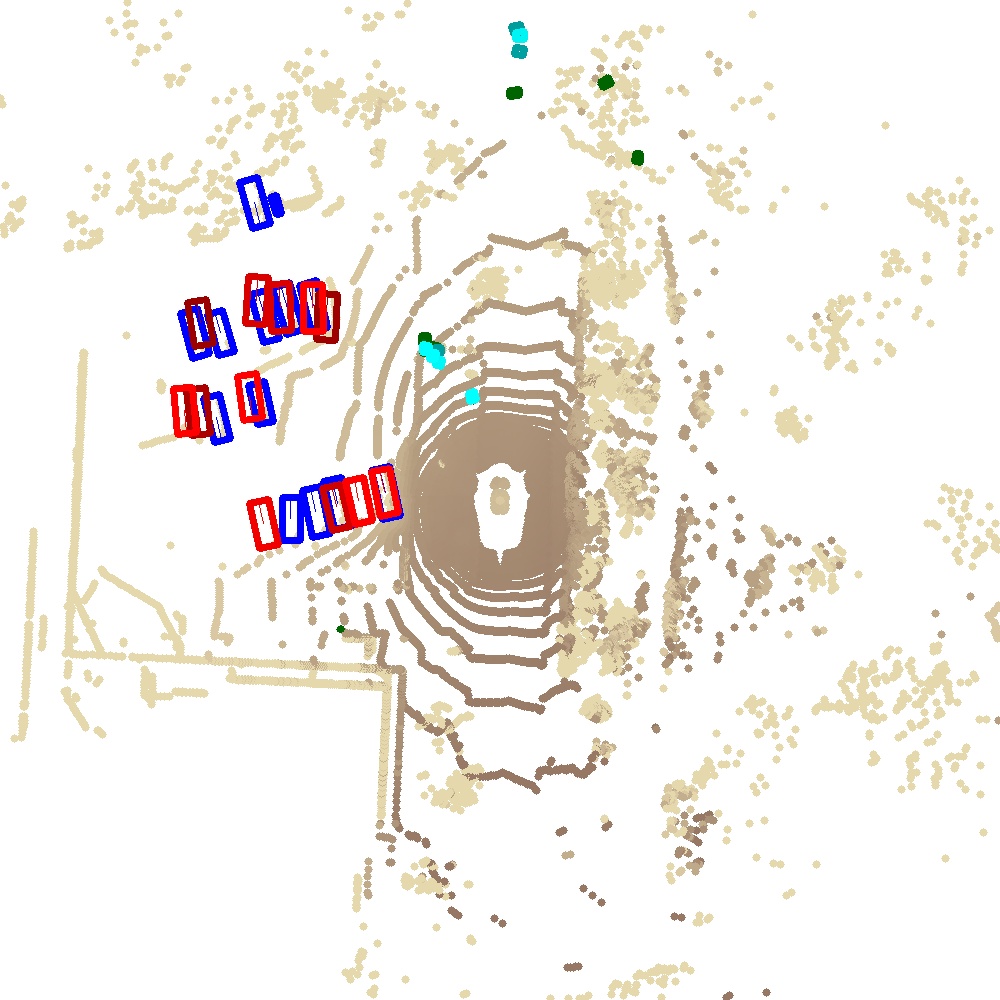}}&
 {\includegraphics[width=0.32\linewidth,height=0.22\linewidth]{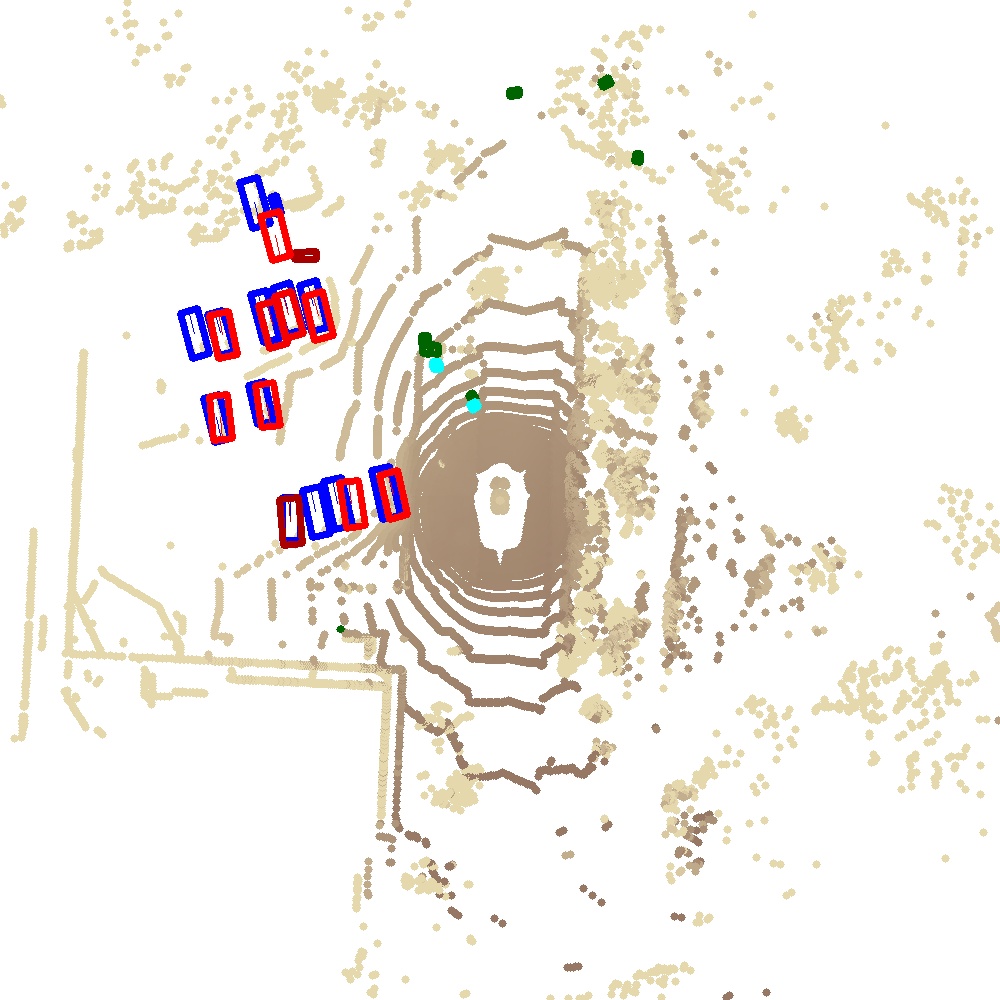}}
 \\
 \multirow{1}*[15mm]{\rotatebox[origin=c]{90}{Attack}} & 
 {\includegraphics[width=0.32\linewidth,height=0.22\linewidth]{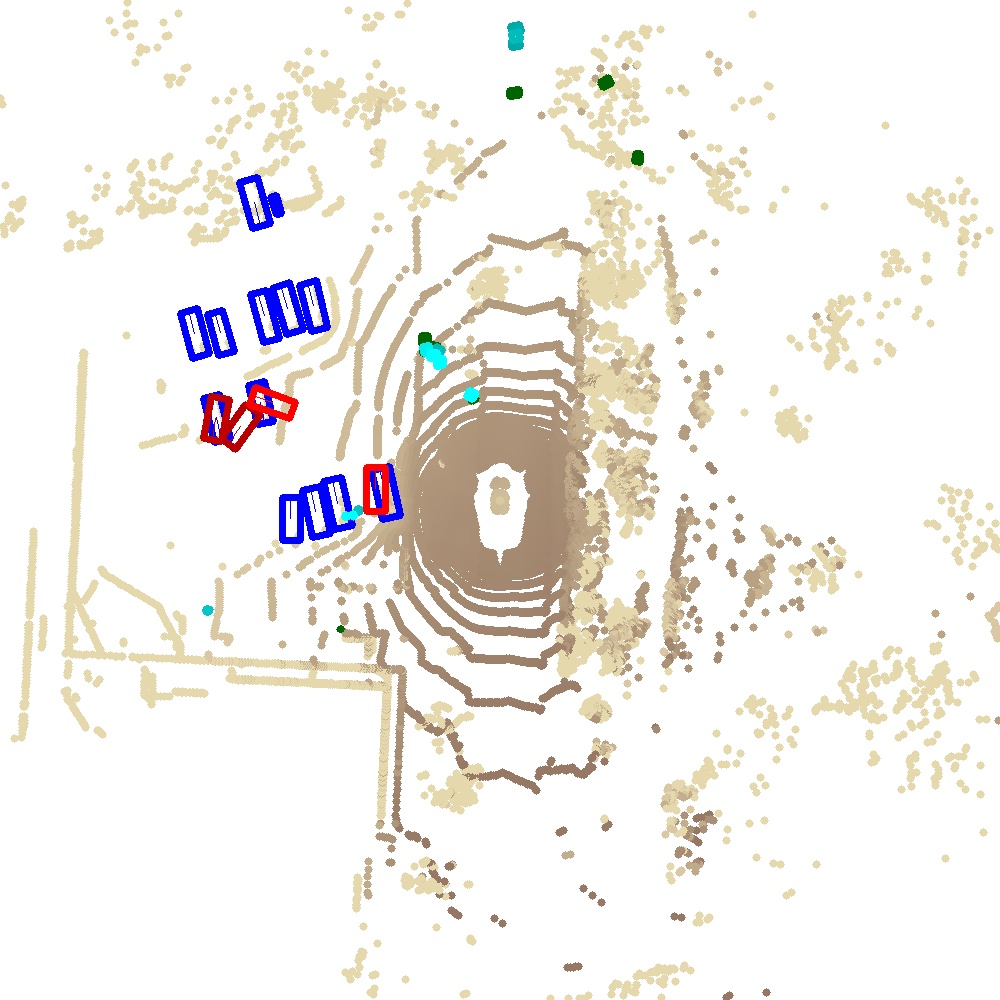}}&
 {\includegraphics[width=0.32\linewidth,height=0.22\linewidth]{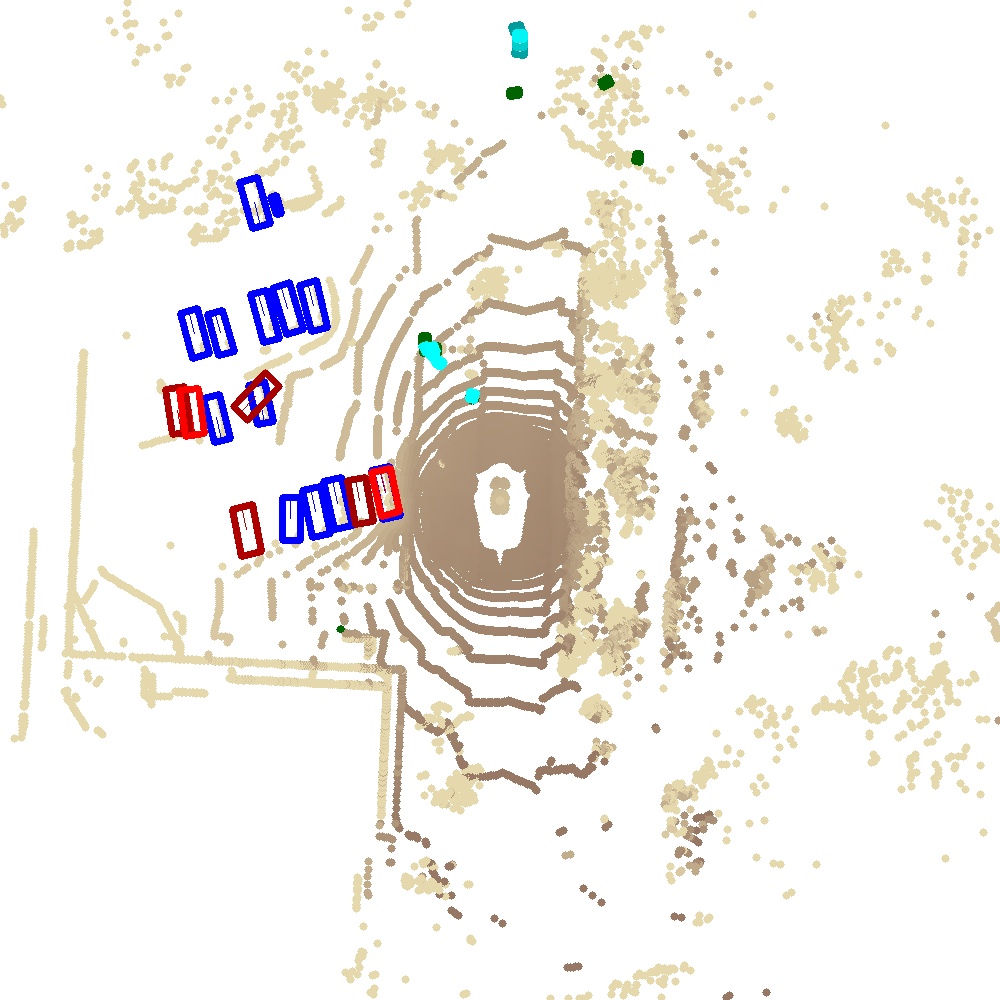}}&
 {\includegraphics[width=0.32\linewidth,height=0.22\linewidth]{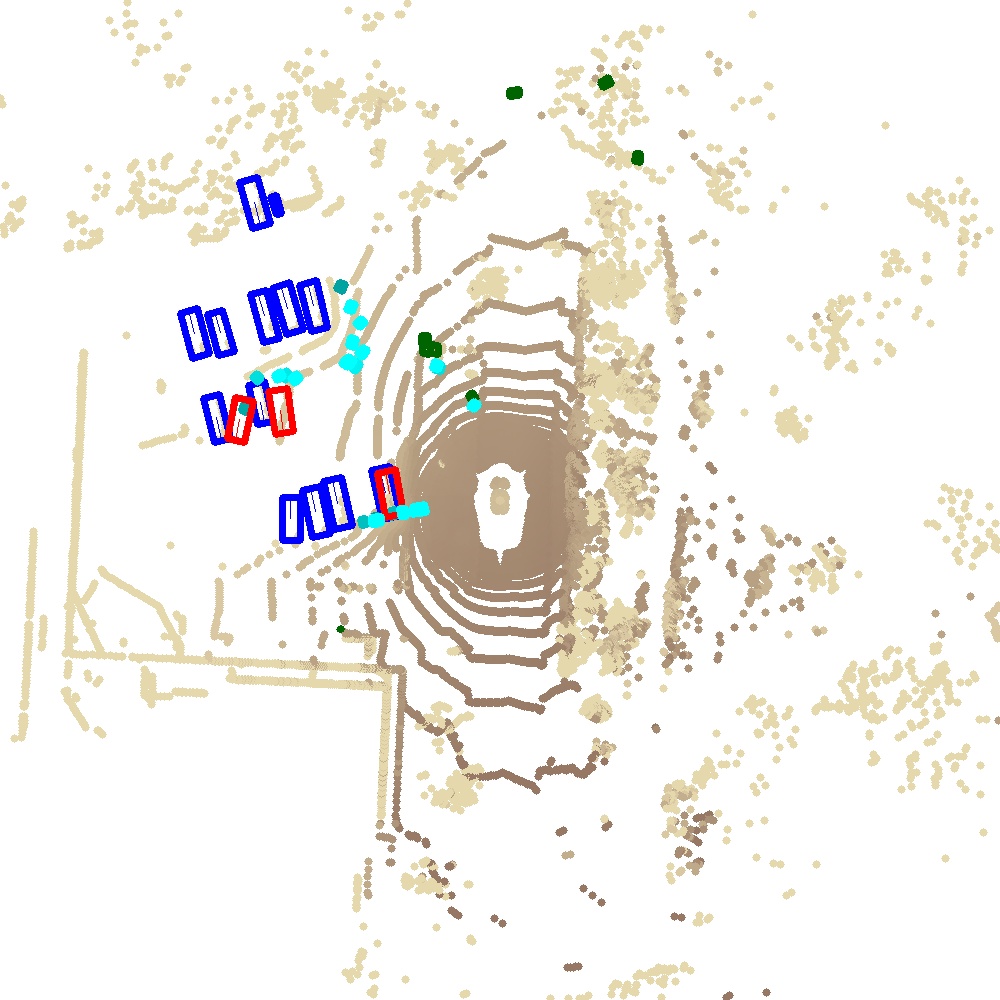}}

\\
   & Ours-BEVDet & Ours-BEVDet4D & Ours-BEVFormer  \\
   \end{tabular} 
    \caption{\textbf{Visualizations of attack effects in the BEV.} Top: predictions with initial objects. Bottom: predictions after inserting the adversarial object. Blue/red indicate ground-truth/predicted boxes for vehicles; green/cyan for other object categories.} 
    \label{fig:bev_compare1}
\end{figure*}

Considering the varying object sizes in the nuScenes dataset, we follow the official nuScenes evaluation protocol, which does not directly utilize IoU.
\cref{tab:white_box_Car} reports the mean Average Precision (mAP) for the vehicle category and detection results of cars at different sensor ranges. For example, $\text{AP}_{car0.5}$  denotes the average precision for vehicles within 0.5 meters of the sensor. The results demonstrate that our 3D adversarial samples consistently reduce vehicle detection accuracy, particularly within 0.5 meters, highlighting both the vulnerability of current BEV models and the potential for severe impact in close-range scenarios. In addition, we show more visualized comparisons in \cref{fig:2d_compare1} and \cref{fig:bev_compare1}.
\begin{table}[t]
\centering
  \setlength{\tabcolsep}{1.5pt} 
  \caption{\textbf{Performance comparisons using additional evaluation metrics} for BEV-based 3D object detection under adversarial attacks with Real Occ. }
  \label{tab:white_box_Car}
    \resizebox{1.0\linewidth}{!}{
     \setlength{\tabcolsep}{1pt}
     \LARGE
  \centering
 \begin{tabular}{l|cc|cc|cc|cc|cc} 
 \toprule

& \multicolumn{2}{c|}{Init/Adv $\text{mAP}_{vel}$}  & \multicolumn{2}{c|}{Init/Adv $\text{AP}_{car0.5}$}  & \multicolumn{2}{c|}{Init/Adv $\text{AP}_{car1.0}$}&  \multicolumn{2}{c|}{Init/Adv $\text{AP}_{car2.0}$} & \multicolumn{2}{c}{Init/Adv $\text{AP}_{car4.0}$}   \\
\midrule
BEVDet    &  0.205  &  0.073 &  0.1684  &  0.0470 &   0.4163 &  0.1924 &   0.6300 & 0.3413 & 0.7481 & 0.4432 \\
BEVDet4D  &  0.216  & 0.085  &  0.1761  &  0.0390 &  0.4282 & 0.1779 & 0.6465 & 0.3303 & 0.7535 & 0.4662 \\
BEVFormer &  0.297  & 0.154  &  0.1790  &  0.0822 &  0.4521 &  0.2976 & 0.7142 & 0.5398 &  0.8367 &0.6835 \\
\bottomrule

     \end{tabular}}
\end{table}

\end{document}